\documentclass[preprint,3p]{elsarticle}

\usepackage{amssymb}
\usepackage{hyperref}
\usepackage{natbib}
\usepackage{algorithm}
\usepackage{algpseudocode}
\usepackage{amsmath}
\usepackage{multirow}
\usepackage{tabularx,siunitx}
\usepackage{threeparttable}
\usepackage{subcaption}
\usepackage{float}
\usepackage{caption}
\usepackage{amsthm}

\usepackage{enumitem}
\usepackage{array}
\usepackage[table]{xcolor}
\usepackage{lineno}

\journal{Journal of Computational Design and Engineering}

\begin{document}

\begin{frontmatter}


% \title{ADAMS-AR: \textbf{A}dams-bashforth \textbf{D}riven \textbf{A}daptive \textbf{M}ulti-\textbf{S}tep \textbf{A}uto-\textbf{R}egression \\for Enhanced Long-Term Spatio-Temporal Prediction}
% \title{ADAMS-AR: \textbf{AD}ams-Bashforth \textbf{A}daptive \textbf{M}ulti-\textbf{S}tep \textbf{A}uto-\textbf{R}egression \\for Robust Long-Term Spatio-Temporal Prediction}
\title{Model-Agnostic AI Framework with Explicit Time Integration\\ for Long-Term Fluid Dynamics Prediction}


\author[inst1]{Sunwoong Yang}
\ead{sunwoongy@kaist.ac.kr}
\affiliation[inst1]{organization={Cho Chun Shik Graduate School of Mobility, Korea Advanced Institute of Science and Technology},%Department and Organization
            % addressline={Address One}, 
            city={Daejeon},
            postcode={34051}, 
            % state={State One},
            country={Republic of Korea}}

\author[inst2]{Ricardo Vinuesa}
\ead{rvinuesa@mech.kth.se}
\affiliation[inst2]{organization={FLOW, Engineering Mechanics, KTH Royal Institute of Technology},%Department and Organization
            % addressline={Address One}, 
            city={Stockholm},
            postcode={SE-100 44}, 
            % state={State One},
            country={Sweden}}

\author[inst1,inst3]{Namwoo Kang\corref{cor1}}

\affiliation[inst3]{organization={Narnia Labs},%Department and Organization
            % addressline={Address One}, 
            city={Daejeon},
            postcode={34051}, 
            % state={State One},
            country={Republic of Korea}}

\cortext[cor1]{Corresponding author. \texttt{nwkang@kaist.ac.kr}. \texttt{https://orcid.org/0000-0003-3475-7477}}

% \linenumbers
\begin{abstract}
The integration of predictive AI into computational design and engineering workflows is often hindered by the challenge of long-term forecasting performance. This study presents a robust framework for AI-accelerated flow simulation, specifically addressing the critical issue of error accumulation in auto-regressive (AR) surrogate models, which is a key bottleneck for their practical use in the design cycle. We introduce the first implementation of the two-step Adams-Bashforth method specifically tailored for data-driven AR prediction, leveraging historical derivative information to enhance numerical stability without additional computational overhead. To validate our approach, we systematically evaluate time integration schemes across canonical 2D PDEs (advection, heat, and Burgers' equations) before extending to complex Navier-Stokes cylinder vortex shedding dynamics. Additionally, we develop three novel adaptive weighting strategies that dynamically adjust the importance of different future time steps during multi-step rollout training. Our analysis reveals that as physical complexity increases, such sophisticated rollout techniques become essential, with the Adams-Bashforth scheme demonstrating consistent robustness across investigated systems and our best adaptive approach delivering an 89\% improvement over conventional fixed-weight methods while maintaining similar computational costs. For the complex Navier-Stokes vortex shedding problem, despite using an extremely lightweight graph neural network with just 1,177 trainable parameters and training on only 50 snapshots, our framework accurately predicts 350 future time steps—a 7:1 prediction-to-training ratio---reducing mean squared error from 0.125 (single-step direct prediction) to 0.002 (Adams-Bashforth with proposed multi-step rollout). Our integrated methodology demonstrates an 83\% improvement over standard noise injection techniques and maintains robustness under severe spatial constraints; specifically, when trained on only a partial spatial domain, it still achieves 58\% and 27\% improvements over direct prediction and forward Euler methods, respectively. Our framework's model-agnostic design, operating at the fundamental level of AR prediction mechanics, enables direct integration with any neural network architecture without requiring model-specific modifications, introducing a versatile solution for robust long-term spatio-temporal predictions across various engineering disciplines.
\end{abstract}


% \begin{highlights}
% \end{highlights}


\end{frontmatter}

% \linenumbers

\section{Introduction}
\label{sec:intro}

In the era of scientific machine learning (SciML), auto-regressive (AR) temporal prediction have emerged as powerful tools for spatio-temporal prediction across various engineering disciplines, particularly in physical domains like fluid dynamics \cite{solera2024beta, liu2024multi, hasegawa2020cnn, lee2019data, list2025differentiability, akhare2023physics}. They predict future states by recursively using their own previous predictions as inputs for subsequent forecasts---specifically, each new prediction becomes part of the input sequence for the next prediction step, creating a chain of sequential forecasts based on historical data. This recursive approach facilitates real-time forecasting and dynamic decision-making across a wide range of engineering applications \cite{taieb2012review, ahani2019statistical, samal2022multi, wang2021automated, chang2008forecast, asadi2012new}. Unlike coordinate-based prediction approaches that incorporate the time coordinate directly into the input and therefore violate temporal causality \cite{wang2024respecting, liu2022causality, nghiem2023causal}—a category to which emerging SciML methods such as physics-informed neural networks \cite{raissi2019physics} and DeepONet \cite{lu2021learning} belong—AR frameworks predict future states based on past historical information: noteworthy SciML models such as Fourier neural operator \cite{li2020fourier} and MeshGraphNet \cite{pfaff2020learning} are implemented within AR frameworks. This sequential prediction aligns with the natural progression of physical processes, preserving causality by ensuring that each snapshot at a given time depends solely on preceding snapshots, without any influence from future information. 

However, AR models have inherent limitations, most notably error accumulation during long-term rollouts \cite{liu2024multi, venkatraman2015improving, kim2024physics, zhang2025goal, gao2024bayesian}. Since each prediction depends on the previous output, any inaccuracies introduced at one step can propagate and amplify in subsequent steps, leading to significant deviations from the true physical behavior over time. Recent approaches attempt to address this limitation by combining data-driven models with traditional numerical solvers, using the latter to recalibrate data-driven predictions when SciML model errors exceed certain thresholds \cite{jeon2022finite, jeon2024residual}. However, such hybrid approaches compromise the primary advantage of ML-based surrogate models, their real-time prediction capability, as demonstrated by insufficient speedup factors: for example, approximately 1.9 times acceleration compared to pure computational fluid dynamics (CFD) simulations \cite{jeon2024residual}. This highlights the critical need for improving the long-term prediction accuracy of purely data-driven AR models while preserving their computational efficiency, enabling real-time predictions without relying on expensive numerical solver re-calibrations.

To enhance the robustness of AR-based spatio-temporal predictions over long-term rollouts, one of the most frequently used approaches is noise injection, where noise is added to the input data during the training phase \cite{kim2024physics, pfaff2020learning, sanchez2020learning,yang2024enhancing}. This method deliberately perturbs the training inputs with random noise to simulate the prediction errors that naturally occur during rollout. By exposing the model to slightly corrupted inputs during training, it learns to handle imperfect data and becomes more robust when its own imperfect predictions are fed back as inputs during inference. However, noise injection requires careful tuning of the noise scale, which is highly data-dependent, and its stochastic nature can lead to inconsistent and unstable training process.

A more structured approach involves framing the learning task as predicting the temporal derivative of the system, which is then advanced in time by a numerical integrator. This concept has shown promise, for instance, in the work by \citet{zhou2025predicting}, which highlights the benefits of predicting change rather than states. However, their framework relies on including the time coordinate $t$ as a direct input to the AI model. While effective for interpolation, this approach is not truly auto-regressive and circumvents the core challenge of error accumulation from recursive, self-generated inputs. Consequently, \citet{zhou2025predicting} evaluated such non-causal models on their ability to interpolate solutions within a trained time domain and do not address the time-extrapolation. Therefore, advancing the time-extrapolation capabilities of AR models is not just a valuable research direction but a fundamental necessity for practical forecasting applications. Our work addresses this critical, unexplored area by being the first to integrate several numerical schemes and novel multi-step rollout training strategies within a true AR framework. We rigorously evaluate our approaches on their time-extrapolation capabilities across multiple datasets, assessing their performance in a forecasting scenario, which provides a more challenging and realistic test of generalization than evaluations conducted within a fixed and trained time domain.

This study presents a novel framework for enhancing long-term AR predictions by integrating numerical time-integration schemes and adaptive multi-step rollout techniques. Importantly, our approach is designed to be model-agnostic and application-independent, focusing on fundamental AR prediction mechanics rather than domain-specific features. This ensures our techniques can be seamlessly integrated with existing AR frameworks—whether based on graph neural networks, convolutional architectures, or neural operators adapted for AR prediction—without requiring specialized adaptations. Our approach is validated across multiple problems, from canonical 2D partial differential equations (heat, advection, and Burgers' equations) to complex Navier-Stokes equations around a circular cylinder, demonstrating its versatility and robustness across different physical systems. Our key contributions can be categorized into three primary areas:

\paragraph{\textbf{Time Integration Innovations}}
\begin{enumerate}  
    \item \textbf{Comprehensive evaluation of time integration schemes for AR prediction:} For the first time in AR prediction within SciML models, we systematically explore various time integration schemes across both canonical PDEs and complex fluid dynamics. We demonstrate that the Adams-Bashforth scheme consistently outperforms other approaches, achieving an improvement over the commonly used forward Euler method in Navier-Stokes simulations while maintaining robust performance across varying prediction horizons and different physical systems.
    
    \item \textbf{Novel adaptation of Adams-Bashforth for AR frameworks:} We introduce the first implementation of the two-step Adams-Bashforth method specifically tailored for data-driven AR prediction, leveraging historical derivative information to enhance numerical stability and long-term accuracy without additional computational overhead.
\end{enumerate}

\paragraph{\textbf{Adaptive Multi-Step Rollout Innovations}}
\begin{enumerate}[resume]
    \item \textbf{Development of adaptive weighting strategies:} We propose three novel adaptive weighting approaches that dynamically adjust the importance of different future time steps during multi-step rollout training. Our best strategy, emphasizing only the first and last future components, delivers an 89\% improvement in rollout performance over conventional fixed-weight multi-step rollout in the Navier-Stokes cylinder dataset while maintaining similar computational costs.
    
    \item \textbf{Systematic comparison with existing techniques:} Our integrated approach achieves an 83\% improvement in prediction accuracy compared to conventional noise injection techniques in the Navier-Stokes cylinder dataset, also revealing previously unidentified negative interactions between multi-step rollout and noise injection strategies.
\end{enumerate}

\paragraph{\textbf{Framework Robustness and Scalability}}
\begin{enumerate}[resume]
    \item \textbf{Performance under resource-constrained conditions:} We validate our framework under intentionally challenging scenarios across three computational limitations. For the Navier-Stokes cylinder flow, we demonstrate: (1) \textit{limited model capacity} - achieving accurate long-term predictions (up to 350 rollouts) with a lightweight model containing only 1,177 trainable parameters; (2) \textit{minimal training data} - using 50 past snapshots for training; and (3) \textit{partial domain training} - maintaining 58\% and 27\% improvements over direct prediction and forward Euler approaches when trained on spatially constrained mesh regions, confirming robustness even when spatial information is severely limited.
    
    \item \textbf{Model-agnostic methodology:} Both our Adams-Bashforth time integration adaptation and adaptive multi-step rollout strategies are designed to be architecture-independent, operating at the fundamental level of AR prediction mechanics. This enables direct integration with any neural network architecture—graph neural networks, convolutional networks, transformers, or neural operators—without requiring model-specific modifications or adaptations.
\end{enumerate}

The remainder of this paper is organized as follows. Section~\ref{sec:method} presents the theoretical framework of our proposed time integration schemes and adaptive multi-step rollout strategies. To establish baseline performance and demonstrate fundamental effectiveness, Section~\ref{sec:numerical_pdes} evaluates our framework on canonical 2D partial differential equations (advection, heat, and Burgers' equations) using multi-layer perceptrons. Section~\ref{sec:NS_all} then transition to complex engineering applications, focusing on vortex shedding dynamics around a circular cylinder governed by Navier-Stokes equations using graph neural networks. Specifically, Section~\ref{sec:Exp} details the experimental setup for the cylinder flow case, Section~\ref{sec:OnlyTime} demonstrates the effectiveness of time integration schemes, Section~\ref{sec:extend_multistep} extends these schemes with conventional multi-step rollout, Section~\ref{sec:AWmulti} introduces our adaptive weighting approaches, Section~\ref{sec:noise} provides comparative analysis with noise injection methods, and Section~\ref{sec:harsh} evaluates performance under challenging partial domain training conditions. Finally, Section~\ref{sec:conclu} concludes with broader implications and future research directions.

\section{Methods: Explicit Time Integration Schemes and Adaptive Multi-Step Rollout}
\label{sec:method}

\subsection{Time integration schemes for auto-regressive prediction}
\label{sec:method_time}

\begin{figure*}[htb!]
    \centering
        \includegraphics[width=\textwidth]{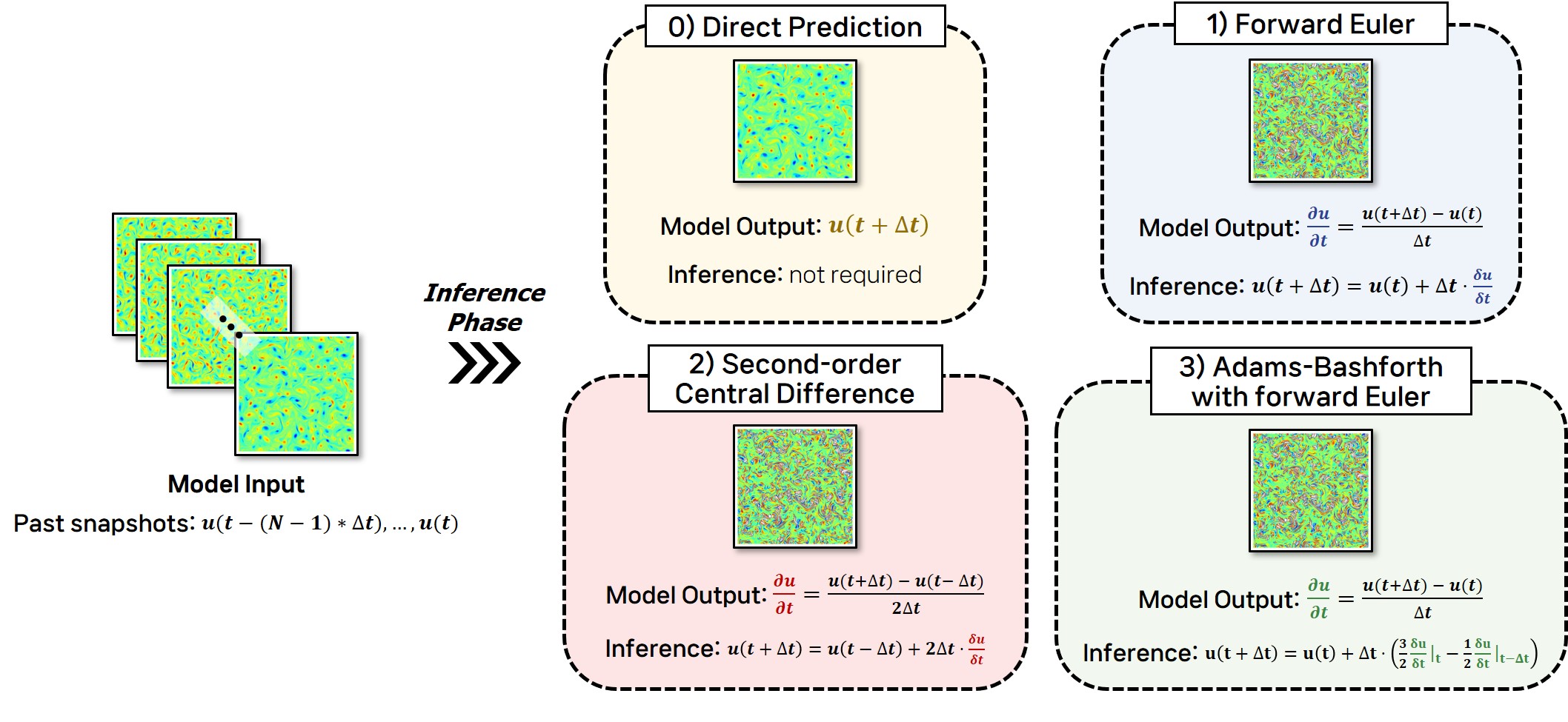}
    \caption{Time integration schemes investigated in this study. Forward Euler and Adams-Bashforth yield identical model outputs but differ in inference stage: forward Euler uses single-step explicit update (Eq.~\ref{eq:time1_3}), while Adams-Bashforth employs two-step approach (Eq.~\ref{eq:time4_0}).}
    \label{fig:timeinteg}
\end{figure*}

While advanced time-stepping schemes are well-established in traditional CFD, their application in AI-driven flow prediction remains limited. Most AI-based temporal prediction research focuses on direct prediction \cite{solera2024beta,hasegawa2020cnn,lee2019data,yang2024enhancing} or simple forward Euler methods \cite{liu2024multi,pfaff2020learning,sanchez2020learning}. We investigate various numerical schemes to enhance long-term AR prediction accuracy, where each finite difference method provides different structural biases for learning temporal dynamics---from simple first-order approximations in forward Euler to symmetric formulations in central difference schemes to the multi-step approach of Adams-Bashforth which incorporates richer temporal history. We explore these three finite difference schemes and compare them with conventional direct prediction. Figure~\ref{fig:timeinteg} visualizes the investigated schemes. The Runge-Kutta method is excluded due to substantial computational overhead (details in \ref{sec:app_Runge}).

\paragraph{\textbf{Direct prediction}} The AR model directly predicts the next time step:
\begin{equation} 
\label{eq:time0} 
\mathbf{u}(t+\Delta t) = AR\left( \mathbf{u}(t), \mathbf{u}(t - \Delta t), \dots, \mathbf{u}(t - (N-1)\Delta t) \right)
\end{equation} 

\paragraph{\textbf{Forward Euler method}} The AR model predicts the temporal derivative:
\begin{equation} 
\label{eq:time1_1} 
\frac{\delta \mathbf{u}}{\delta t} = AR\big( \mathbf{u}(t), \mathbf{u}(t-\Delta t), \dots, \mathbf{u}(t - (N-1)\Delta t) \big)
\end{equation}
where $\frac{\delta \mathbf{u}}{\delta t} = \frac{\mathbf{u}(t+\Delta t) - \mathbf{u}(t)}{\Delta t}$, resulting in the following equation for next snapshot prediction:
\begin{equation} 
\label{eq:time1_3} 
\mathbf{u}(t+\Delta t) = \mathbf{u}(t) + \Delta t \cdot \frac{\delta \mathbf{u}}{\delta t} 
\end{equation}

\paragraph{\textbf{Second-order central difference}} Similar to forward Euler but with:
\begin{equation} 
\label{eq:time2_1} 
\frac{\delta \mathbf{u}}{\delta t} = \frac{\mathbf{u}(t+\Delta t) - \mathbf{u}(t-\Delta t)}{2\Delta t}
\end{equation}
resulting in: $\mathbf{u}(t+\Delta t) = \mathbf{u}(t-\Delta t) + 2\Delta t \cdot \frac{\delta \mathbf{u}}{\delta t}$ for next snapshot prediction \cite{hussain2025integrating}.

\paragraph{\textbf{Adams-Bashforth with forward Euler (Adams-Euler)}} The two-step Adams-Bashforth predicts the next snapshot as:
\begin{equation} 
\label{eq:time4_0} 
\mathbf{u}(t+\Delta t) = \mathbf{u}(t) + \Delta t \cdot \left( \frac{3}{2} \frac{\delta \mathbf{u}}{\delta t}\big|_{t} - \frac{1}{2} \frac{\delta \mathbf{u}}{\delta t}\big|_{t-\Delta t} \right)
\end{equation}

A critical aspect of implementing the Adams-Bashforth method in an AR framework is the choice of how to approximate the derivatives $\frac{\delta \mathbf{u}}{\delta t}\big|_{t}$ and $\frac{\delta \mathbf{u}}{\delta t}\big|_{t-\Delta t}$. We implement the forward Euler method for time derivatives (Eq. \ref{eq:time4_1} and Eq. \ref{eq:time4_2}), a common approach in AI-based temporal prediction \cite{liu2024multi,pfaff2020learning,sanchez2020learning}: for brevity, we refer to this combined Adams-Bashforth with forward Euler method as ``Adams-Euler'' throughout this paper. Since forward Euler and Adams-Euler methods employ the forward Euler scheme to compute time derivatives, those approaches yield identical model outputs: see Eq. \ref{eq:time0} and Eq. \ref{eq:time4_1}. Given this equivalence in training, a single trained model could theoretically be used for both forward Euler and Adams-Euler inference. However, for ease of experimental implementation, we simply trained separate models for each scheme. To ensure this approach provides a fair and robust comparison, each experiment was repeated multiple times using different random seeds, and the averaged results are presented throughout the paper. Consequently, the minor variations in reported training times between the forward Euler and Adams-Euler schemes are a natural result of these separate training runs and normal stochastic fluctuations in GPU computation.

\begin{equation} 
\label{eq:time4_1} 
\frac{\delta \mathbf{u}}{\delta t}\big|_{t} = \frac{\mathbf{u}(t+\Delta t) - \mathbf{u}(t)}{\Delta t} = AR\left( \underbrace{\mathbf{u}(t), \mathbf{u}(t - \Delta t), \dots , \mathbf{u}(t - (N-1)\Delta t)}_{\text{past $N$ snapshots from $t$}} \right)
\end{equation}

\begin{equation} 
\label{eq:time4_2} 
\frac{\delta \mathbf{u}}{\delta t}\big|_{t-\Delta t} = \frac{\mathbf{u}(t) - \mathbf{u}(t-\Delta t)}{\Delta t} = AR\left( \underbrace{\mathbf{u}(t-\Delta t), \mathbf{u}(t-2\Delta t), \dots, \mathbf{u}(t - N\Delta t)}_{\text{past $N$ snapshots from ($t-\Delta t$)}} \right)
\end{equation}

\subsection{Adaptive multi-step rollout}
\label{sec:method_multi}

Conventional AR models trained with single-step prediction often struggle with error accumulation during long-term rollouts. In single-step training, the model learns to predict only one step ahead using ground truth data as input, but during inference, it must use its own predictions as inputs for subsequent steps. This creates a mismatch between training and inference conditions, as the model is never exposed to its own prediction errors during training.

\begin{figure*}[htb!]
    \centering
        \includegraphics[width=\textwidth]{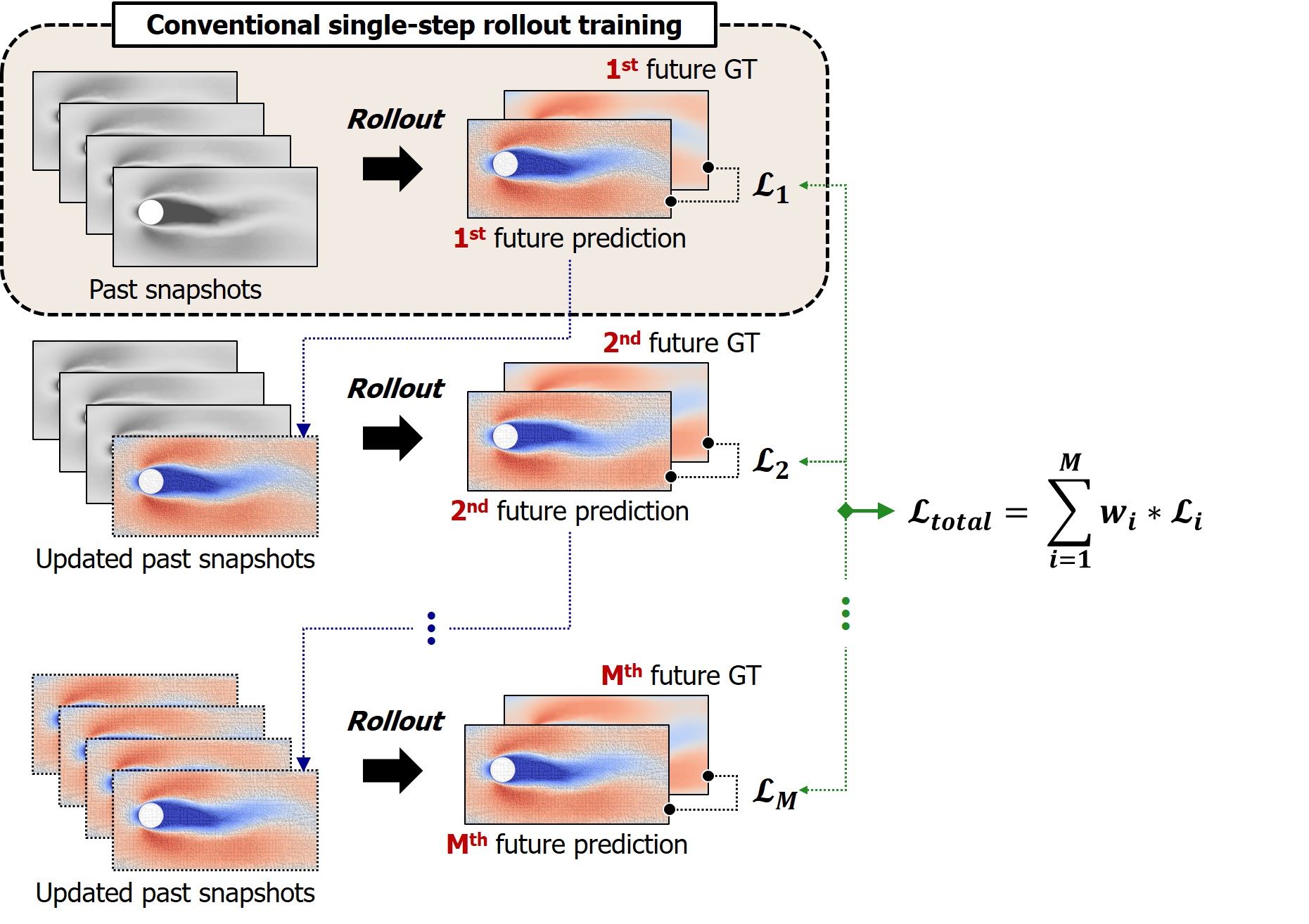}
    \caption{Multi-step rollout: the model predicts $M$ future steps during training, with total loss computed as weighted sum of individual prediction losses. Our adaptive weighting schemes dynamically determine weights ($w_i$).}
    \label{fig:multistep}
\end{figure*}

Multi-step rollout \cite{wu2022learning} addresses this limitation by incorporating multiple future time steps into the training process: Figure \ref{fig:multistep}. Specifically, given input snapshots $\mathbf{u}(t), \ldots, \mathbf{u}(t-(N-1))$, the AR framework first predicts $\hat{\mathbf{u}}(t+1)$. The input history is then updated to $\hat{\mathbf{u}}(t+1), \ldots, \mathbf{u}(t-(N-2))$, allowing prediction of $\hat{\mathbf{u}}(t+2)$. This iterative process continues for $M$ steps, producing predictions from $\hat{\mathbf{u}}(t+1)$ to $\hat{\mathbf{u}}(t+M)$ based on the initial input history. The total loss is defined as a weighted sum of individual step losses:

\begin{equation}
\label{eq:multi-loss}
\mathcal{L}_{multi-step} = \sum_{i=1}^{M} {w_i \mathcal{L}_i}
\end{equation}
where $w_i$ represents the weight assigned to the loss at each future time step $t+i$, and $\mathcal{L}_i$ is the loss between the model's prediction $\hat{\mathbf{u}}(t+i)$ and the ground truth $\mathbf{u}(t+i)$. By explicitly optimizing across multiple steps, the model enhances its resilience to compounding errors during inference.

However, conventional multi-step rollout approaches present two main challenges: (1) Manual weight tuning ($w_i$) is problem-dependent and requires trial-and-error adjustment. For instance, a conventional strategy is to use fixed weights, setting $w_1 = 1$ for the first future step and $w_i = 0.1$ for all subsequent steps \cite{wu2022learning}. Such fixed values may not be optimal as they vary significantly across different datasets and problem types, and reckless weight tuning can destabilize training, leading to overfitting or underfitting issues. (2) The simultaneous consideration of all $M$ future steps complicates the optimization process. This approach requires the model to account for every snapshot in the long-term prediction, which can be particularly problematic during the early stages of training when long-term predictions are highly uncertain.

To overcome these limitations, we propose three adaptive weighting strategies designed to enhance training efficiency and improve prediction accuracy over long time horizons:

\paragraph{\textbf{AW1: Adaptive weighting without learnable parameter}} The first approach mitigates the need for manual weight selection by automatically assigning weights based on prediction error magnitude. Weights are computed as normalized mean squared error (MSE) for each prediction step:
\begin{equation} 
\label{eq:AW1} w_i = \frac{\text{MSE}_i}{\sum_{j=1}^{M} \text{MSE}_j} 
\end{equation}
This method inherently emphasizes time steps with larger errors, as they receive higher weights. However, the approach may lack flexibility since weights are simply proportional to MSE values.

\paragraph{\textbf{AW2: Adaptive weighting with learnable parameter}} To introduce more flexibility, we incorporate a learnable parameter $k$ that dynamically adjusts the weighting scheme. The adaptive weights are computed using a power function of MSE values, modulated by the effective parameter $k_e$:
\begin{equation} 
\label{eq:AW2}
w_i = \frac{\text{MSE}_i^{k_e}}{\sum_{j=1}^M \text{MSE}_j^{k_e}}, \quad k_e = 0.5 + 2.5 \cdot \sigma(sk)
\end{equation}
where $\sigma$ represents a sigmoid activation function. The parameter $k$ is learnable, and $s$ acts as a scaling factor. We set $s=10$ to ensure sufficient sensitivity in the sigmoid function, allowing meaningful gradient flow during training. The range of $k_e$ is bounded between 0.5 and 3.0: values below 0.5 would flatten weight differences excessively, while values above 3.0 could create overly sharp weight distributions that destabilize training. Note that these values are tunable hyperparameters; for this study, the scaling factor $s$ and the bounds for $k_e$ were determined empirically and can be flexibly adjusted for other problems. The overall procedure of AW2 can be found in Algorithm \ref{alg:AW2}. 

\begin{algorithm}[htb!] 
\caption{AW2: adaptive weighting with learnable parameter} \label{alg:AW2} 
\begin{algorithmic}[1] 
\Require Model $\mathcal{M}$, input data $\mathbf{u}(t), \dots, \mathbf{u}(t-N+1)$, ground truth $\mathbf{u}(t+1), \dots, \mathbf{u}(t+M)$ 
\Ensure Updated model parameters and learnable parameter $k$ 
\State Initialize learnable parameter $k$ 
\For{each training iteration} 
\State Perform $M$-step rollout predictions recursively
\State Compute MSE for each step: $\text{MSE}_i = \text{MSE}(\hat{\mathbf{u}}(t+i), \mathbf{u}(t+i))$ for $i=1,\dots,M$ 
\State Calculate $k_e = 0.5 + 2.5 \cdot \text{sigmoid}(sk)$ 
\State Compute adaptive weights: $w_i = \frac{\text{MSE}_i^{k_e}}{\sum_{j=1}^M \text{MSE}_j^{k_e}}$
\State Calculate total loss: $\mathcal{L} = \sum_{i=1}^M w_i \cdot \text{MSE}_i$ 
\State Update model parameters and parameter $k$ through backpropagation
\EndFor 
\State \Return Updated model parameters and $k$ 
\end{algorithmic} 
\end{algorithm}

\paragraph{\textbf{AW3: Simplified adaptive weighting}} Recognizing that error accumulation is the primary failure mode in AR prediction, we frame the multi-step learning task as a trajectory control problem. The goal is to enforce both short-term accuracy and long-term stability. To this end, we propose a simplified and robust weighting strategy that focuses only on the two most critical points of the rollout trajectory: the beginning and the end.
\begin{itemize}
    \item \textbf{First step ($w_1$):} This term enforces short-term accuracy. A precise prediction for the very next time step is crucial, as any initial error will immediately propagate and compound, causing the trajectory to diverge quickly from the ground truth. This penalty ensures the forecast begins on the correct path, translating the momentum of the historical data into the very first predicted step.
    \item \textbf{Last step ($w_M$):} This term enforces long-term stability. By penalizing the error at the final point of the rollout window, the model is explicitly regularized against divergence. This loss acts as a constraint on the future, forcing the model to learn dynamics that remain stable over the entire horizon.
\end{itemize}
This approach is motivated by curriculum learning \cite{elman1993learning}, an AI training strategy where a model is first taught simpler concepts before progressing to more complex ones. Our adaptive weighting mechanism automatically manages this curriculum. It dynamically adjusts the focus between the short-term (easy) and long-term (hard) objectives based on their relative errors, allowing the model to efficiently learn a stable trajectory without the chaotic gradients from intermediate steps; however, please note that this approach can fail when excessively hard objective is adopted such as very large $M$ value in this study. The adaptive weights are computed as:

\begin{align}\label{eq:AW3} 
w_1 = \frac{\text{MSE}_1^{k_e}}{\text{MSE}_1^{k_e} + \text{MSE}_M^{k_e}}, \quad 
w_M = \frac{\text{MSE}_M^{k_e}}{\text{MSE}_1^{k_e} + \text{MSE}_M^{k_e}}
\end{align}

The progression from AW1 to AW3 provides increasing sophistication: AW1 offers simplicity with automatic weighting, AW2 adds flexibility through learnable parameter, and AW3 reduces complexity while retaining adaptivity by focusing on the most critical prediction steps.

\subsection{Evaluation metric}
\label{sec:method_eval}

Throughout this paper, we assess long-term prediction capability using rollout performance over extended time horizons. All presented rollout performance results are evaluated using the following error metric, averaged MSE over all predicted snapshots:
\begin{equation} 
\label{eq:MSE} 
\text{MSE} = \frac{1}{S} \sum_{s=1}^S \left( \frac{1}{N} \sum_{n=1}^N \left( y_{n,s} - \hat{y}_{n,s} \right)^2 \right)
\end{equation}
where $\hat{y}_{n,s}$ and $y_{n,s}$ are the predicted and ground-truth values at spatial location $n$ for future time step $s$, respectively. $N$ represents the total number of spatial locations and $S$ denotes the number of future snapshots evaluated during the rollout process. This MSE metric serves as our primary indicator of rollout performance throughout all experiments in this study.

In addition to MSE, for the periodic cylinder flow problem, we evaluate the physical fidelity of the predictions using the Strouhal number ($St$). This non-dimensional quantity characterizes the dominant frequency of vortex shedding, making it a powerful metric for assessing whether the model has learned the underlying flow dynamics beyond simple numerical accuracy. To calculate it, we apply a fast Fourier transform to the $x$-velocity time series at a certain point in the cylinder's wake. The signal is first preprocessed by removing the mean and applying a Hamming window to reduce spectral leakage. The dominant frequency, $f_d$, is identified from the peak of the resulting power spectrum. The Strouhal number is then computed as:
\begin{equation} 
\label{eq:Strouhal} 
St = \frac{f_d D}{U_{\infty}}
\end{equation}
where $D$ is the characteristic cylinder diameter and $U_{\infty}$ is the freestream velocity.

\section{Preliminary Studies on Canonical PDEs}
\label{sec:numerical_pdes}

To establish a baseline and demonstrate the versatility of our proposed framework, we first evaluate its performance on a set of canonical PDEs. These simpler systems provide a controlled environment to analyze the fundamental interactions between the time integration schemes and the AR model before moving to more complex fluid dynamics problems in Section~\ref{sec:NS_all}.

\subsection{Dataset generation and experimental setup}
\label{sec:pde_setup}

We generate datasets for three fundamental 2D PDEs: the advection equation (pure convection), the heat equation (pure diffusion), and the Burgers' equation (convection-diffusion).

\subsubsection{Explored PDEs}

The governing equations for the scalar quantity $u(x,y,t)$ are defined as:
\begin{enumerate}
    \item \textbf{2D Advection Equation:} Represents convective transport.
    \begin{equation}
        \partial_t u + \mathbf{c} \cdot \nabla u = 0
    \end{equation}
    where $\mathbf{c} = [c_x, c_y]$ is the constant velocity vector.
    \item \textbf{2D Heat Equation:} Represents diffusive processes.
    \begin{equation}
        \partial_t u - \nu \nabla^2 u = 0
    \end{equation}
    where $\nu$ is the diffusion coefficient (viscosity).
    \item \textbf{2D Burgers' Equation:} A non-linear equation combining convection and diffusion.
    \begin{equation}
        \partial_t u + u(\mathbf{c} \cdot \nabla u) - \nu \nabla^2 u = 0
    \end{equation}
\end{enumerate}

\subsubsection{Initial and boundary conditions}

For all three PDEs, the solution is computed on a square domain of $(x,y) \in [-1, 1]^2$, which is discretized using a uniform $64 \times 64$ grid, and periodic boundary conditions are applied. The simulation runs from $t=0$ to $t=2s$, captured over 500 uniformly discretized time snapshots. To generate a diverse dataset, the initial condition at $t=0$ for each sample is randomly generated using a sum of five sine waves:
\begin{equation}
\label{eq:sine_ic}
u(x,y,0) = \sum_{j=1}^{5} A_j \sin\left(\frac{2\pi l_{xj}x}{L} + \frac{2\pi l_{yj}y}{L} + \phi_j\right)
\end{equation}
where the domain size $L=2$, the amplitude $A_j$ is sampled from $\mathcal{U}[-0.5, 0.5]$, the wavenumbers $l_{xj}, l_{yj}$ are sampled from $\{1, 2, 3\}$, and the phase shift $\phi_j$ is sampled from $\mathcal{U}[0, 2\pi]$. These settings are adopted from \citet{zhou2024strategies}.

\subsubsection{Dataset generation}

We generate 50 distinct simulation samples for each of the three PDEs. The physical parameters for each sample are randomly drawn from the following ranges:

\begin{itemize}
    \item \textbf{Advection Equation:} $c_x, c_y \in [0.1, 2.5]$
    \item \textbf{Heat Equation:} $\nu \in [2 \times 10^{-3}, 2 \times 10^{-2}]$
    \item \textbf{Burgers' Equation:} $c_x, c_y \in [0.5, 1.0]$ and $\nu \in [7.5 \times 10^{-3}, 1.5 \times 10^{-2}]$
\end{itemize}

\subsection{Model training details}
\label{sec:pde_training}

For each of the 50 samples per PDE type, a separate standard multi-layer perceptron (MLP) model is trained, which is designed to be independent of spatial coordinates. This simple, pointwise model was deliberately chosen to test our model-agnostic temporal framework using a basic architecture with no inherent spatial inductive bias. This ensures the observed performance gains are directly attributable to the temporal methods themselves, providing a strong baseline before extending the framework to more complex, spatially-aware GNNs in Section~\ref{sec:NS_all}. Specifically, its input is a time-series vector representing the last 30 snapshots of the solution $u$ at a single query point in the domain, and its output is the prediction for that same point's next state. During both training and inference, the trained MLP is applied simultaneously and independently to every point in the spatial grid, effectively processing all spatial grid points in a large batch. This approach allows the model to learn a general rule for temporal evolution that can be utilized for inference at any spatial location. The final performance metrics are averaged across the 50 independent runs from 50 cases to ensure statistical robustness. The training hyperparameters are kept consistent across all experiments to facilitate a fair comparison:

\begin{itemize}
    \item \textbf{MLP Architecture:} A 5-layer MLP with hidden layer sizes of [32, 32, 32, 32, 32].
    \item \textbf{Input History:} The model uses the last $N=30$ snapshots as input.
    \item \textbf{Data Split:} The data is split chronologically along the time axis, where the first 80\% of the snapshots are used for training and the subsequent 20\% are reserved for testing to evaluate AR rollout performance.
    \item \textbf{Training Epochs:} Each model is trained for 250 epochs.
    \item \textbf{Optimizer and Learning Rate:} The Adam optimizer is used with an initial learning rate of $10^{-3}$, which is decayed by a factor of 0.9 every 50 epochs.
\end{itemize}

% #### Results

\subsection{Predictive performance of time integration schemes}
\label{sec:pde_results_table1}

We first evaluate the different time integration methods using a single-step rollout ($M=1$). As shown in Table \ref{tab:pde_time_mse_comparison}, all schemes that predict temporal derivatives consistently outperform the conventional direct prediction method. Note that training time based on NVIDIA 3090 GPU is also shown---the same GPU machine is utilized throughout this study. And also since this study adopted lightweight AI models, inference time is estimated to be negligible, and therefore its time is not reported separately.

\begin{table}[H]
	\centering
	\caption{Comparison of averaged test MSE and training time for different time integration schemes across the three canonical PDEs.}
	\label{tab:pde_time_mse_comparison}
	\begin{tabular}{ll|rr}
		\hline
		\textbf{PDE} & \textbf{Time Integration Scheme} & \textbf{Test MSE [$\times 10^{-6}$]} & \textbf{Training Time [s]} \\
		\hline
		\multirow{4}{*}{2D Advection} 
			& Direct prediction      & 7723    & 29.62 \\
			& Forward Euler          & \textbf{5143}  & 29.51 \\
			& Second-order central   & 8852    & 29.45 \\
			& Adams-Euler            & 7273    & 29.36 \\
		\hline
		\multirow{4}{*}{2D Heat} 
			& Direct prediction      & 76       & 34.65 \\
			& Forward Euler          & 21       & 34.90 \\
			& Second-order central   & 24       & 34.30 \\
			& Adams-Euler            & \textbf{16} & 35.15 \\
		\hline
		\multirow{4}{*}{2D Burgers} 
			& Direct prediction      & 357      & 34.53 \\
			& Forward Euler          & 112      & 35.17 \\
			& Second-order central   & 112      & 34.24 \\
			& Adams-Euler            & \textbf{95}  & 35.41 \\
		\hline
	\end{tabular}
\end{table}

Adams-Euler scheme demonstrates the most robust capability, achieving the lowest test MSE for both the diffusive heat equation and the non-linear Burgers' equation. An interesting and noteworthy result is observed for the purely convective 2D advection equation, where the simpler forward Euler method excels. This suggests that for systems with straightforward, linear dynamics, the additional historical information from a second past derivative ($\frac{\delta \mathbf{u}}{\delta t}\big|_{t}$ and $\frac{\delta \mathbf{u}}{\delta t}\big|_{t-\Delta t}$ in Eq~\ref{eq:time4_0}) used by Adams-Euler may be unnecessary, and a less complex scheme provides a more direct and effective learning target. However, as the physics become more complex with the introduction of diffusion (heat equation) and non-linearity (Burgers' equation), the advantage shifts decisively to the Adams-Euler scheme. Its ability to capture richer temporal dynamics by incorporating information from prior steps proves crucial for modeling these more intricate systems accurately. Crucially, as the training times for all integration schemes are nearly identical, the meaningful accuracy gains from the more advanced methods are achieved with no additional computational cost.

\begin{figure*}[htb!]
    \centering
    \begin{subfigure}[b]{0.7\textwidth}
        \centering
        \includegraphics[width=\textwidth]{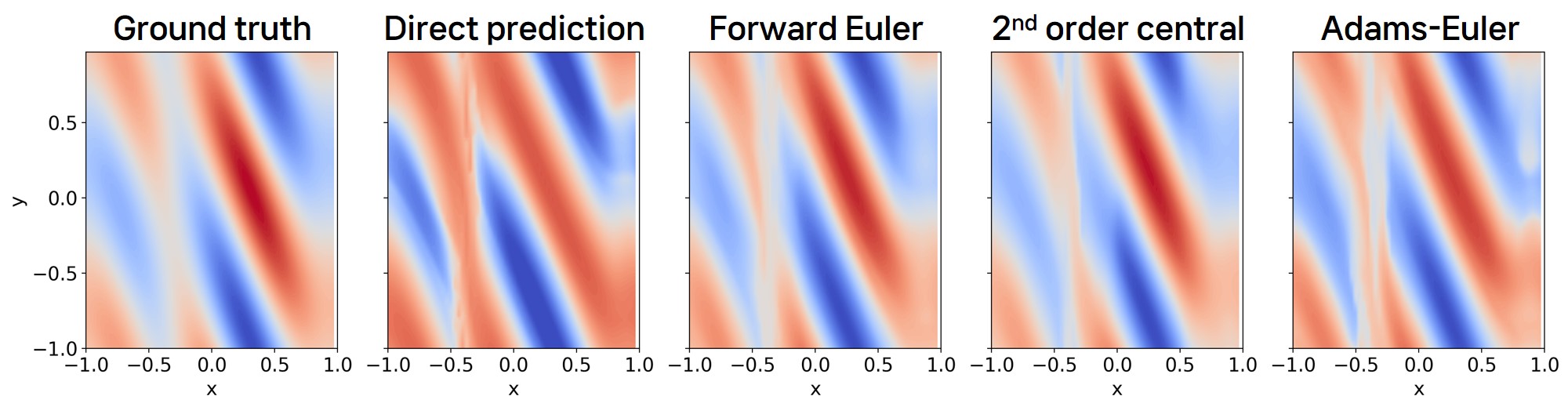}
        \caption{2D advection equation}
        \label{fig:pde_final_snapshot_advection}
    \end{subfigure}
    \vfill
    \begin{subfigure}[b]{0.7\textwidth}
        \centering
        \includegraphics[width=\textwidth]{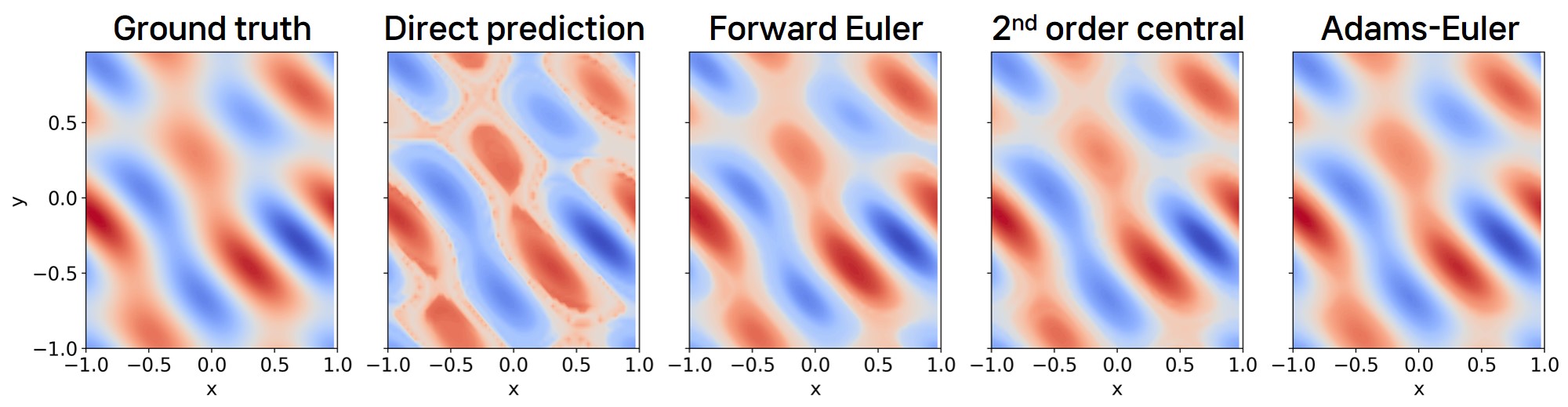}
        \caption{2D heat equation}
        \label{fig:pde_final_snapshot_heat}
    \end{subfigure}
    \vfill
    \begin{subfigure}[b]{0.7\textwidth}
        \centering
        \includegraphics[width=\textwidth]{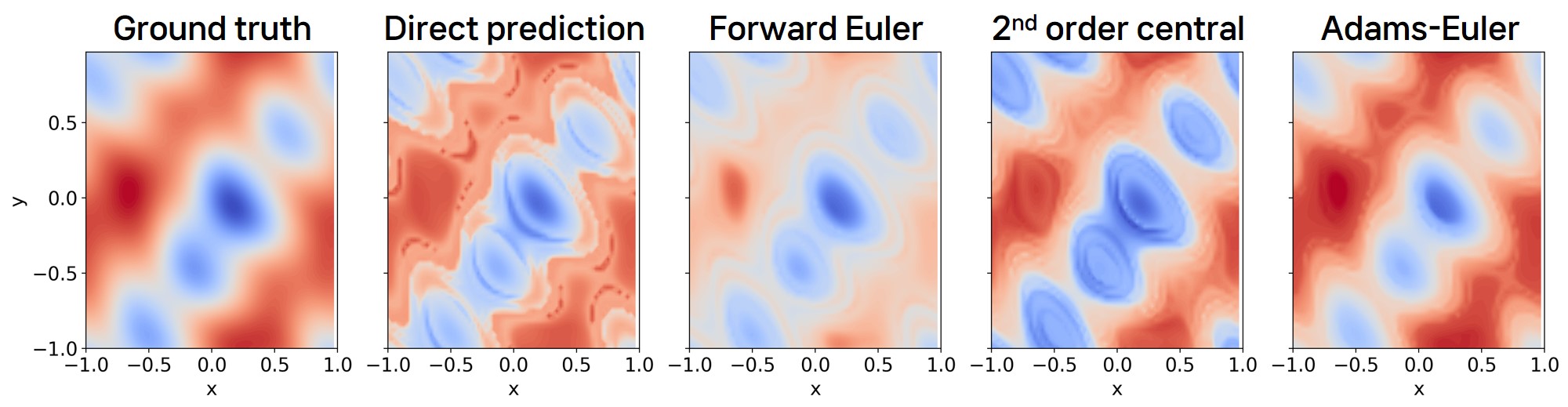}
        \caption{2D Burgers' equation}
        \label{fig:pde_final_snapshot_burgers}
    \end{subfigure}
    \caption{Qualitative comparison of the final predicted snapshot for representative samples, assessing time-extrapolation performance. Models were trained on data from $t=0$ to $t=1.6$s, and these figures show the prediction at the final test time of $t=2.0$s. For each PDE, all sub-images corresponding to different time integration schemes share a consistent color bar range for a fair comparison.}
    \label{fig:pde_final_snapshots_combined}
\end{figure*}

To provide a qualitative assessment, we visualize the final predicted snapshot of the test phase (time-extrapolation) for representative samples of the 2D advection, heat, and Burgers' equations in Figures \ref{fig:pde_final_snapshot_advection}, \ref{fig:pde_final_snapshot_heat}, and \ref{fig:pde_final_snapshot_burgers}, respectively. These qualitative results align directly with the quantitative findings in Table \ref{tab:pde_time_mse_comparison}. For the purely convective advection case, the forward Euler scheme yields the best prediction, visually confirming that simpler integration methods are effective for linear systems. However, for the more complex heat and Burgers' equations, the prediction from the Adams-Euler scheme most closely resembles the ground truth, successfully capturing the primary structures of the final solution field where other methods show more noticeable diffusion and error. This visual evidence further underscores the superior stability and accuracy of the Adams-Euler method for time-extrapolative forecasting tasks across systems with varying physical complexity.

\subsection{Application of proposed adaptive multi-step rollout}
\label{sec:pde_results_table3}

Building on the single-step results, we now investigate the impact of multi-step rollouts ($M=4$) combined with the different time integration schemes. This experiment directly evaluates the effectiveness of the fixed-weight vanilla approach against our proposed adaptive weighting strategies (AW1, AW2, and AW3). The results are presented in Table \ref{tab:pde_aw_comparison}.

\begin{table}[H]
	\centering
	\caption{Comparison of test MSE values in single-step and multi-step ($M=4$) rollout. For each time integration scheme, the best performing strategy is highlighted in bold. Test MSE values are scaled by $10^6$.}
	\label{tab:pde_aw_comparison}
	\begin{tabular}{ll|r|rrrr}
        & & \textbf{Single-step}&\multicolumn{4}{c}{\textbf{Multi-Step Rollout ($M$=4)}}\\
		\hline
		\textbf{PDE} & \textbf{Time Integration Scheme} & \multicolumn{1}{c|}{w/o AW} & w/o AW & AW1 & AW2 & AW3 \\
		\hline
		\multirow{4}{*}{2D Advection}
            & Direct prediction & \textbf{7723} & 11219 & 9110 & 8659 & 11851 \\
            & Forward Euler & 5143 & \textbf{4812} & 6125 & 5035 & 5845 \\
            & Second-order central   & \textbf{8852} & 17111 & 51953 & 19988 & 17903 \\
            & Adams-Euler & 7273 & 7482 & 9874 & \textbf{6337} & 6448 \\
        \hline
        \multirow{4}{*}{2D Heat}
            & Direct prediction & 76 & 123 & 59 & 72 & \textbf{58} \\
            & Forward Euler & 21 & \textbf{18} & 14 & 62 & 78 \\
            & Second-order central   & 24 & \textbf{16} & 27 & 65 & 80 \\
            & Adams-Euler & \textbf{16} & 19 & 20 & 62 & 64 \\
        \hline
        \multirow{4}{*}{2D Burgers}
            & Direct prediction & 357 & 238 & \textbf{196} & 218 & 316 \\ 
            & Forward Euler & 112 & \textbf{63} & 77 & 95 & 103 \\
            & Second-order central & 112 & 98 & \textbf{83} & 109 & 107 \\
            & Adams-Euler & 95 & 137 & \textbf{66} & 97 & 96 \\
		\hline
	\end{tabular}
\end{table}

The results reveal a clear progression: as the underlying physics increases in complexity, more sophisticated multi-step rollout techniques become advantageous. For the purely convective advection equation, single-step rollout generally performs best, indicating that simple transport dynamics require minimal regularization. The diffusive heat equation benefits most from vanilla multi-step rollout without adaptive weighting, suggesting that moderate regularization suffices for systems with smoothing dynamics. For the nonlinear Burgers' equation, which combines convection and diffusion, adaptive weighting strategies (particularly AW1) achieve optimal performance, demonstrating that complex, coupled physics require sophisticated training approaches to manage error accumulation effectively. This progression underscores that the choice of training strategy should align with the physical complexity of the system being modeled. Following this trend, one can anticipate that even more advanced adaptive strategies (AW2 and AW3) will prove most effective for highly complex systems such as the Navier-Stokes cylinder flow dynamics, as demonstrated in subsequent sections. With respect to time integration schemes, both forward Euler and Adams-Euler demonstrate consistent robustness across all three PDE systems, maintaining competitive performance regardless of the underlying physics or training strategy employed.

\section{Experiments for Navier-Stokes Cylinder Flow Dynamics}
\label{sec:NS_all}

\subsection{Experimental setup}
\label{sec:Exp}

In this section, we transition from the canonical PDEs to a more challenging and realistic test case: the vortex shedding phenomenon behind a two-dimensional (2D) circular cylinder. We specifically evaluate the performance of our framework in a mesh-agnostic SciML paradigm using a Graph U-Net model \cite{gao2019graph, yang2024enhancing}. The goal is to assess whether key insights gained from the simpler systems---that the Adams-Euler scheme remains robust and that advanced adaptive weighting becomes increasingly necessary for more complex physics---hold true for unsteady complex fluid dynamics.

\subsubsection{Dataset and preprocessing}

\begin{figure}[htb!] 
\centering 
\includegraphics[width=.5\textwidth]{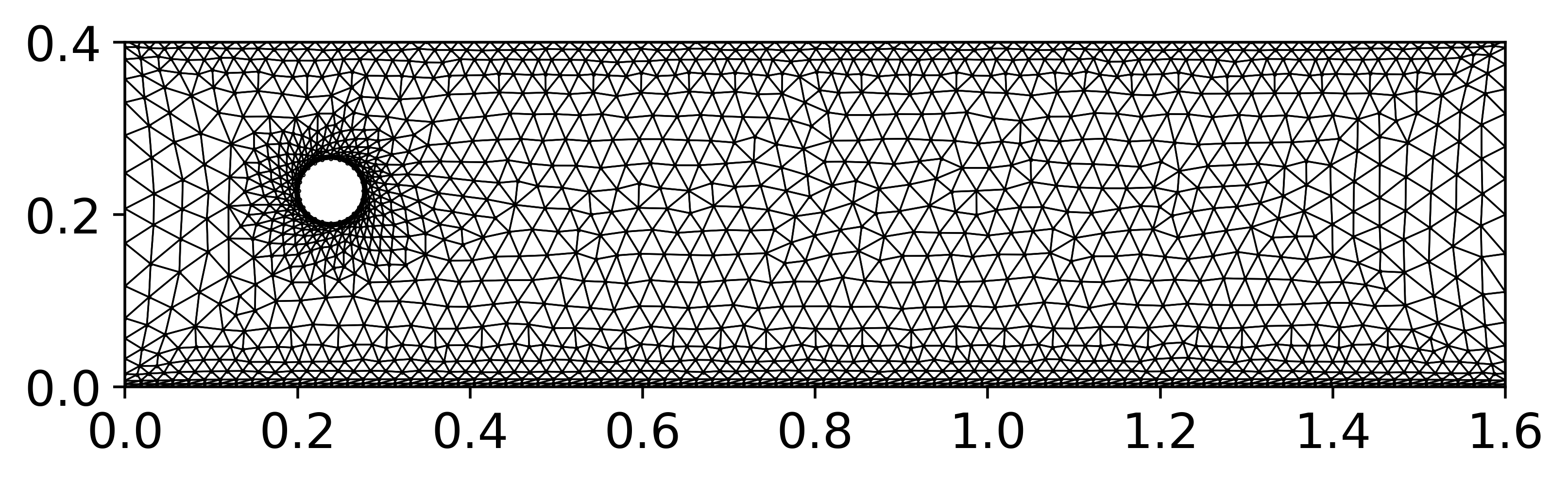} 
\caption{Mesh used for training, containing 1,946 nodes, 11,208 edges, and 3,658 volume cells. The flow is from left to right.} \label{fig:mesh} 
\end{figure}

The dataset features a mesh scenario with a cylinder placed in a fluid flow, generating vortex shedding characterized by oscillatory flow patterns. The training mesh consists of 1,946 nodes, 11,208 edges, and 3,658 volume cells, as shown in Figure \ref{fig:mesh}: this training dataset is adapted from the work of Google DeepMind \cite{pfaff2020learning}. The flow conditions include a maximum inlet velocity of $1.78$ m/s with a parabolic velocity distribution and a cylinder diameter of $0.074$ m. A single vortex shedding period comprises approximately 29 snapshots (or trajectories), where the time step between each is fixed at $\Delta t$=0.01 second.

For training, we use only 50 consecutive $x$-velocity snapshots. Our investigation focuses exclusively on the $x$-velocity, as this single component provides a sufficient and challenging testbed for our primary contribution: a model-agnostic temporal prediction framework. The $x$-velocity field encapsulates the most dominant physical features of the flow, including the characteristic von Kármán vortex shedding. By demonstrating our framework's ability to accurately forecast this dynamically rich variable, we focus on a validation of the proposed temporal methods themselves. The model input consists of a sequence of $N=20$ past snapshots, and its task is to predict the next immediate snapshot.

\subsubsection{Model training details}

\begin{figure}[htb!] 
\centering 
\includegraphics[width=.9\textwidth]{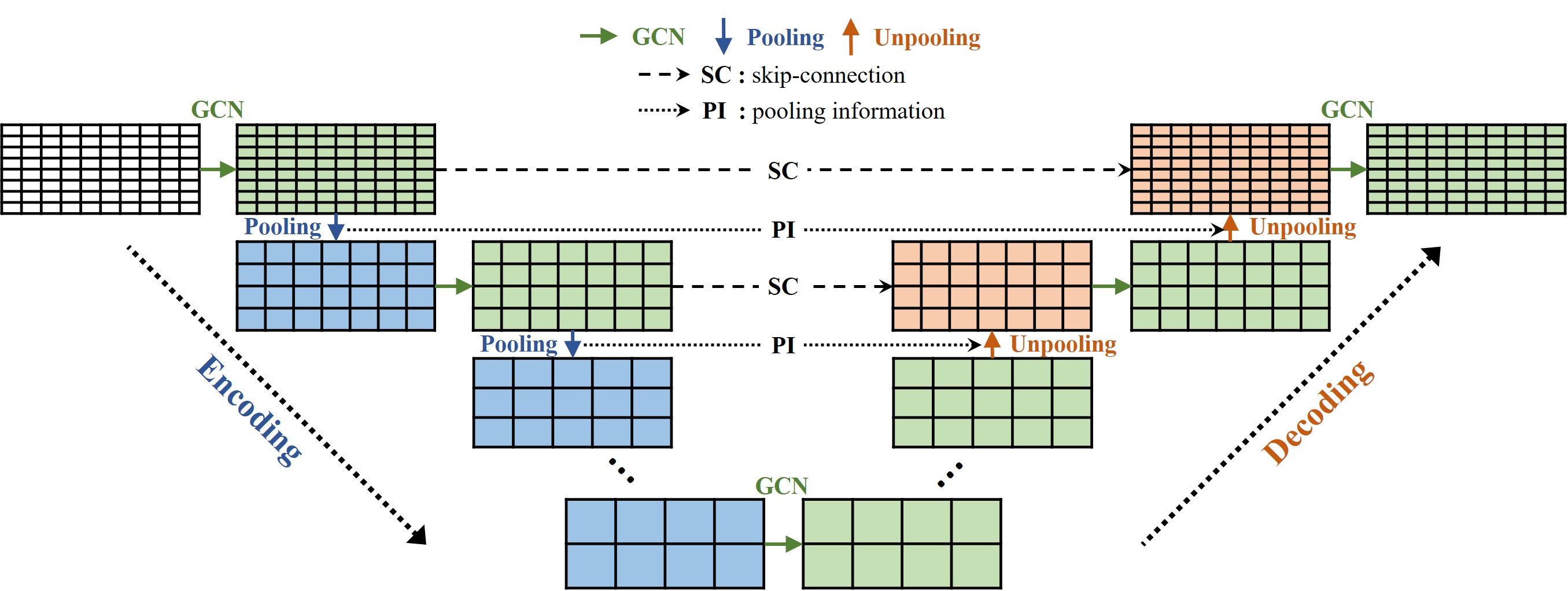} 
\caption{The Graph U-Net architecture consists of an encoder and a decoder, with skip connections facilitating the transfer of information from the encoder to the decoder.} \label{fig:GUNet} 
\end{figure}

The Graph U-Net architecture employed in this study is designed with a focus on computational efficiency and simplicity, comprising only 1,177 trainable parameters (Figure \ref{fig:GUNet}). The encoder part of the model consists of four graph convolutional network (GCN) layers \cite{ogoke2021graph}, with channel dimensions decreasing from 20 to 1 in the sequence 20, 15, 10, 5, and 1, where the initial dimension corresponds to the 20-channel input graph. A pooling ratio of 0.6 is applied using the gPool layer \cite{gao2019graph} after each GCN layer in the encoder, as this ratio was found to be optimal in our previous work \cite{yang2024enhancing}. This pooling operation reduces the number of nodes by 60\% at each step, effectively capturing multi-scale features while maintaining computational tractability.

The decoder mirrors the encoder with four GCN layers, all having a channel dimension of 1. Skip-connections between corresponding layers of the encoder and decoder are incorporated to facilitate the flow of information and improve reconstruction accuracy. The unpooling operations restore the graph to its original size, ensuring that the output mesh has the same dimensionality as the input. The model is trained for 5,000 epochs using the Adam optimizer with an initial learning rate of $10^{-3}$. To ensure robustness and capture variability, each training process is repeated three times under identical settings, with the mean performance metrics presented as experimental results. During training, the MSE between predicted and ground-truth snapshots serves as the loss function.

In summary, the Graph U-Net model used in this section is evaluated under intentionally harsh conditions. We use a strict chronological split for our data: the model is trained on a minimal dataset of only the first 50 snapshots and then tested on its ability to forecast a long prediction horizon of the subsequent 350 snapshots. This setup, combined with the model's limited capacity (1,177 parameters), creates a challenging 7:1 prediction-to-training ratio to rigorously assess its time-extrapolation capabilities. These constraints simulate practical engineering scenarios where only small datasets are available and computational infrastructure for model training is limited. This challenging setup rigorously tests the model's robustness and its ability to manage error accumulation over time, particularly for complex temporal patterns like vortex shedding.

\subsection{Application of time integration schemes into auto-regressive GNNs}
\label{sec:OnlyTime}

We present results of applying the four different time integration schemes, along with conventional direct prediction, as described in Section \ref{sec:method_time}. The objective is to assess their impact on long-term prediction accuracy and stability of the Graph U-Net model before incorporating multi-step rollout techniques.

\subsubsection{Predictive performance comparison}\label{sec:OnlyTime_1}

We evaluate each time integration scheme by computing MSE between predicted and ground-truth flow fields over the entire spatial domain. MSE values are averaged over 350 future snapshots using Eq. \ref{eq:MSE}, providing comprehensive assessment of long-term prediction capability. Results are summarized in Table \ref{tab:time_mse_comparison}.

\begin{table}[htb!]
    \centering
    \caption{Comparison of MSE for different time integration schemes over 350 future snapshots.}
    \label{tab:time_mse_comparison}
    \begin{tabular}{lcc}
        \hline
        \textbf{Time Integration Scheme} & \textbf{MSE} & \textbf{Training time [s]}  \\
        \hline
        Direct prediction & \textbf{0.125} & 2004\\
        Forward Euler & \textbf{0.138} & 1867 \\
        Second-order central & 65.024 & 1879 \\
        Adams-Euler & \textbf{0.139} & 1895 \\
        \hline
    \end{tabular}
\end{table}

As observed in Table \ref{tab:time_mse_comparison}, direct prediction, forward Euler, and Adams-Euler schemes yield relatively low MSE values with better prediction accuracy. In contrast, second-order central difference exhibits significantly higher errors. Training times are comparable across all schemes, indicating that different time integration methods do not significantly affect computational requirements.

\begin{figure*}[htb!]
    \begin{subfigure}[h]{0.45\textwidth}
        \centering
        \includegraphics[width=\textwidth]{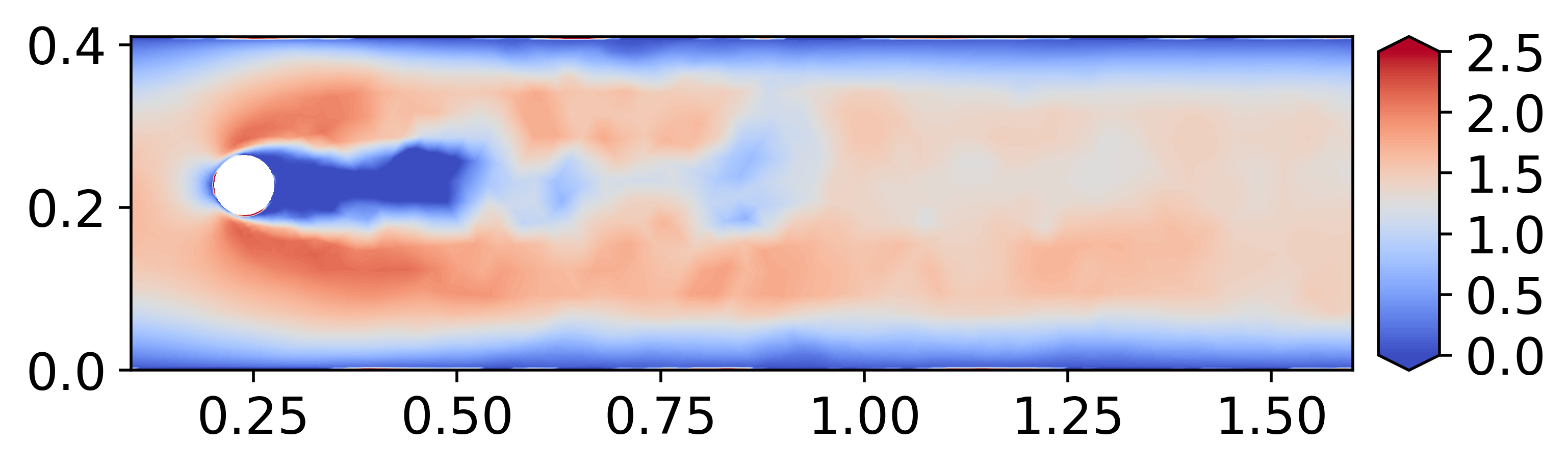}
        \caption{After 100 rollout steps}
        \label{fig:1st_central_1}
    \end{subfigure}
    \hfill
    \begin{subfigure}[h]{0.45\textwidth}
        \centering
        \includegraphics[width=\textwidth]{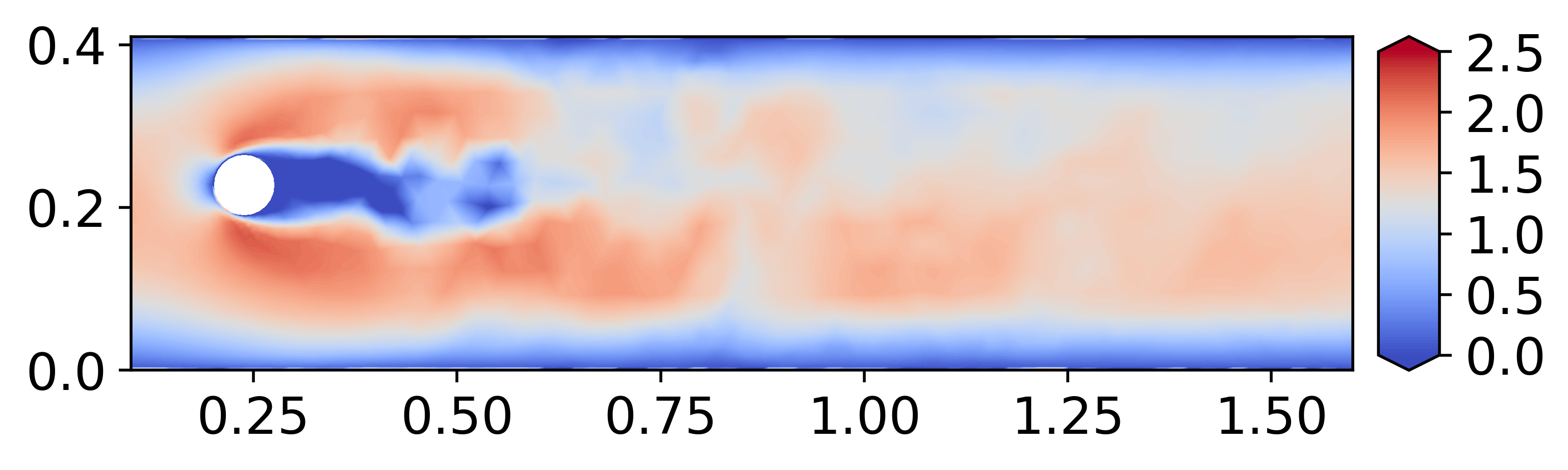}
        \caption{After 101 rollout steps}
        \label{fig:1st_central_2}
    \end{subfigure}
    
    \vfill
    
    \begin{subfigure}[h]{0.45\textwidth}
        \centering
        \includegraphics[width=\textwidth]{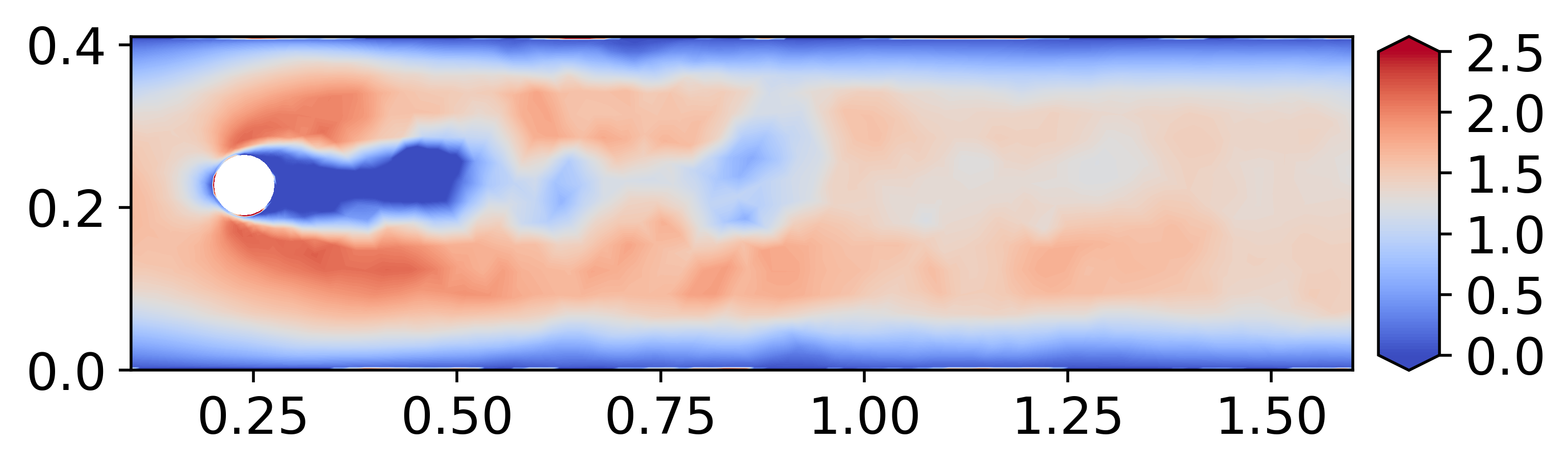}
        \caption{After 102 rollout steps}
        \label{fig:1st_central_3}
    \end{subfigure}
    \hfill
    \begin{subfigure}[h]{0.45\textwidth}
        \centering
        \includegraphics[width=\textwidth]{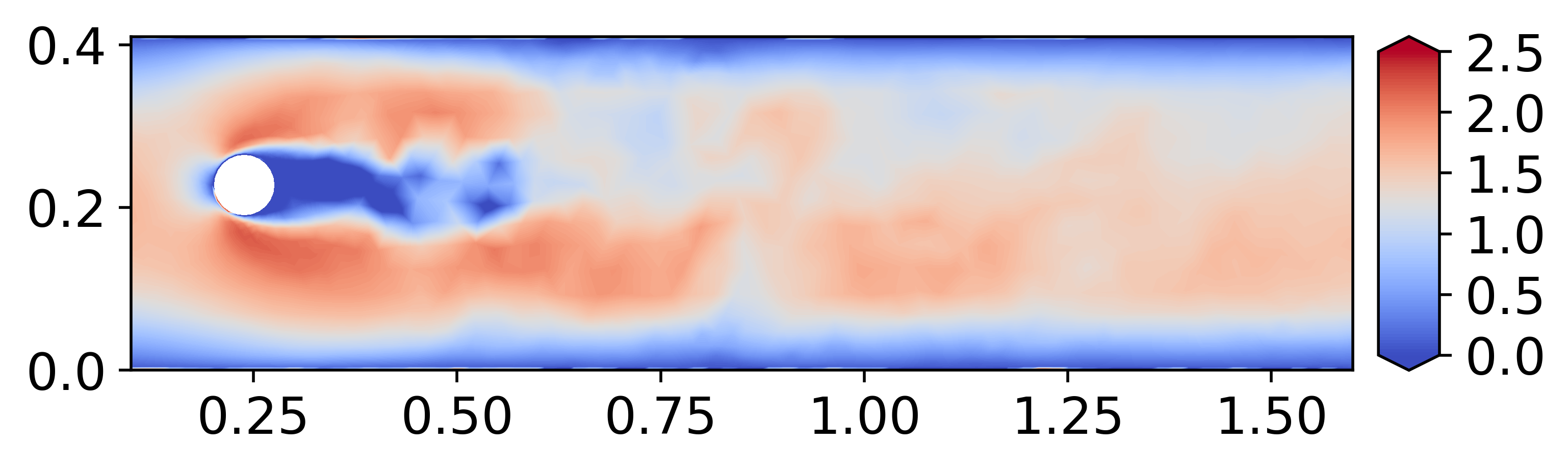}
        \caption{After 103 rollout steps}
        \label{fig:1st_central_4}
    \end{subfigure}
    
    \caption{Second-order central difference scheme predictions showing oscillatory instability. Snapshots after 100 and 102 rollouts display similar patterns, while those after 101 and 103 show alternating patterns.}
    \label{fig:1st_central}
\end{figure*}

Among the various approaches, Figure \ref{fig:1st_central} illustrates the fundamental instability of the second-order central difference scheme. Four consecutive snapshots after 100 rollouts reveal highly oscillatory behavior: snapshots after 100 (Figure \ref{fig:1st_central_1}) and 102 (Figure \ref{fig:1st_central_3}) rollouts appear similar, while those after 101 (Figure \ref{fig:1st_central_2}) and 103 (Figure \ref{fig:1st_central_4}) exhibit similar patterns. This alternating pattern persists beyond 100 rollouts, reflecting the scheme's fundamental limitation where $\mathbf{u}(t+\Delta t)$ depends heavily on $\mathbf{u}(t-\Delta t)$ while bypassing the intermediate state $\mathbf{u}(t)$; since $\mathbf{u}(t+\Delta t) = \mathbf{u}(t-\Delta t) + 2\Delta t \cdot \frac{\delta \mathbf{u}}{\delta t}$. These results align with established findings that central difference schemes are generally unsuitable for time integration due to inherent instability.

\begin{figure*}[htb!]
    \centering
    \begin{subfigure}[h]{0.45\textwidth}
        \centering
        \includegraphics[width=\textwidth]{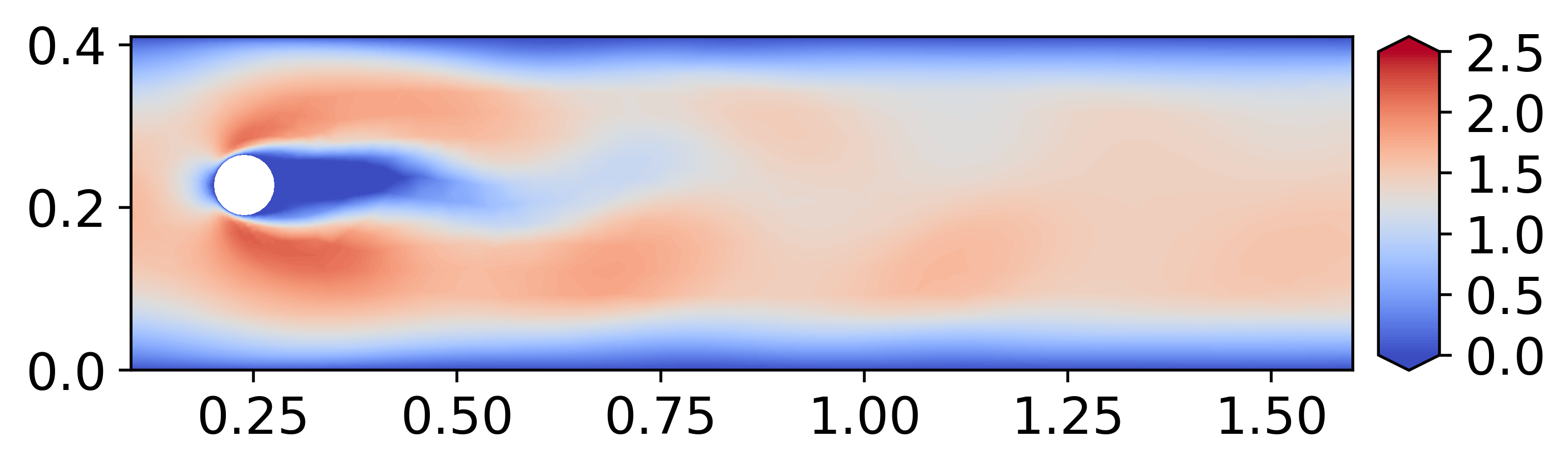}
        \caption{Ground truth flow field at snapshot $t+100$}
    \end{subfigure}
    
    \vfill
    
    \begin{subfigure}[h]{0.45\textwidth}
        \centering
        \includegraphics[width=\textwidth]{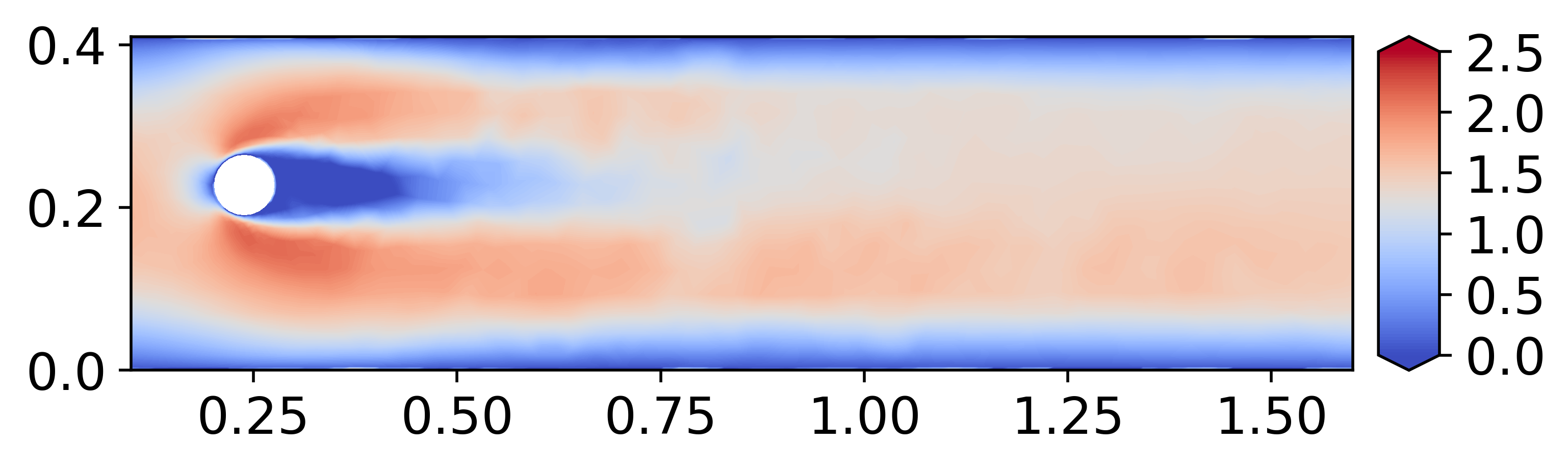}
        \caption{Direct prediction}
    \end{subfigure}
    \hfill
    \begin{subfigure}[h]{0.45\textwidth}
        \centering
        \includegraphics[width=\textwidth]{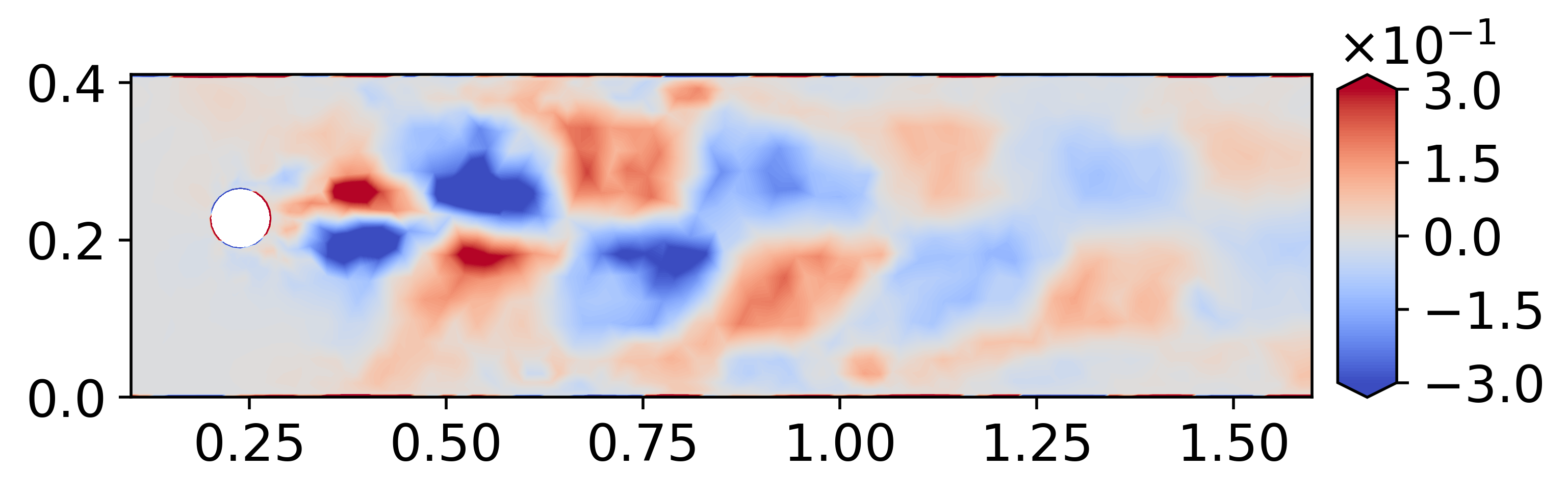}
        \caption{Direct prediction error}
    \end{subfigure}
    
    \vfill
    
    \begin{subfigure}[h]{0.45\textwidth}
        \centering
        \includegraphics[width=\textwidth]{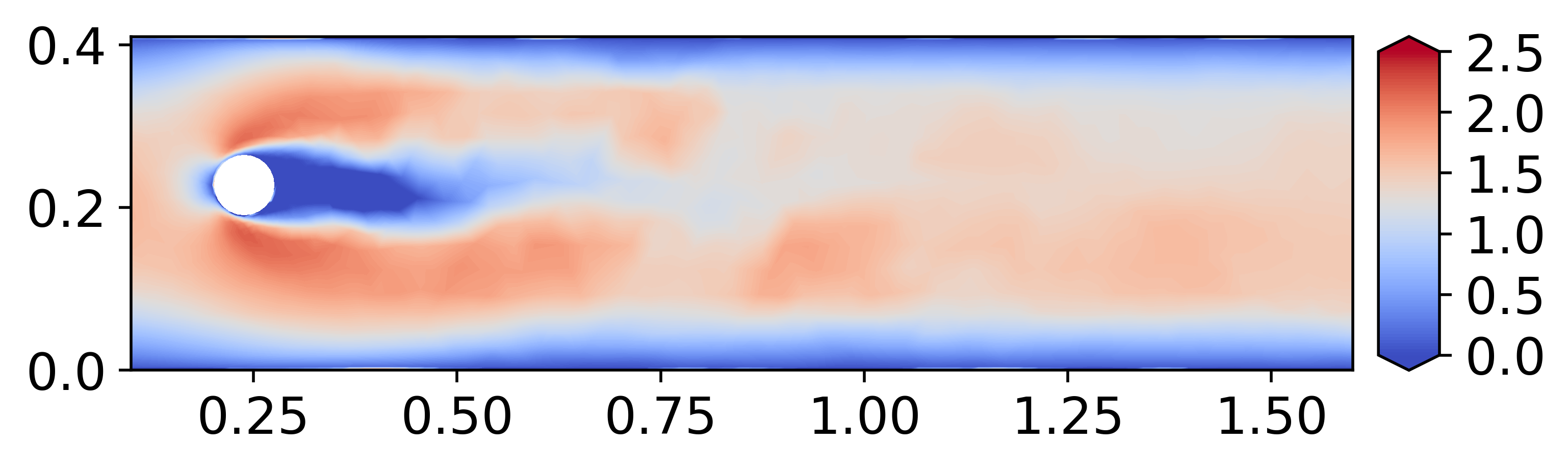}
        \caption{Forward Euler prediction}
    \end{subfigure}
    \hfill
    \begin{subfigure}[h]{0.45\textwidth}
        \centering
        \includegraphics[width=\textwidth]{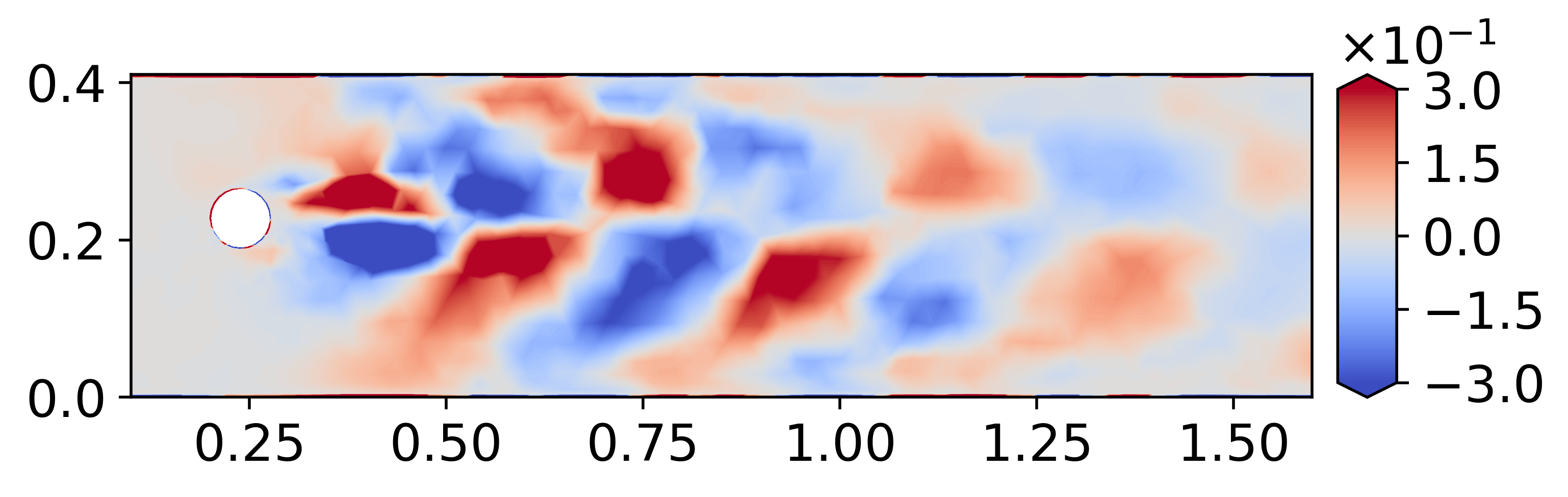}
        \caption{Forward Euler error}
    \end{subfigure}
    
    \vfill
    
    \begin{subfigure}[h]{0.45\textwidth}
        \centering
        \includegraphics[width=\textwidth]{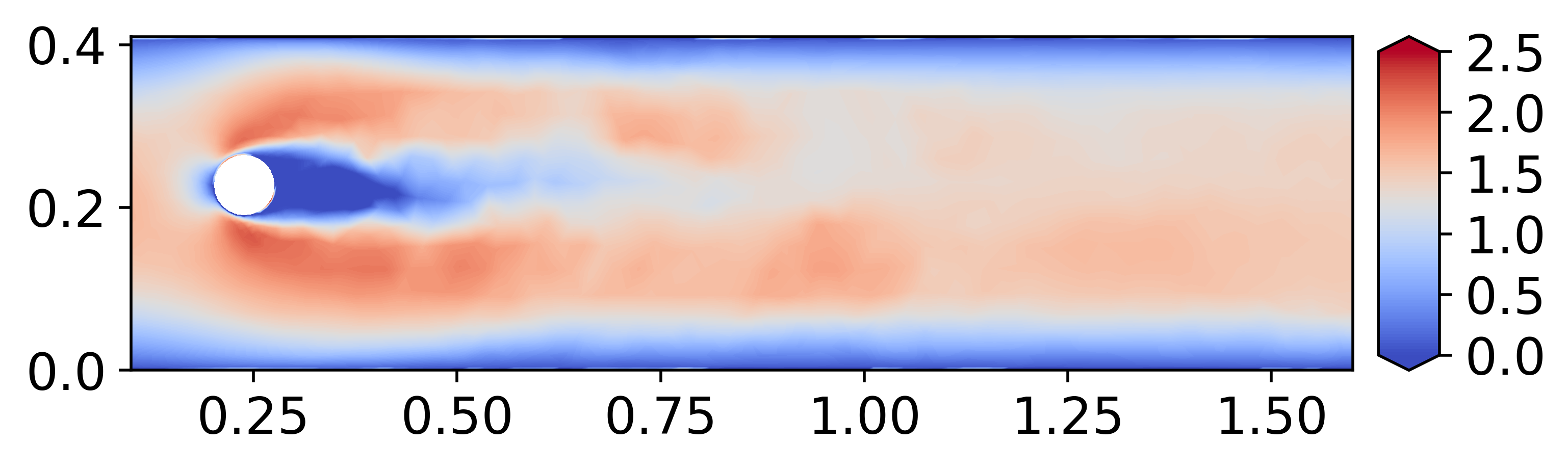}
        \caption{Adams-Euler prediction}
    \end{subfigure}
    \hfill
    \begin{subfigure}[h]{0.45\textwidth}
        \centering
        \includegraphics[width=\textwidth]{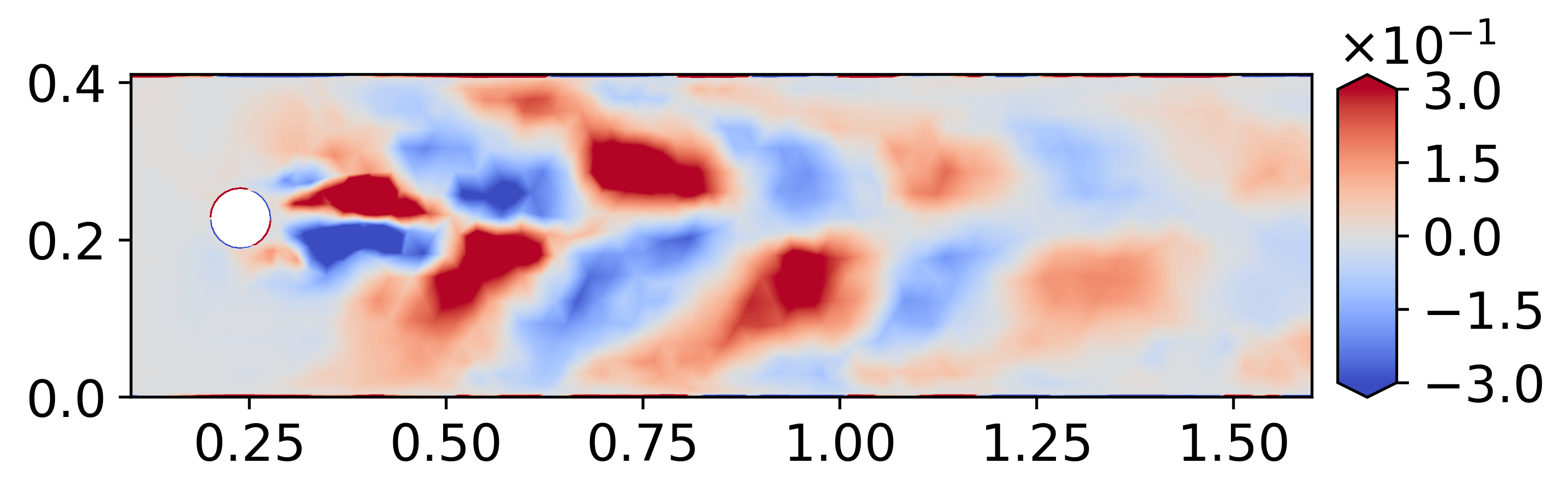}
        \caption{Adams-Euler error}
    \end{subfigure}
    
    \caption{$x$-velocity field predictions and error distributions after 100 rollout steps, showing limitations of time integration methods alone.}
    \label{fig:flow_field_visualization}
\end{figure*}

To further illustrate performance limitations, Figure \ref{fig:flow_field_visualization} presents flow field predictions after 100 rollout steps for the three outperforming schemes alongside ground truth. The ground truth shows a specific vortex shedding phase behind the cylinder. All methods show noticeable discrepancies from ground truth after only 100 rollout steps. Direct prediction provides time-averaged results where flow fields remain constant during rollout progression. The corresponding error plots highlight significant prediction deviations, especially in the wake region where complex flow dynamics are most pronounced.

\subsubsection{Need for additional techniques}

Despite relatively low MSE values for some schemes, the Graph U-Net models struggle to maintain accurate long-term predictions over extended rollout horizons. Error accumulation becomes apparent within the first 100 rollout steps, well short of our 350-step prediction objective. These findings underscore the limitations of directly applying traditional finite difference schemes within AR architectures, revealing insufficient performance for extended prediction horizons. This motivates the need for enhanced techniques to improve stability and accuracy. In subsequent sections, we integrate multi-step rollout strategies with these time integration schemes to enhance long-term prediction performance and achieve more reliable forecasts over extended time horizons.

\subsection{Extension of time integration schemes into conventional multi-step rollout scenario}
\label{sec:extend_multistep}

We extend the previously investigated time integration schemes by incorporating conventional multi-step rollout techniques with different values of $M$, where $M$ represents the number of future snapshots considered during training. Specifically, we examine $M = 1$, $2$, $4$, and $8$. When $M=1$, multi-step rollout is not applied, corresponding to the vanilla AR model. We adopt the conventional weighting strategy suggested by \citet{wu2022learning}, where loss weights are set to $w_1=1$ for the first future step and $w_i=0.1$ for subsequent steps (see Eq. \ref{eq:multi-loss}).

We evaluate all time integration schemes with varying values of $M$, with MSE results presented in Table \ref{tab:mse_comparison_multistep}. Increasing multi-step rollout length $M$ introduces significant computational overhead, as evidenced by training times increasing from 1,914s ($M=1$) to 3,131s ($M=8$). However, this increased computational cost does not necessarily improve performance. The results reveal a clear distinction between time integration schemes: while Adams-Euler demonstrates consistently robust performance across all $M$ values (MSE improving from 0.139 at $M=1$ to 0.070 at $M=4$), other schemes exhibit significant instability. Direct prediction breaks down completely with NaN errors at $M=4$ and $M=8$, while forward Euler shows instability at $M=2$ (MSE of 581.945) and $M=8$ (NaN). Second-order central difference displays extremely high MSE values across increasing $M$.

\begin{table}[htb!]
    \centering
    \caption{MSE comparison for different time integration schemes with varying multi-step rollout lengths $M$: erroneous values are shown with a gray background. Strouhal numbers ($St$) are also provided under each MSE value, where ground truth $St$ value is 0.1438.}
    \label{tab:mse_comparison_multistep}
    \renewcommand{\arraystretch}{1.25}
    \begin{tabular}{>{\raggedright\arraybackslash}m{5.cm}*{4}{>{\centering\arraybackslash}m{2.3cm}}}
        \hline
        \multirow{2}{*}{\textbf{Time Integration Scheme}} & \multicolumn{4}{c}{\textbf{Number of considered future snapshots}} \\
        & \textbf{$M=1$} & \textbf{$M=2$} & \textbf{$M=4$} & \textbf{$M=8$} \\
        \hline
        \multirow{2}{*}{Direct prediction}
          & 0.125 & 0.102 & \cellcolor{gray!30}NaN & \cellcolor{gray!30}NaN \\
          & ($St=0.3261$) & ($St=0.1434$) & \cellcolor{gray!30}($St=0.3566$) & \cellcolor{gray!30}($St=0.3248$) \\
        \multirow{2}{*}{Forward Euler}
          & 0.138 & \cellcolor{gray!30}581.945 & 0.075 & \cellcolor{gray!30}NaN \\
          & ($St=0.1353$) & \cellcolor{gray!30}($St=0.0101$) & ($St=0.1434$) & \cellcolor{gray!30}($St=0.0115$) \\
        \multirow{2}{*}{Second-order central difference}
          & \cellcolor{gray!30}65.024 & \cellcolor{gray!30}1248.223 & \cellcolor{gray!30}138.025 & 0.475 \\
          & \cellcolor{gray!30}($St=0.0101$) & \cellcolor{gray!30}($St=0.0101$) & \cellcolor{gray!30}($St=0.0101$) & ($St=0.0095$) \\
        \multirow{2}{*}{Adams-Euler}
          & 0.139 & 0.092 & \textbf{0.070} & 0.071 \\
          & ($St=0.1319$) & ($St=0.1387$) & ($St=0.1367$) & ($St=0.1455$) \\
        \hline
        \textbf{Averaged time [s]} & 1914 & 2086 & 2633 & 3131 \\
        \hline
    \end{tabular}
\end{table}

These results reveal two key insights: First, naively increasing $M$ imposes excessive training constraints on the lightweight model (1,177 parameters), often leading to performance degradation despite the theoretical benefit of learning longer-term dependencies. Second, the time integration scheme proves more critical than the choice of $M$---while most methods fail even with optimal $M$ values, Adams-Euler maintains robust performance across all $M$ values, demonstrating inherent stability for long-term AR prediction.

The superior performance of Adams-Euler can be attributed to its ability to leverage information from multiple previous time steps in a mathematically principled way. Since three successive snapshots, $\mathbf{u}(t-\Delta t)$, $\mathbf{u}(t)$, and $\mathbf{u}(t+\Delta t)$, are explicitly considered for predicting the next snapshot (Eq. \ref{eq:time4_0}, \ref{eq:time4_1}, \ref{eq:time4_2}), this scheme enables stable predictions even with limited model capacity.

\begin{figure*}[htb!]
    \centering
    \begin{subfigure}[h]{0.48\textwidth}
        \centering
        \includegraphics[width=\textwidth]{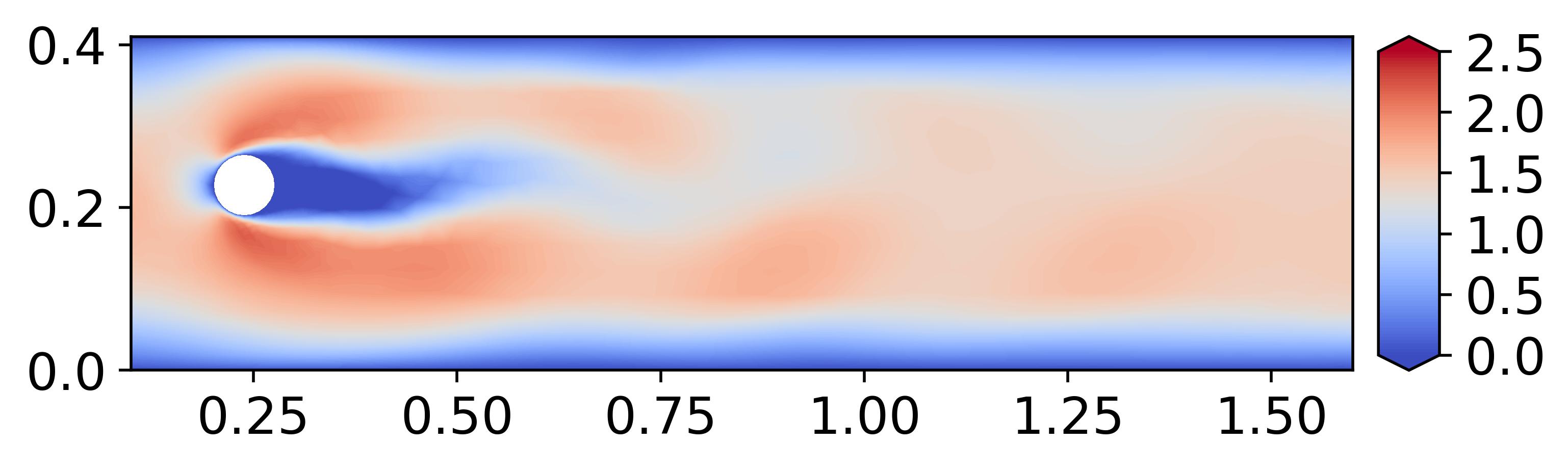}
        \caption{Ground truth at snapshot $t+200$}
        \label{fig:velocity_field_comparison_a}
    \end{subfigure}
    \hfill
    \begin{subfigure}[h]{0.48\textwidth}
        \centering
        \includegraphics[width=\textwidth]{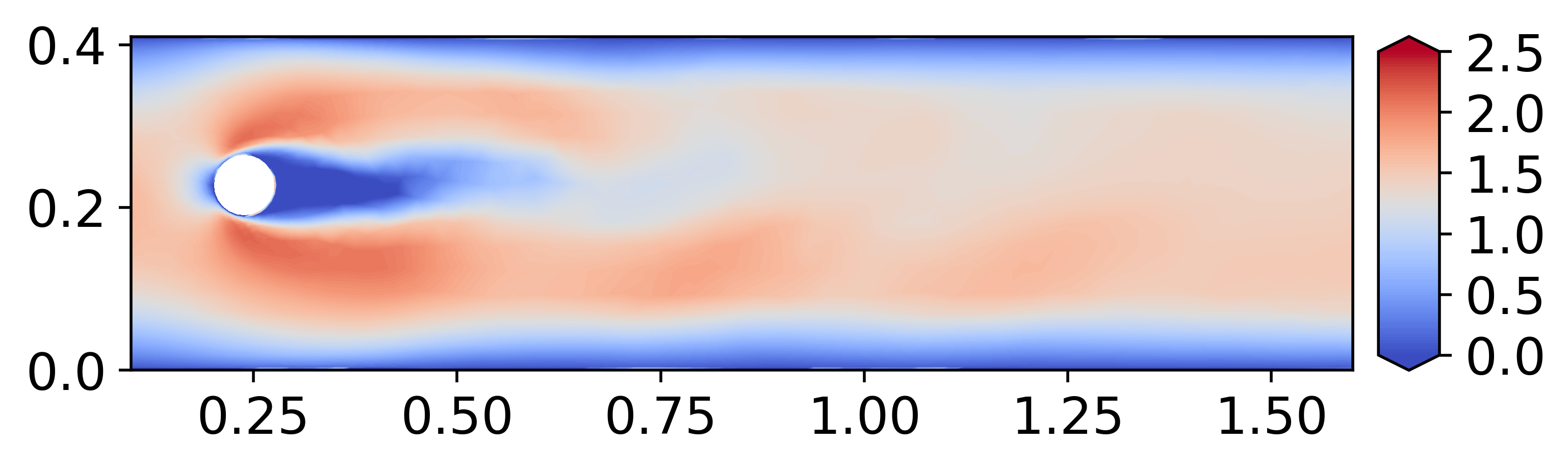}
        \caption{Adams-Euler prediction at $t+200$ (MSE=0.091)}
        \label{fig:velocity_field_comparison_b}
    \end{subfigure}
    
    \vfill
    
    \begin{subfigure}[h]{0.48\textwidth}
        \centering
        \includegraphics[width=\textwidth]{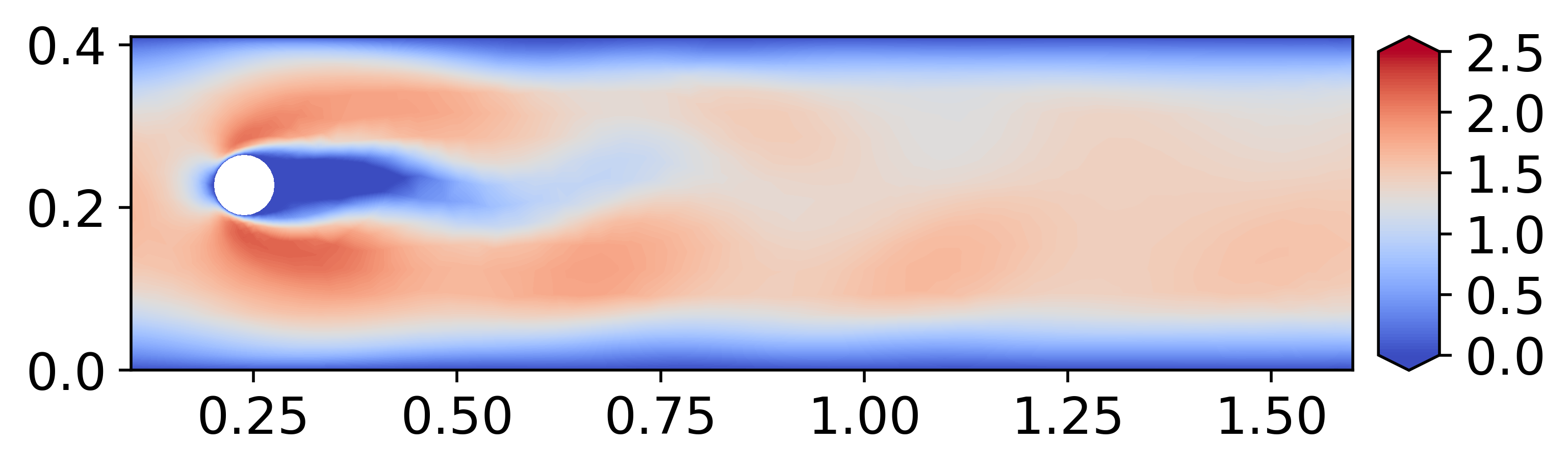}
        \caption{Ground truth at snapshot $t+300$}
        \label{fig:velocity_field_comparison_c}
    \end{subfigure}
    \hfill
    \begin{subfigure}[h]{0.48\textwidth}
        \centering
        \includegraphics[width=\textwidth]{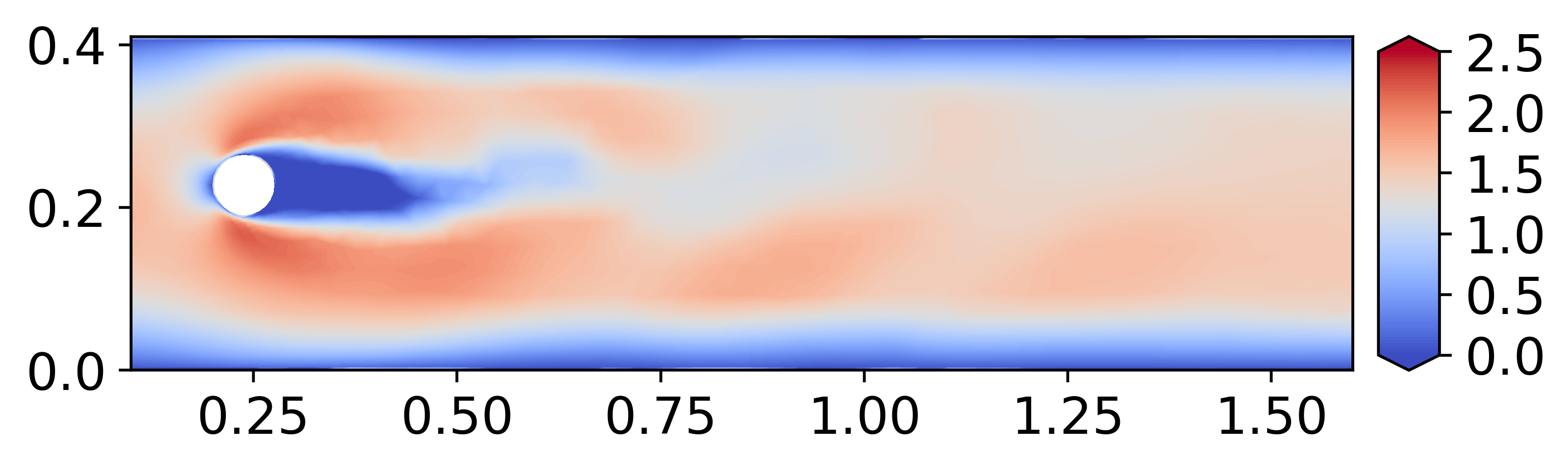}
        \caption{Adams-Euler prediction at $t+300$ (MSE=0.105)}
        \label{fig:velocity_field_comparison_d}
    \end{subfigure}
    
    \caption{$x$-velocity field predictions using Adams-Euler scheme with $M=4$ multi-step rollout, showing improved performance until $t+200$.}
    \label{fig:velocity_field_comparison}
\end{figure*}

To illustrate the improved performance achieved with Adams-Euler and $M=4$ multi-step rollout (best case in Table \ref{tab:mse_comparison_multistep}), Figure \ref{fig:velocity_field_comparison} shows flow field predictions after 200 and 300 rollout steps. At snapshot $t+200$ (Figures \ref{fig:velocity_field_comparison_a} and \ref{fig:velocity_field_comparison_b}), the prediction shows reasonable agreement with ground truth, capturing overall flow patterns and vortex shedding behavior. This represents substantial improvement over Figure \ref{fig:flow_field_visualization}, where accurate predictions were unattainable even at 100 rollout steps, clearly demonstrating the effectiveness of combining conventional multi-step rollout with the Adams-Euler scheme. However, at snapshot $t+300$ (Figures \ref{fig:velocity_field_comparison_c} and \ref{fig:velocity_field_comparison_d}), MSE increases from 0.091 to 0.105, showing noticeable divergence from ground truth: flow structure accuracy diminishes with different vortex shedding patterns. While the incorporation of multi-step rollout with Adams-Euler significantly enhances long-term prediction performance, room for improvement remains, particularly for prediction horizons extending beyond $t+200$. This motivates the development of adaptive multi-step rollout strategies explored in subsequent sections.

\subsection{Application of proposed adaptive multi-step rollout} \label{sec:AWmulti}

We evaluate the three adaptive multi-step rollout approaches elaborated in Section \ref{sec:method_multi}, aiming to enhance robustness and accuracy of long-term predictions by automatically adjusting loss function weights during training.

\begin{figure}[htb!]
    \centering
    \includegraphics[width=0.6\textwidth]{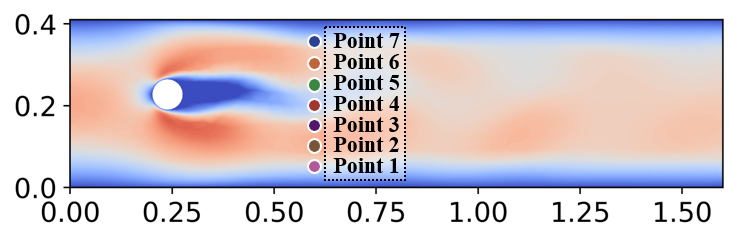}
    \caption{Seven probe points for quantitative assessment of vortex shedding prediction performance.}
    \label{fig:probe_points_locations}
\end{figure}

Performance of the three adaptive weighting strategies across different time integration schemes is assessed using a focused evaluation approach. To better evaluate the models' ability to capture complex wake oscillatory vortex behavior, MSE results from this section are calculated based on $x$-velocity data from seven strategic probe points in the wake region (Figure \ref{fig:probe_points_locations}), rather than across the entire field. This targeted analysis provides precise assessment of how well each method captures critical vortex shedding dynamics.

Results are summarized in Table \ref{tab:mse_comparison_adaptive_weighting}. For direct prediction, the vanilla approach with fixed weights (MSE: 0.011) outperforms adaptive variants, likely because the direct prediction's inherent simplicity does not benefit from complex weighting schemes. However, for derivative-based time integration methods, adaptive approaches show clear advantages. Both AW2 and AW3 demonstrate strong performance with forward Euler (MSE: 0.007 and 0.010 respectively), indicating that adaptive weighting helps balance short-term and long-term prediction accuracy more effectively than fixed weights. The most remarkable performance is achieved by combining Adams-Euler with AW3 (MSE: 0.002), which leverages both the superior temporal integration of Adams-Euler and the focused weighting strategy of AW3. This represents a significant 89\% reduction in mean squared error compared to the conventional fixed-weight multi-step rollout (which had an MSE of 0.018), highlighting the practical benefit of our adaptive approach.

\begin{table}[htb!]
    \centering
    \begin{threeparttable}
        \caption{MSE comparison for different adaptive weighting approaches across time integration schemes with $M=4$ multi-step rollout. MSE values calculated from seven probe points in Figure 9. Strouhal numbers ($St$) are also provided under each MSE value, where ground truth $St$ value is 0.1438.}
        \label{tab:mse_comparison_adaptive_weighting}
        \renewcommand{\arraystretch}{1.25}
        \begin{tabular}{lcccc}
            \hline
            \multirow{2}{*}{\textbf{Time Scheme}} & \textbf{Vanilla} & \textbf{AW1} & \textbf{AW2} & \textbf{AW3} \\
            & (fixed weights) & (without learnable $k$) & (with learnable $k$) & (first and last only) \\
            \hline
            \multirow{2}{*}{Direct prediction}
                & \textbf{0.011} & 0.025 & 0.027 & 0.027 \\
                & ($St=0.1407$) & ($St=0.0365$) & ($St=0.0074$) & ($St=0.0453$) \\
            \multirow{2}{*}{Forward Euler}
                & 0.019 & 0.023 & \textbf{0.007}\tnote{*} & \textbf{0.010} \\
                & ($St=0.1374$) & ($St=0.1421$) & ($St=0.1489$) & ($St=0.1407$) \\
            \multirow{2}{*}{Adams-Euler}
                & 0.018 & 0.017 & \textbf{0.010} & \textbf{0.002}\tnote{**} \\
                & ($St=0.1380$) & ($St=0.1428$) &($St=0.1489$) & ($St=0.1434$) \\
            \hline
            \textbf{Averaged time [s]} & 2420 & 2384 & 2408 & 2354 \\
            \hline
        \end{tabular}
        \begin{tablenotes}\footnotesize\raggedleft
            \item[*] Model A
            \item[**] Model B
        \end{tablenotes}
    \end{threeparttable}
\end{table}

The superiority of adaptive weighting methods can be attributed to their ability to dynamically allocate training focus based on prediction difficulty. AW1 provides automatic error-based weighting, naturally emphasizing challenging time steps. AW2 extends this with learnable flexibility, allowing the model to discover optimal weighting patterns through training. AW3 achieves the best performance by recognizing that intermediate steps often contain redundant information—focusing only on immediate accuracy (first step) and long-term stability (last step) creates a more efficient learning signal that avoids overfitting to intermediate predictions.

\begin{figure}[htb!]
    \centering
    \includegraphics[width=0.95\textwidth]{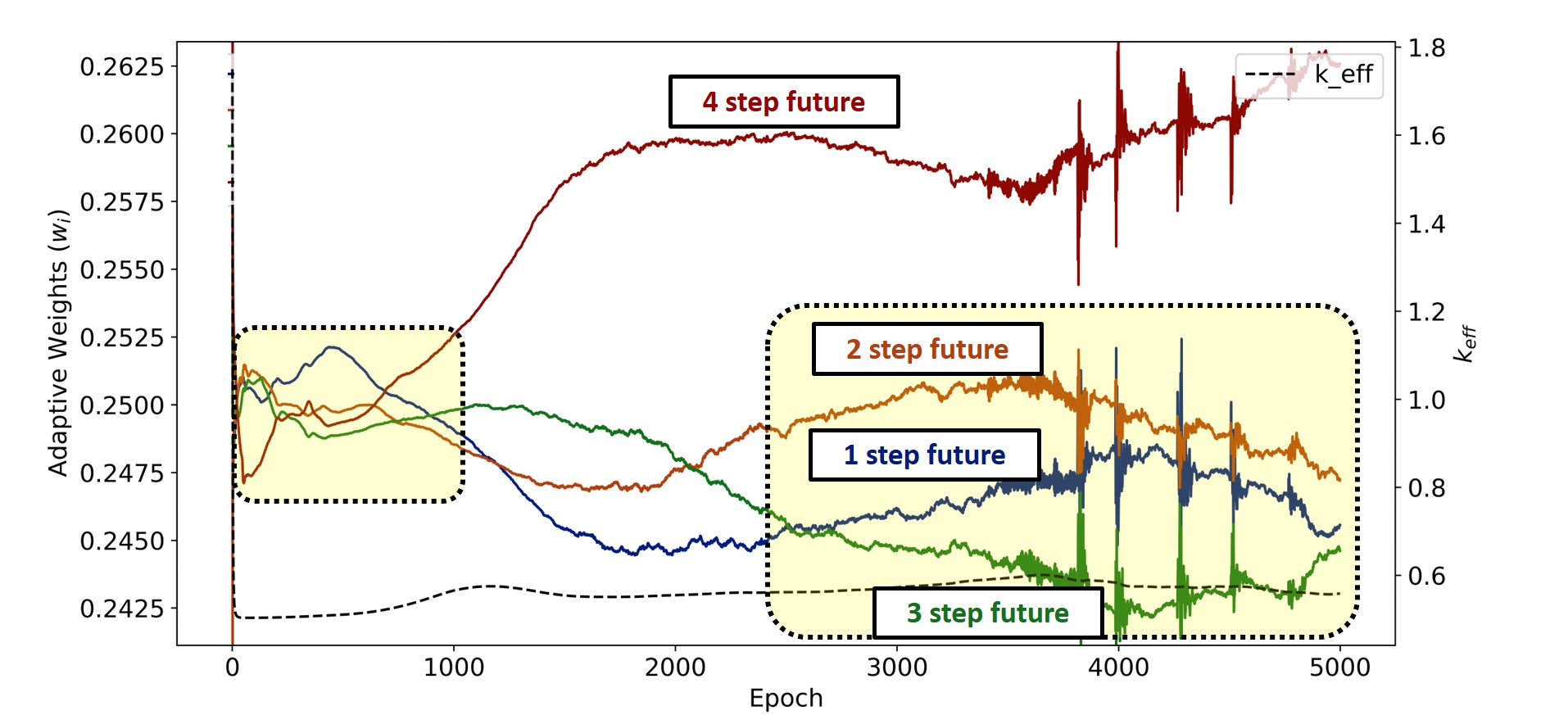}
    \caption{Evolution of adaptive weights for each loss term over epochs in AW2 with Adams-Euler scheme ($M=4$). Adaptive weights ($w_i$) and effective parameter ($k_{eff}$) shown on y-axes.}
    \label{fig:adaptive_weights_evolution}
\end{figure}

To understand how adaptive weights evolve during training, Figure \ref{fig:adaptive_weights_evolution} shows weight evolution for AW2 with Adams-Euler. Two distinct phases emerge: (1) Stabilization phase (before 1000 epochs): minimal weight differences as the model learns basic temporal patterns; (2) Adaptation phase (after 1000 epochs): increased weight on the 4th loss term, indicating greater difficulty in long-term predictions. This trend validates the AW3 approach—since the first three loss terms show little distinction, emphasizing only the first and last steps captures the essential trade-off between immediate and long-term accuracy: evolution of adaptive weights under AW3 approach can be found in Figure \ref{fig:adaptive_weights_evolution_AW3}, \ref{sec:app_AW3}.

\begin{figure*}[htb!]
    \centering
    \begin{subfigure}[h]{0.49\textwidth}
        \centering
        \includegraphics[width=\textwidth]{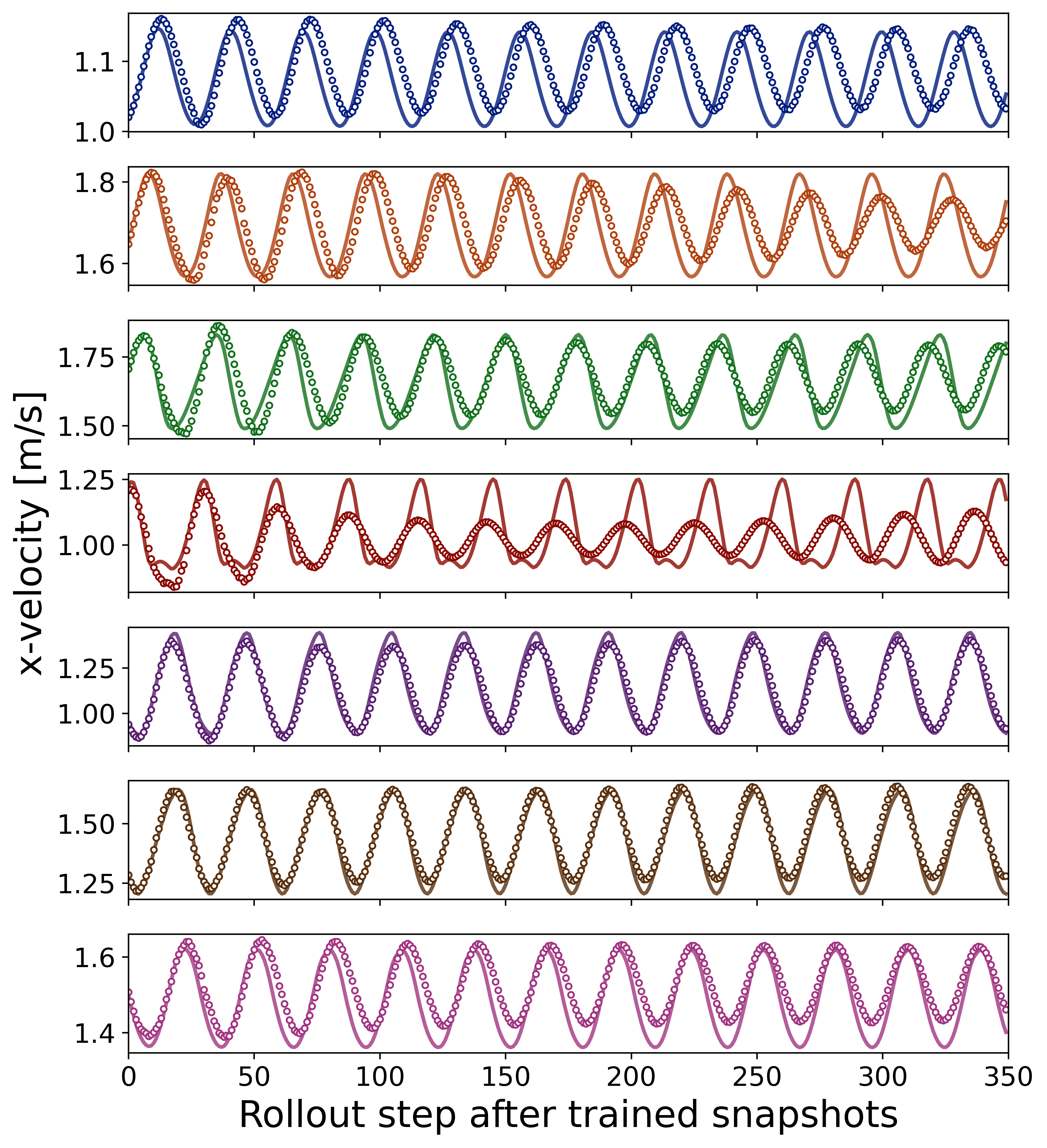}
        \caption{Model A: Forward Euler with AW2}
        \label{fig:time_series_probe_points_a}
    \end{subfigure}
    \hfill
    \begin{subfigure}[h]{0.49\textwidth}
        \centering
        \includegraphics[width=\textwidth]{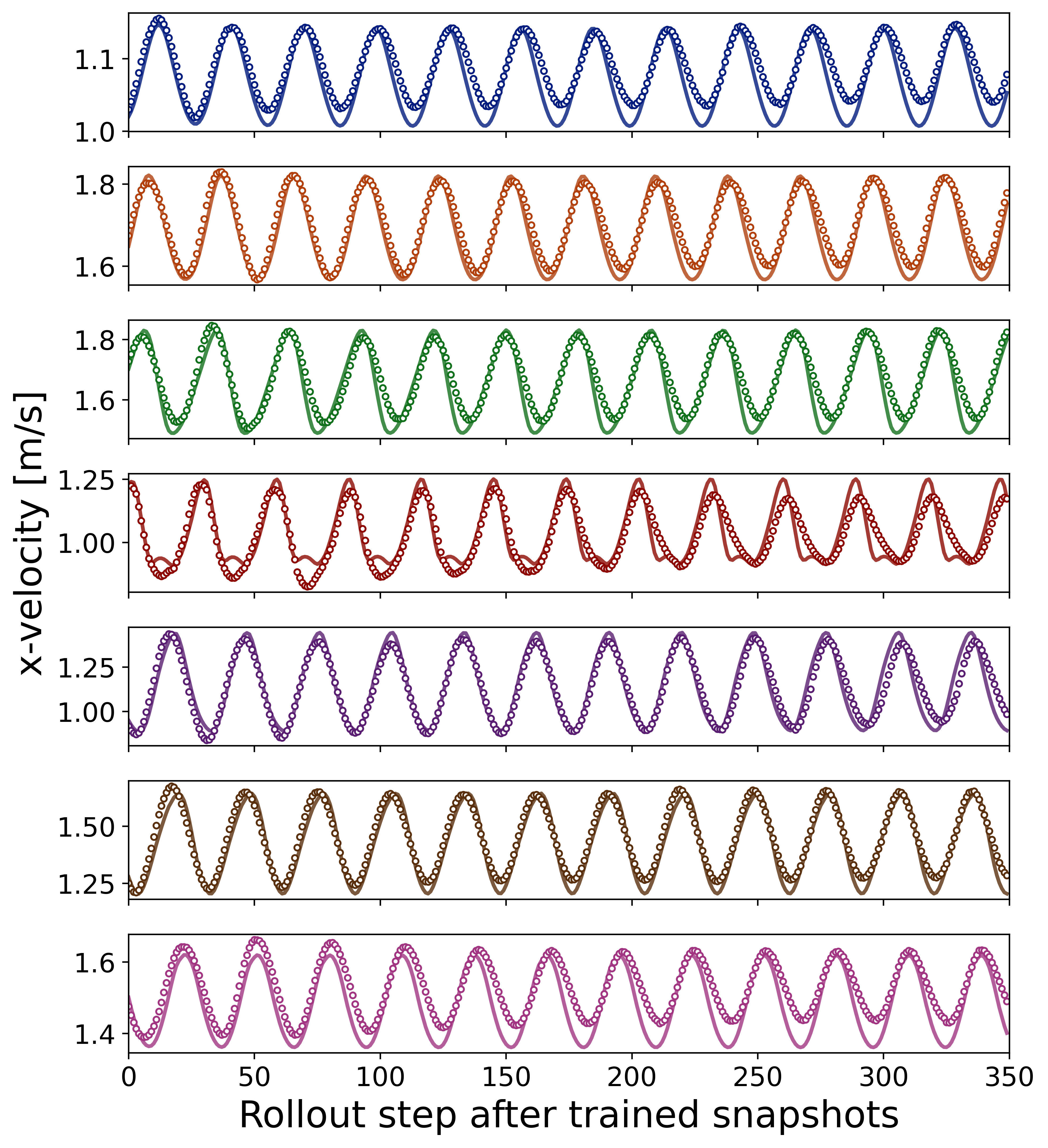}
        \caption{Model B: Adams-Euler with AW3}
        \label{fig:time_series_probe_points_b}
    \end{subfigure}

    \begin{subfigure}[h]{0.49\textwidth}
        \centering
        \includegraphics[width=\textwidth]{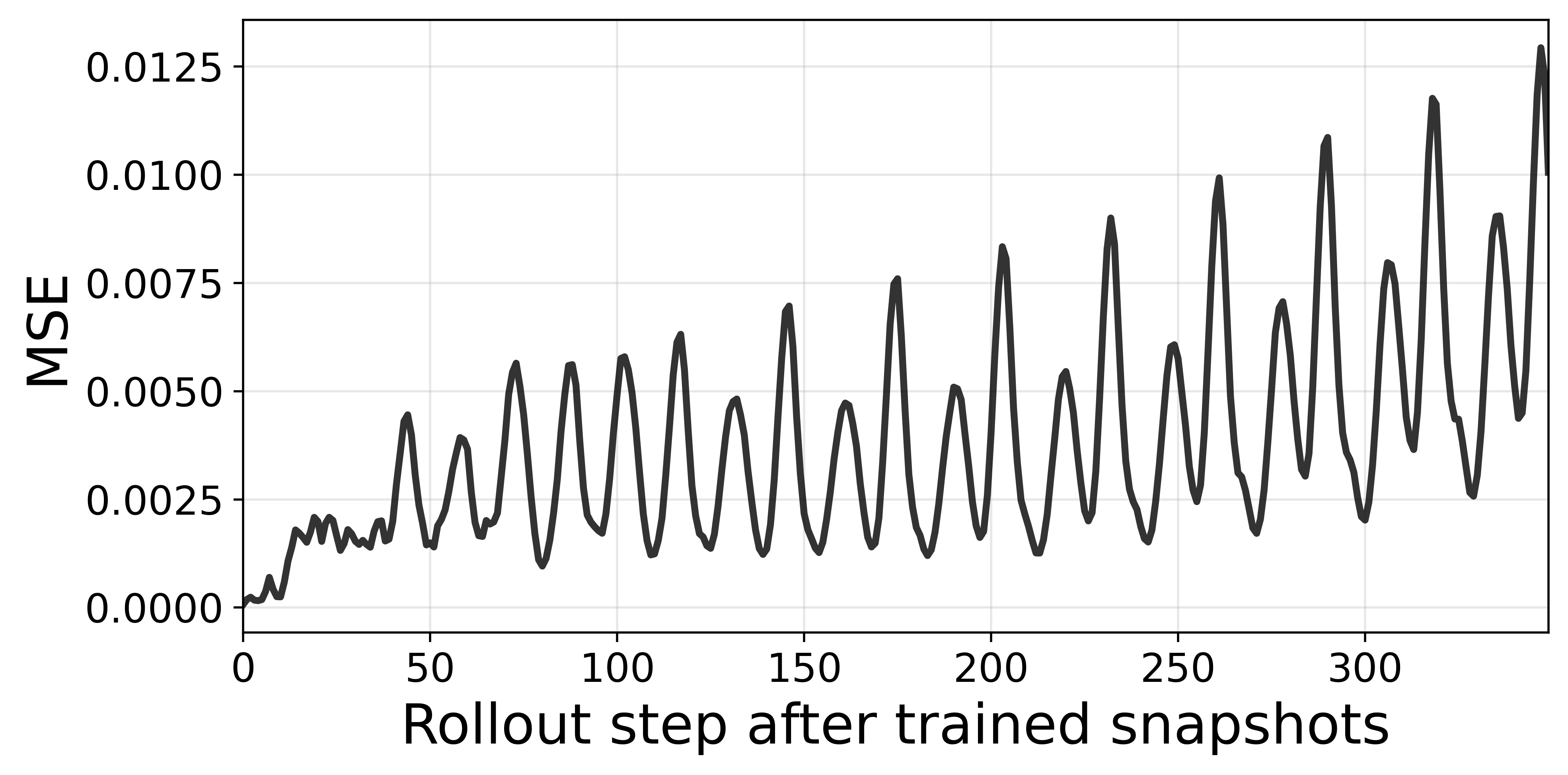}
        \caption{Time-varying MSE of Model A}
        \label{fig:time_series_probe_points_c}
    \end{subfigure}
    \hfill
    \begin{subfigure}[h]{0.49\textwidth}
        \centering
        \includegraphics[width=\textwidth]{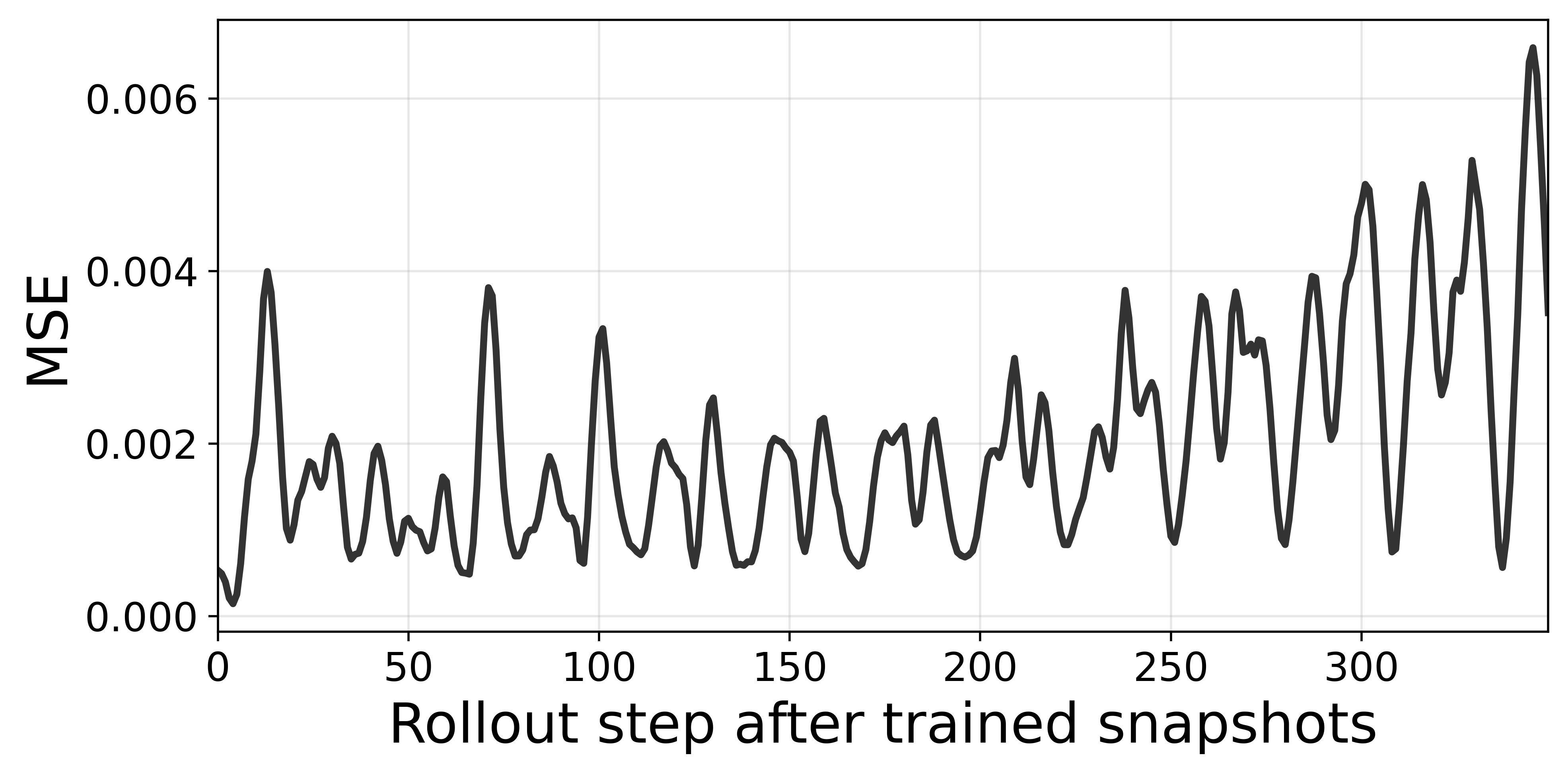}
        \caption{Time-varying MSE of Model B}
        \label{fig:time_series_probe_points_d}
    \end{subfigure}
    
    \caption{(a-b): $x$-velocity time series at seven probe points over 350 future snapshots. From bottom to top: probe points 1-7 (Figure \ref{fig:probe_points_locations}). Solid lines: ground truth; circles: model predictions. (c-d): time-varying MSE for each model, averaged over the seven probe points.}
    \label{fig:time_series_probe_points}
\end{figure*}

To quantitatively evaluate these improvements, we analyze the temporal evolution of the predictions by Model A (forward Euler with AW2) and B (Adams-Euler with AW3) selected in Table \ref{tab:mse_comparison_adaptive_weighting}. Figure \ref{fig:time_series_probe_points} shows the 350-snapshot forecast, with the corresponding time-varying MSE plots providing a measure of error accumulation. Model A's predictions (Figure \ref{fig:time_series_probe_points_a}) show noticeable discrepancies in amplitude and phase. This is confirmed by its time-varying MSE plot (Figure \ref{fig:time_series_probe_points_c}), which reveals a clear upward trend indicating steady error accumulation that reaches a maximum MSE of approximately 0.0125. In stark contrast, Model B's predictions (Figure \ref{fig:time_series_probe_points_b}) closely match the ground truth. Its superior stability is highlighted in the corresponding MSE plot (Figure \ref{fig:time_series_probe_points_d}), where the error stays bounded below an MSE of 0.004 for the majority of the horizon (up to 250 steps), peaking at just 0.006—less than half the maximum error of Model A. This demonstrates that Model B's training strategy, Adams-Euler with AW3, more effectively mitigates the compounding error inherent in long-term auto-regressive forecasting.

These results demonstrate successful long-term prediction over 350 rollout steps despite challenging conditions: severely constrained model capacity (1,177 parameters), minimal training data (50 snapshots), and extensive prediction horizon (future 350 snapshots). This achievement stems from two key innovations: (1) Adams-Euler time integration providing enhanced numerical stability, and (2) AW3's adaptive weighting strategy that strategically focuses on the most critical prediction steps during multi-step rollout training.

The following subsections present critical comparisons against conventional noise injection approaches (Section \ref{sec:noise}) and evaluation under even more challenging partial domain training conditions (Section \ref{sec:harsh}) to further validate the robustness and superiority of our proposed methodology.

\subsection{Comparison with conventional noise-injection approach for long-term rollout} \label{sec:noise}

We compare the long-term rollout performance of our proposed framework against noise injection, a widely adopted technique for enhancing model robustness against error accumulation in AR prediction \cite{kim2024physics,pfaff2020learning,sanchez2020learning,yang2024enhancing}. This method deliberately adds random noise to input data during training to enhance model robustness against error accumulation. The underlying principle is that by exposing the model to perturbed input data during training, it becomes more resilient to imperfect data when its own flawed predictions are fed back as inputs during inference.

We evaluate three model configurations (direct prediction, forward Euler, and Adams-Euler time integration schemes) trained with Gaussian noise $\mathcal{N}(0, 0.16^2)$ injected into input data. This noise level, identified as optimal in our previous work \cite{yang2024enhancing}, is consistently applied during training without multi-step rollouts. 

\begin{figure*}[htb!]
    \centering
    \begin{subfigure}[h]{0.49\textwidth}
        \centering
        \includegraphics[width=\textwidth]{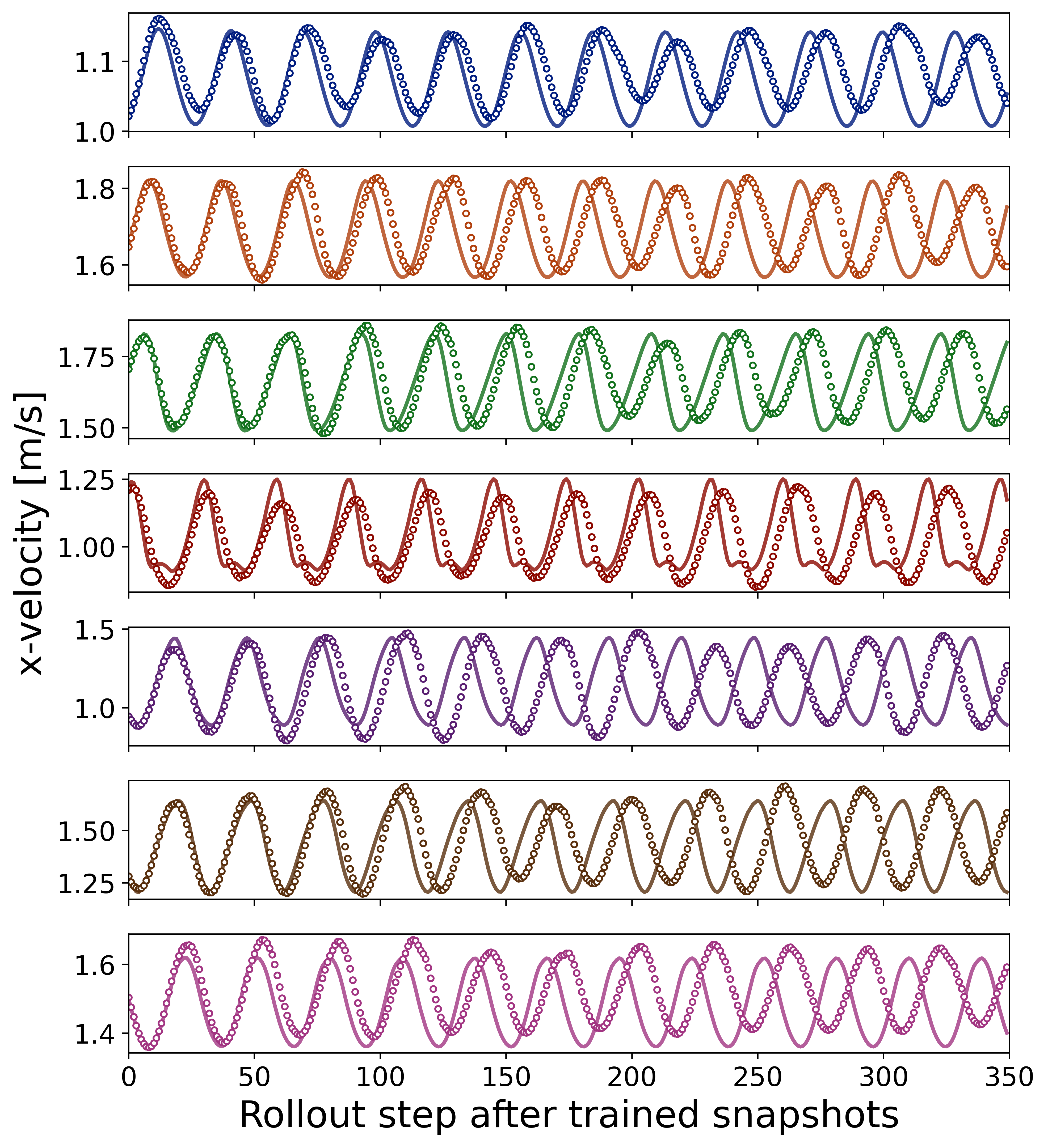}
        \caption{Forward Euler with noise injection}
        \label{fig:time_series_noise_injection_a}
    \end{subfigure}
    \hfill
    \begin{subfigure}[h]{0.49\textwidth}
        \centering
        \includegraphics[width=\textwidth]{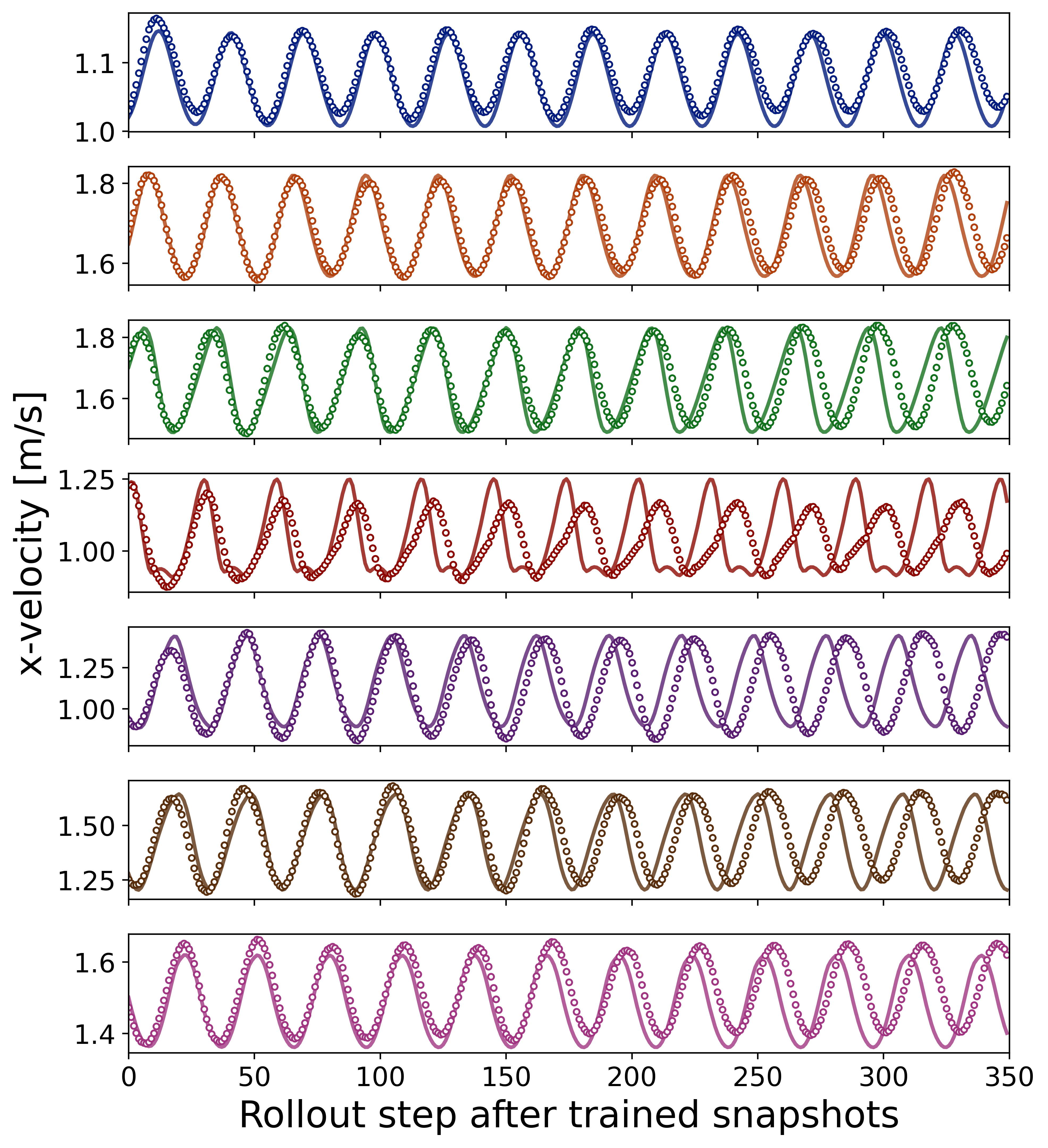}
        \caption{Adams-Euler with noise injection}
        \label{fig:time_series_noise_injection_b}
    \end{subfigure}

    \begin{subfigure}[h]{0.49\textwidth}
        \centering
        \includegraphics[width=\textwidth]{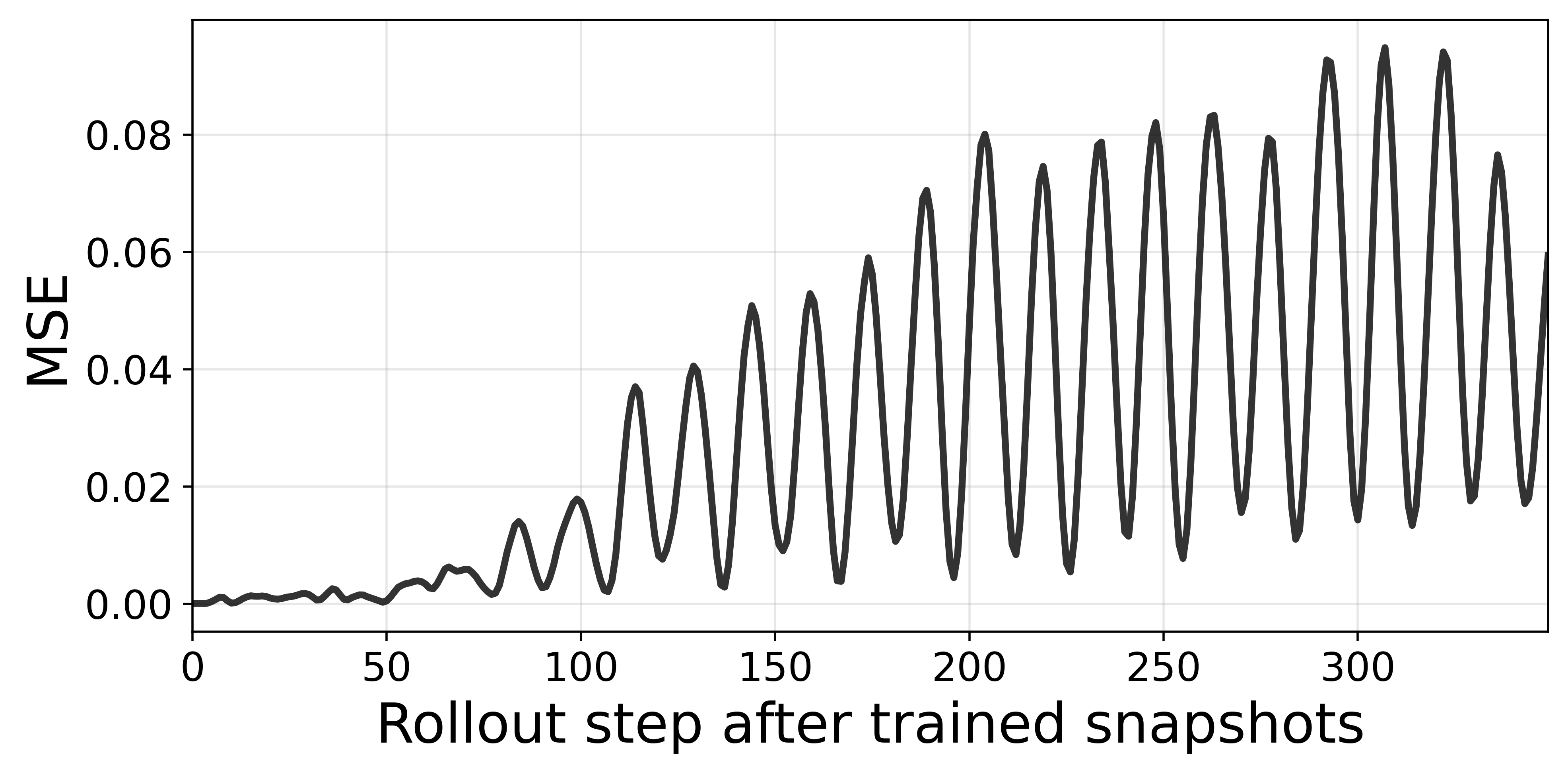}
        \caption{Time-varying MSE of forward Euler with noise injection model}
        \label{fig:time_series_noise_injection_c}
    \end{subfigure}
    \hfill
    \begin{subfigure}[h]{0.49\textwidth}
        \centering
        \includegraphics[width=\textwidth]{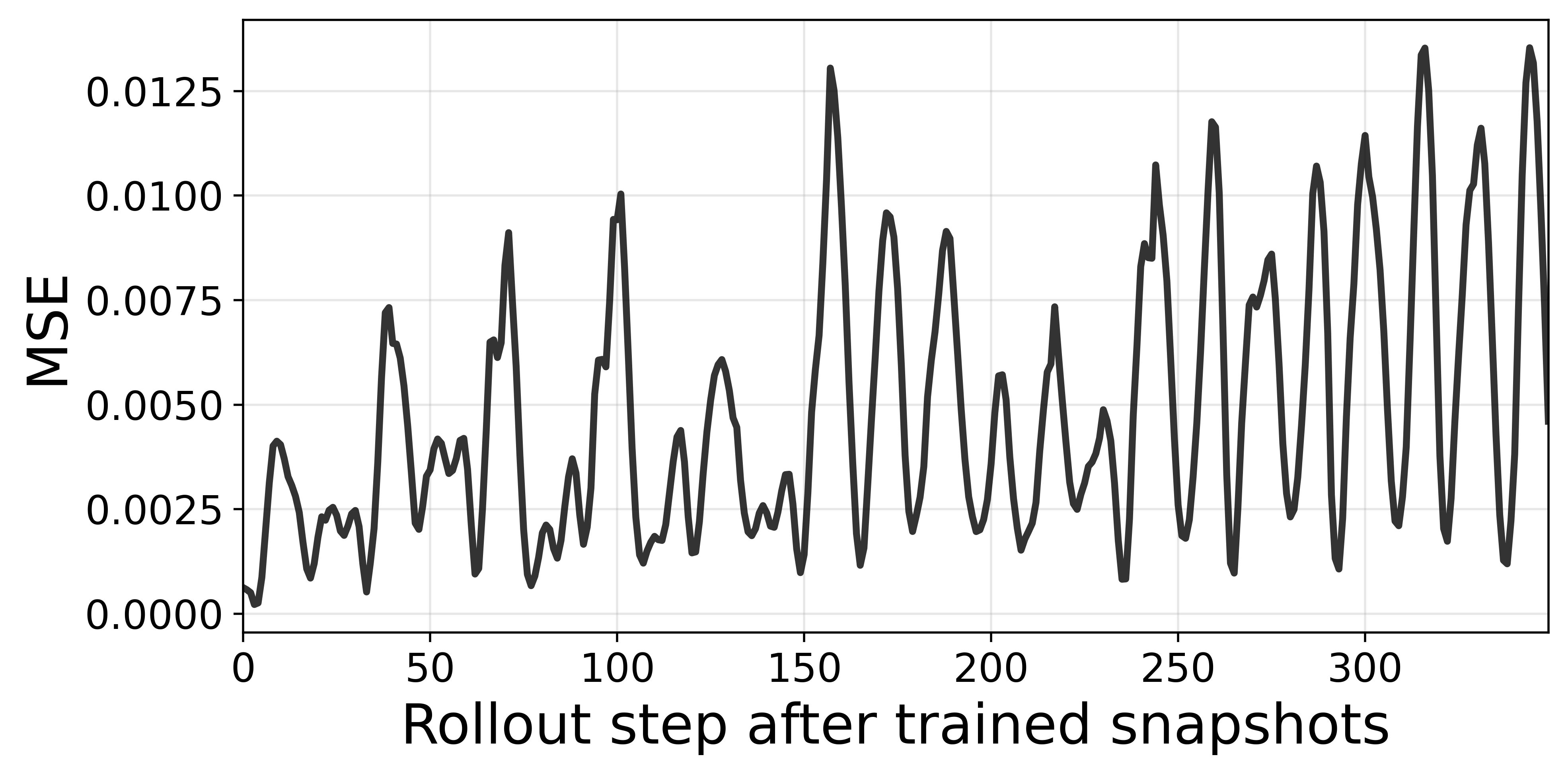}
        \caption{Time-varying MSE of Adams-Euler with noise injection model}
        \label{fig:time_series_noise_injection_d}
    \end{subfigure}
    
    \caption{(a-b) $x$-velocity time series at probe points over 350 future snapshots using noise injection. Solid lines: ground truth; circles: model predictions. (c-d): time-varying MSE for each model, averaged over the seven probe points.}
    \label{fig:time_series_noise_injection}
\end{figure*}

The effectiveness of this approach is assessed using the temporal behavior of the $x$-velocity and the time-varying MSE at seven probe points downstream of the cylinder (Figure \ref{fig:time_series_noise_injection}). The forward Euler model's time-series plot (Figure \ref{fig:time_series_noise_injection_a}) shows significant deviations from the ground truth in both amplitude and phase, indicating that noise injection alone is insufficient for maintaining long-term accuracy. This instability is confirmed quantitatively by its time-varying MSE plot (Figure \ref{fig:time_series_noise_injection_c}), where the error grows rapidly to a maximum of approximately 0.1. In comparison, the Adams-Euler model, while still less accurate than our full framework, demonstrates greater stability in its time-series (Figure \ref{fig:time_series_noise_injection_b}). This is clearly evidenced by its corresponding MSE plot (Figure \ref{fig:time_series_noise_injection_d}), which shows the error peaking at just 0.0135—nearly an order of magnitude smaller than that of the forward Euler model. Comparing two schemes with noise injection, Adams-Euler (Figure \ref{fig:time_series_noise_injection_b}) still significantly outperforms forward Euler (Figure \ref{fig:time_series_noise_injection_a}), further demonstrating the robust performance of the Adams-Bashforth time integration scheme across various training approaches.

Table \ref{tab:mse_comparison_adaptive_noise} quantifies performance differences with MSE calculations for seven probe points. It's important to note that noise injection itself provides substantial improvements over baseline approaches—comparing to Table \ref{tab:time_mse_comparison} where no multi-step rollout or noise injection was applied (Direct: 0.125, Forward Euler: 0.138, Adams-Euler: 0.139), noise injection achieves significant performance gains (Direct: 0.019, Forward Euler: 0.017, Adams-Euler: 0.012), demonstrating its established effectiveness in the AR prediction community. However, for all time integration schemes, models trained with adaptive multi-step rollout consistently outperform even these improved noise injection results. The Adams-Euler scheme shows particularly dramatic enhancement, with the adaptive multi-step rollout (MSE of 0.002) achieving an 83\% error reduction compared to the noise injection approach (MSE of 0.012)—representing not only a substantial improvement over baseline methods but also a significant advance beyond the already powerful noise injection technique.

\begin{table}[htb!]
    \centering
    \begin{threeparttable}
        \caption{MSE comparison between adaptive multi-step rollout, noise injection, and a combined approach. Each multi-step rollout configuration uses the best adaptive weighting approach (footnoted). MSE values calculated from seven probe points in Figure \ref{fig:probe_points_locations}.}
        \label{tab:mse_comparison_adaptive_noise}
        \begin{tabular}{lccc}
            \hline
            \textbf{Time Scheme} & \textbf{Multi-Step Rollout} & \textbf{Noise Injection} & \textbf{Combined} \\
            \hline
            Direct Prediction & 0.011\tnote{1} & 0.019 & \textbf{0.008} \\
            Forward Euler & \textbf{0.007}\tnote{2} & 0.017 & 0.029 \\
            Adams-Euler & \textbf{0.002}\tnote{3} & 0.012 & 0.021 \\ \hline
            \textbf{Averaged time [s]} & 2354 & \textbf{1806} & 2586 \\
            \hline
        \end{tabular}
        \begin{tablenotes}[flushleft]
            \footnotesize
            \item[1] Without adaptive weighting
            \item[2] With AW2
            \item[3] With AW3
        \end{tablenotes}
    \end{threeparttable}
\end{table}

The performance gap stems from fundamental differences in addressing error accumulation. Noise injection builds resilience through input perturbations, making models more robust to small deviations, but does not explicitly address temporal dependencies and error propagation mechanisms inherent in AR predictions. In contrast, adaptive multi-step rollout enables learning from multiple future steps simultaneously, using strategically adjusted loss weights to focus on both immediate and distant predictions. This fundamental difference explains why adaptive multi-step rollout more accurately captures vortex shedding patterns compared to noise injection, as demonstrated in the time series comparisons (Figures \ref{fig:time_series_probe_points} and \ref{fig:time_series_noise_injection}).

We also investigated combining multi-step rollout with noise injection techniques, which revealed contrasting behaviors as shown in Table \ref{tab:mse_comparison_adaptive_noise}. The combined approach yields the best performance for direct prediction, while for derivative-based methods, this combination leads to training instability and significantly degrades performance. This contrasting behavior can be attributed to the interaction between the two regularization techniques. For direct prediction, the methods are complementary: noise injection robustifies the model to the exact type of input errors it will encounter during the multi-step rollout. Conversely, for derivative-based methods, the two techniques impose conflicting objectives. Multi-step rollout pushes the model to learn a precise, deterministic temporal trajectory. Noise injection, a form of stochastic regularization, forces the model to be robust to a distribution of inputs around that trajectory. When combined, noise is repeatedly propagated and potentially amplified through the unrolled computational graph of the multi-step loss. We hypothesize that this compounding of stochastic perturbations leads to unstable optimization process with higher MSE values.

\subsection{Robustness evaluation under challenging conditions}
\label{sec:harsh}
To further validate our framework's robustness, we evaluate its performance under three challenging conditions designed to simulate practical engineering constraints: (1) training with limited spatial information, where the model only sees a subset of the domain; (2) training with a larger time-step, which tests the numerical stability of the integration schemes; and (3) training on multiple flow scenarios simultaneously to assess the model's generalization capability.

\subsubsection{Performance under partial domain training}
\label{sec:harsh_1}
First, we evaluate performance under partial domain training, where models are trained on only a spatial subset ($0.3 < x < 0.75$ and $0.128 < y < 0.328$) of the original vortex shedding domain. This challenging scenario simulates practical applications with limited spatial coverage or memory constraints.

\begin{figure*}[htb!]
    \centering
    \begin{subfigure}[h]{0.45\textwidth}
        \centering
        \includegraphics[width=\textwidth]{GT0_test_idx4_snap520_.png}
        \caption{Ground truth flow field at snapshot $t+300$ (full domain reference)}
        \label{fig:truncated_velocity_comparison_a}
    \end{subfigure}
    
    \vfill
    
    \begin{subfigure}[h]{0.45\textwidth}
        \centering
        \includegraphics[width=\textwidth, trim=0 0 0 21, clip]{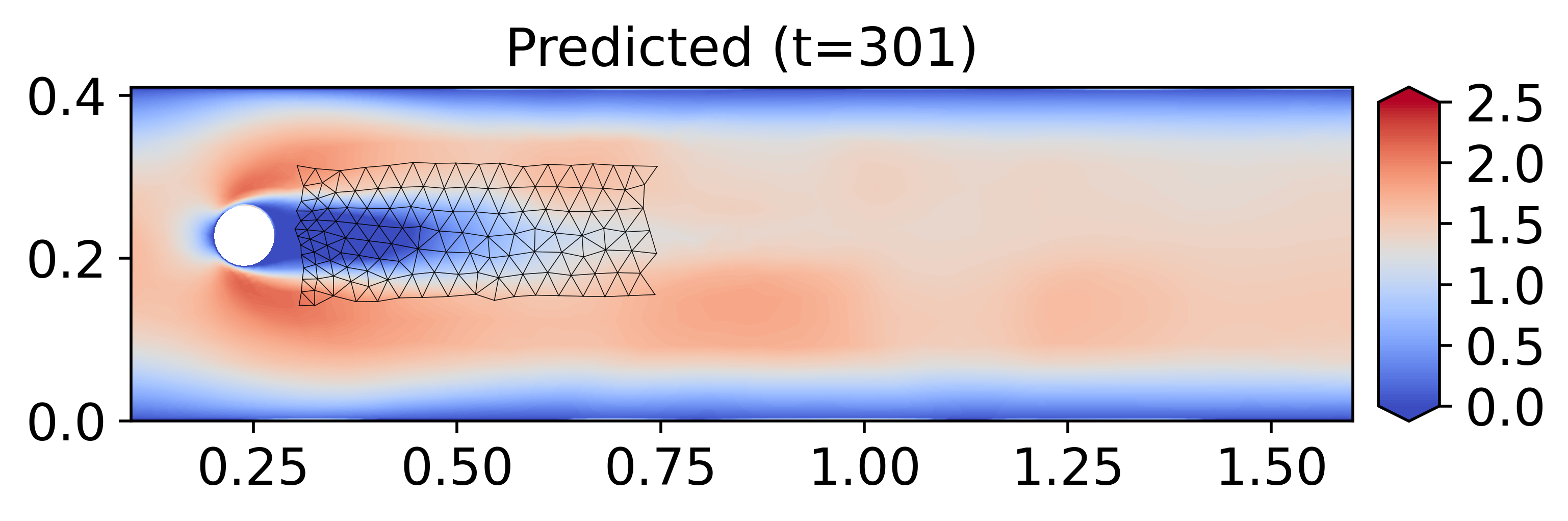}
        \caption{Direct prediction with vanilla multi-step rollout}
        \label{fig:truncated_velocity_comparison_b}
    \end{subfigure}
    \hfill
    \begin{subfigure}[h]{0.45\textwidth}
        \centering
        \includegraphics[width=\textwidth, trim=0 0 0 21, clip]{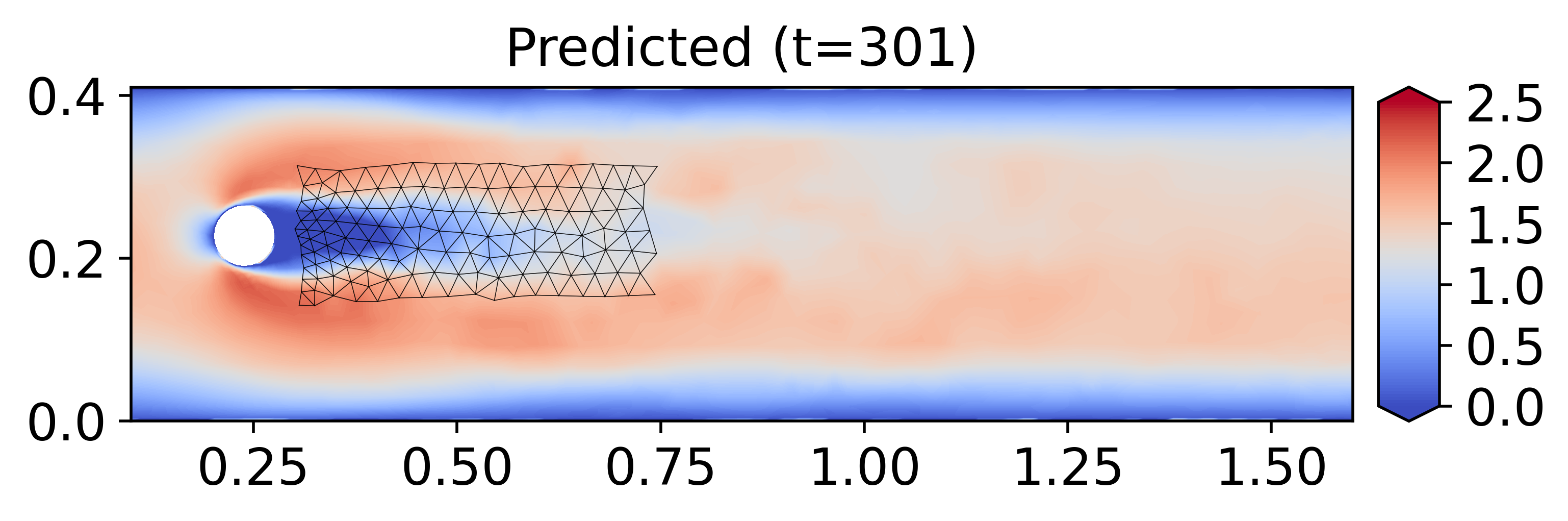}
        \caption{Forward Euler with AW2}
        \label{fig:truncated_velocity_comparison_c}
    \end{subfigure}
    
    \vfill
    
    \begin{subfigure}[h]{0.45\textwidth}
        \centering
        \includegraphics[width=\textwidth, trim=0 0 0 21, clip]{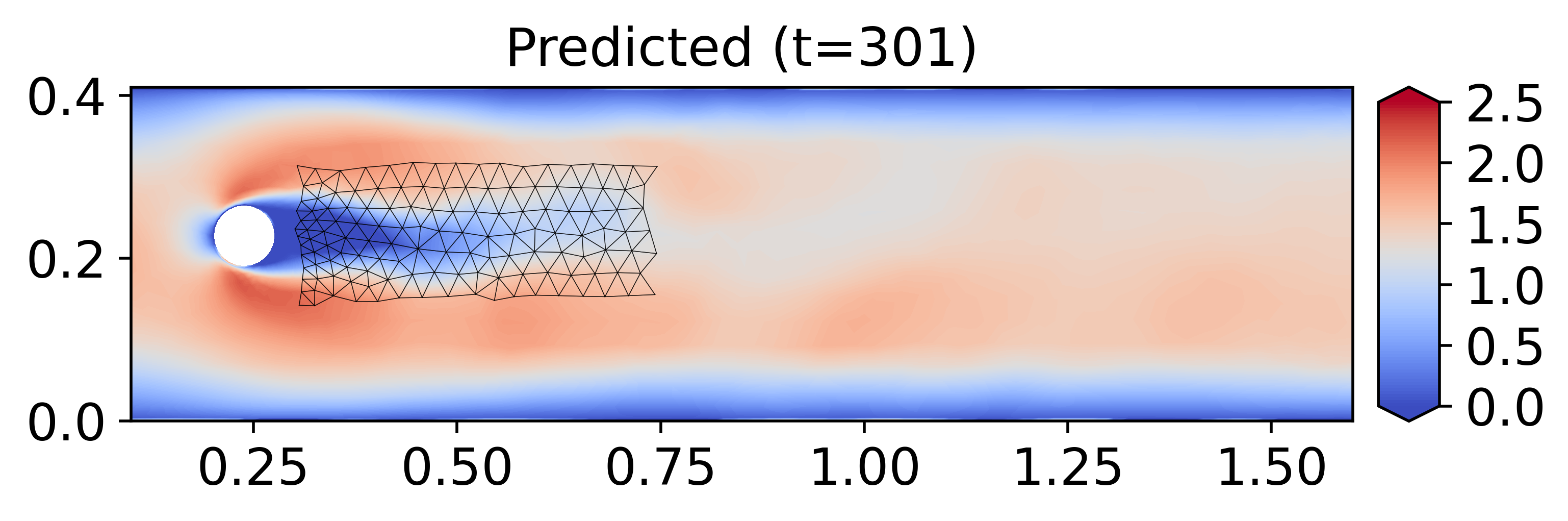}
        \caption{Forward Euler with AW3}
        \label{fig:truncated_velocity_comparison_d}
    \end{subfigure}
    \hfill
    \begin{subfigure}[h]{0.45\textwidth}
        \centering
        \includegraphics[width=\textwidth, trim=0 0 0 21, clip]{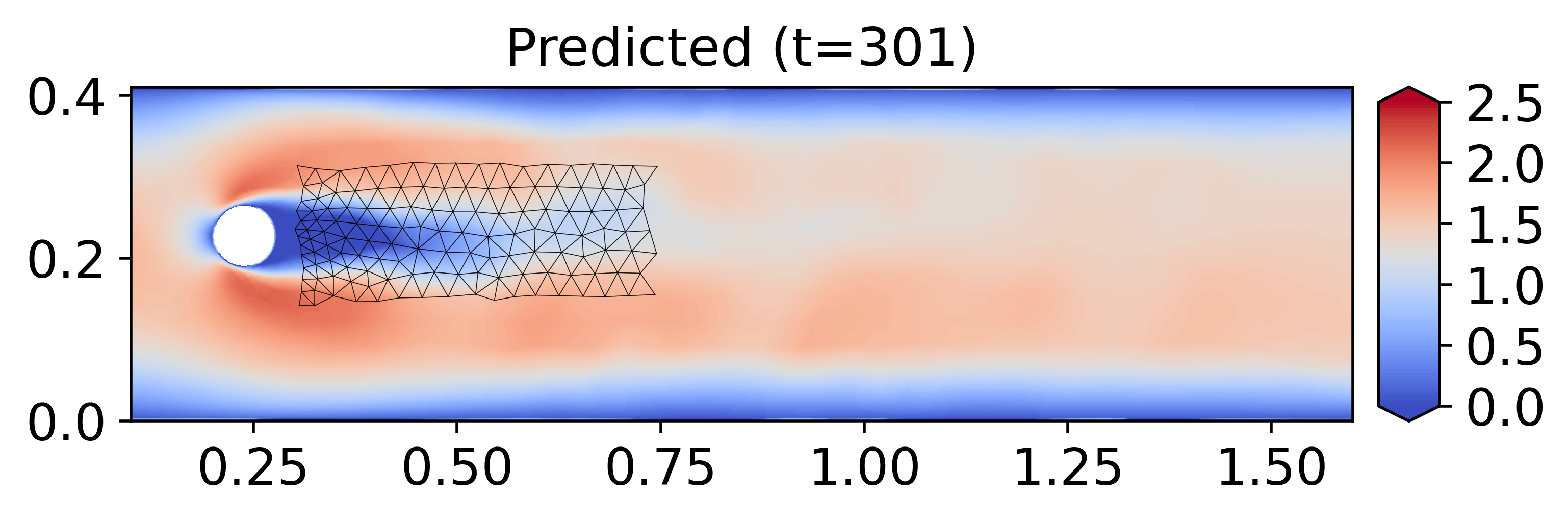}
        \caption{Adams-Euler with AW3}
        \label{fig:truncated_velocity_comparison_e}
    \end{subfigure}
    
    \caption{Predicted $x$-velocity fields in partial domain after 300 rollout steps. The partial domain mesh region shown in (b)-(e) was used for training.}
    \label{fig:truncated_velocity_comparison}
\end{figure*}

We test the four models that previously demonstrated satisfactory results from Table \ref{tab:mse_comparison_adaptive_weighting}: direct prediction with vanilla multi-step rollout, forward Euler with AW2, forward Euler with AW3, and Adams-Euler with AW3. Figure \ref{fig:truncated_velocity_comparison} shows flow field predictions after 300 rollout steps: partial domain mesh used for training is visualized in Figure \ref{fig:truncated_velocity_comparison_b}$\sim$\ref{fig:truncated_velocity_comparison_e}. Direct prediction fails completely, while forward Euler methods show limited success. Adams-Euler with AW3 demonstrates the best performance, maintaining high accuracy and successfully reproducing vortex shedding patterns within the constrained region.

\begin{figure*}[htb!]
    \centering
    \begin{subfigure}[h]{0.49\textwidth}
        \centering
        \includegraphics[width=\textwidth]{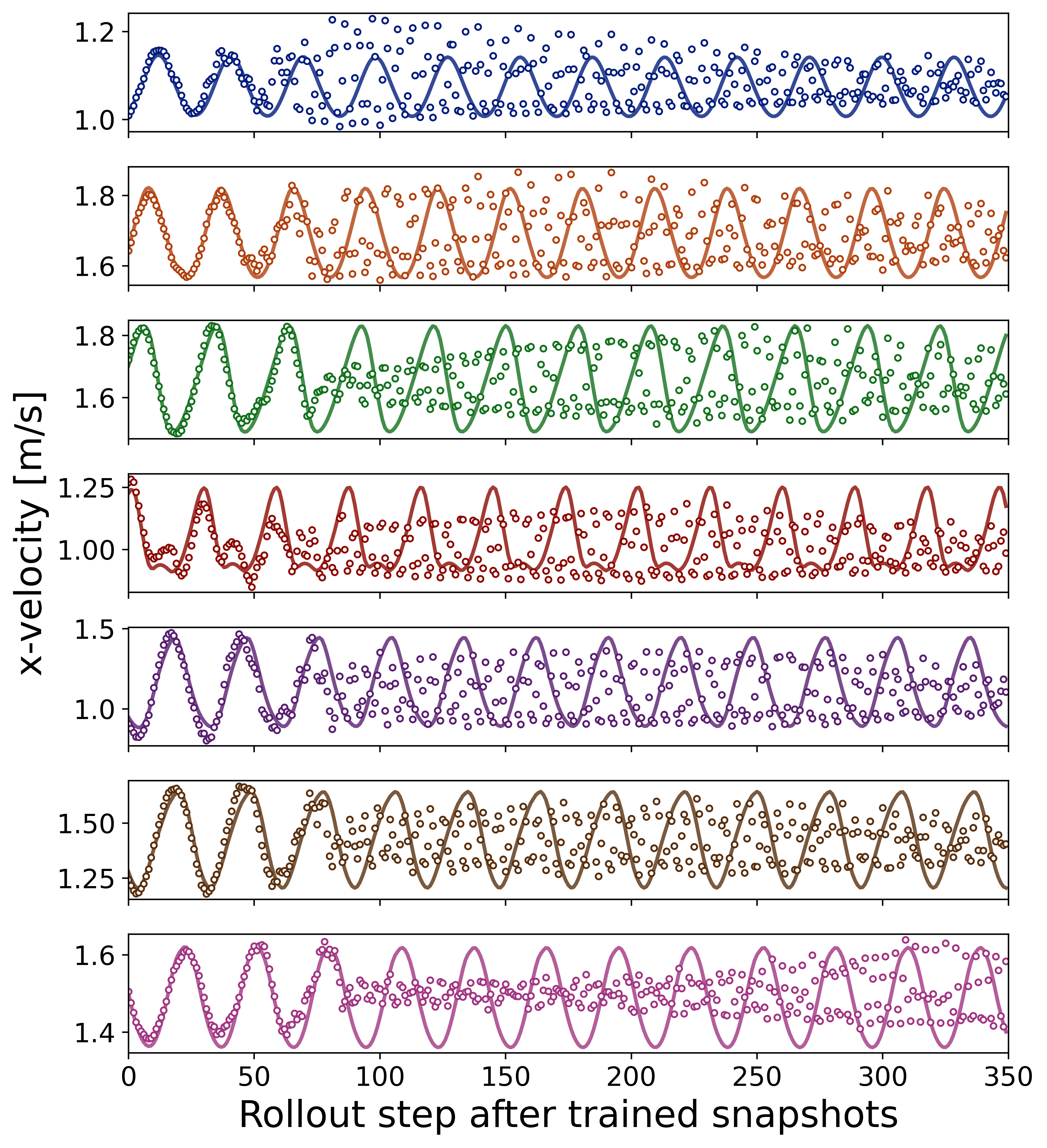}
        \caption{Direct prediction with vanilla multi-step rollout (MSE=0.019)}
        \label{fig:probe_4models_a}
    \end{subfigure}
    \hfill
    \begin{subfigure}[h]{0.49\textwidth}
        \centering
        \includegraphics[width=\textwidth]{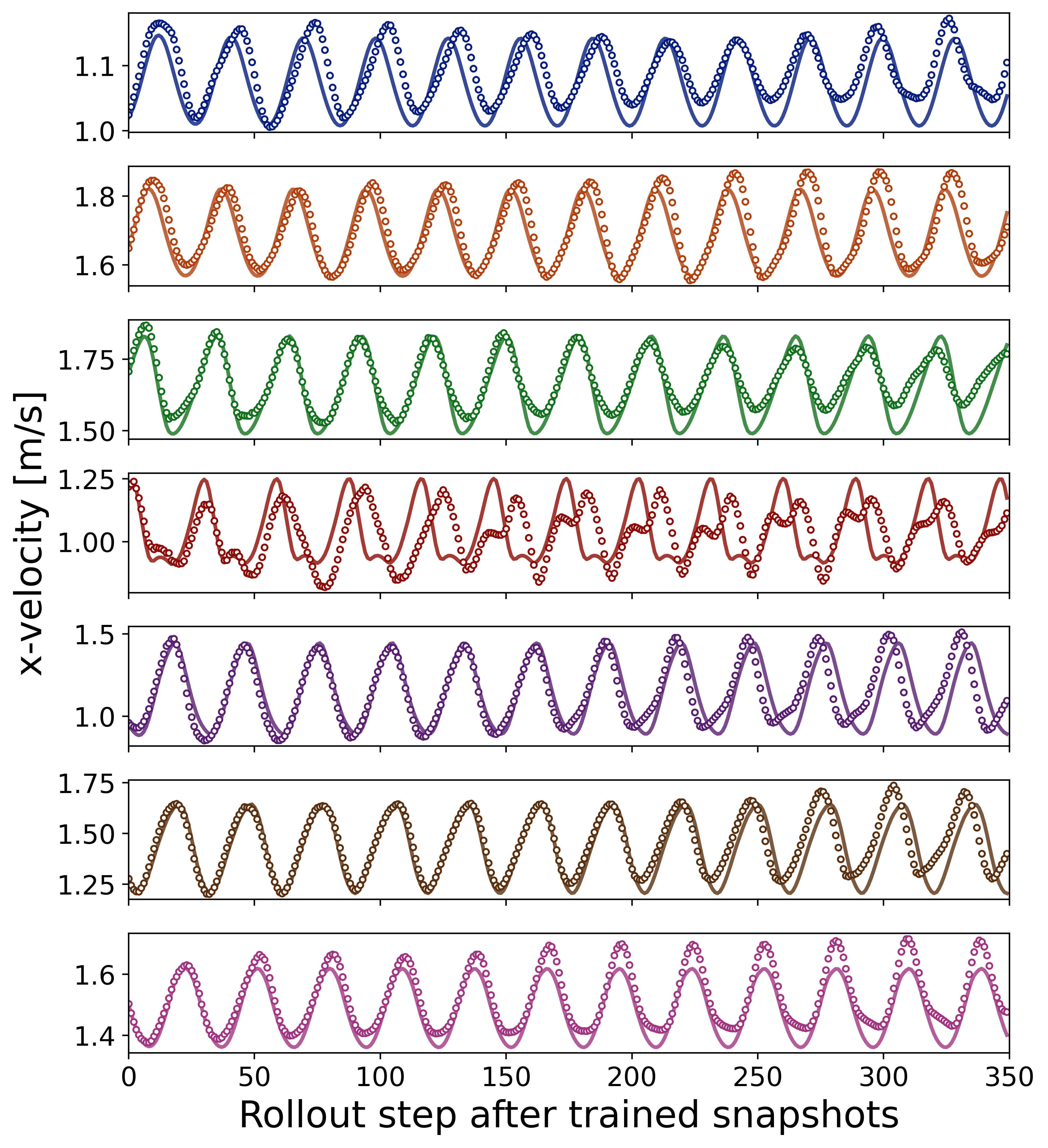}
        \caption{Forward Euler with AW2 (MSE=0.011)}
        \label{fig:probe_4models_b}
    \end{subfigure}
    \vfill 
    \begin{subfigure}[h]{0.49\textwidth}
        \centering
        \includegraphics[width=\textwidth]{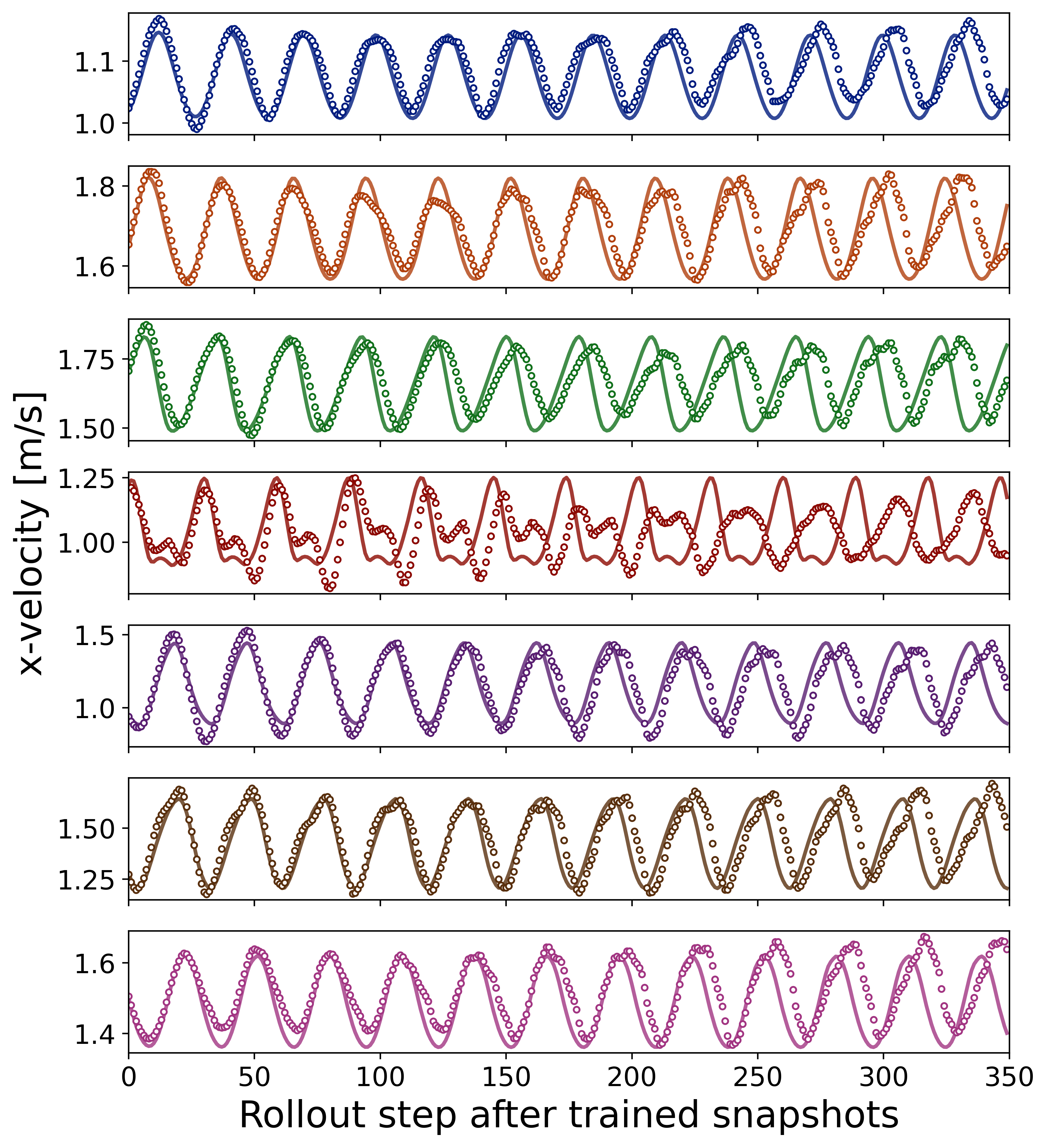}
        \caption{Forward Euler with AW3 (MSE=0.013)}
        \label{fig:probe_4models_c}
    \end{subfigure}
    \hfill
    \begin{subfigure}[h]{0.49\textwidth}
        \centering
        \includegraphics[width=\textwidth]{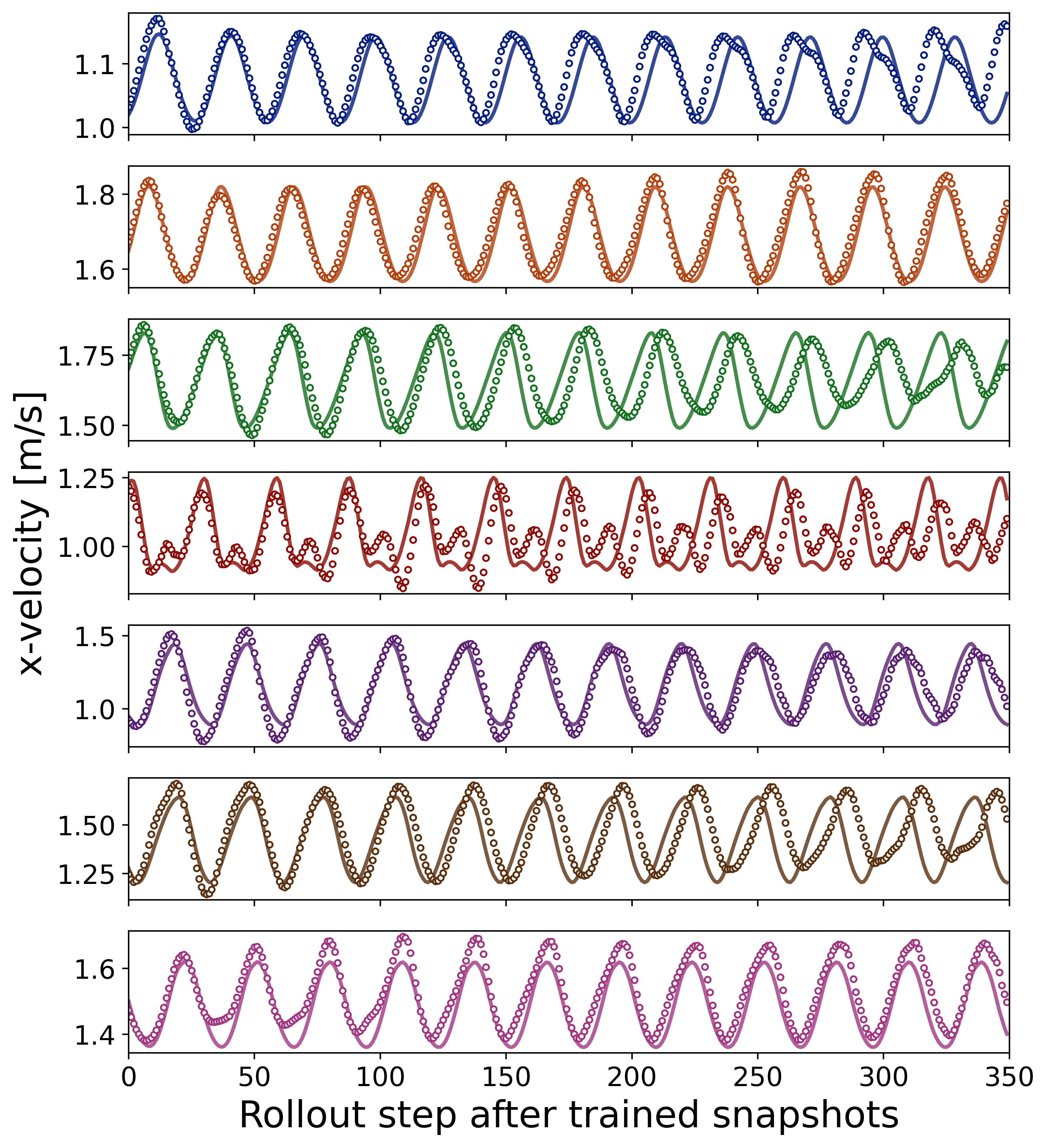}
        \caption{Adams-Euler with AW3 (MSE=0.008)}
        \label{fig:probe_4models_d}
    \end{subfigure}
    
    \caption{Time series of $x$-velocity at probe points over 350 future snapshots for partial domain training. Solid lines: ground truth; circles: predictions.}
    \label{fig:probe_4models}
\end{figure*}

Quantitative analysis using seven probe points (Figure \ref{fig:probe_4models}) confirms these results. Direct prediction completely fails to capture vortex shedding oscillations (MSE: 0.019), while forward Euler methods show gradual degradation (MSE: 0.011-0.013). Adams-Euler with AW3 achieves the lowest error (MSE: 0.008), accurately maintaining vortex shedding frequency, amplitude, and phase throughout the entire 350-step prediction horizon. These results validate the exceptional robustness of our Adams-Bashforth time integration with adaptive multi-step rollout framework, confirming its broad applicability for practical engineering applications where complete spatial information may be unavailable.

\subsubsection{Robustness to increased time-step size}
\label{sec:harsh_2}
A critical challenge for derivative-based prediction methods is their sensitivity to the temporal discretization, $\Delta t$. Because the model learns to approximate the temporal derivative based on a specific, fixed time step from the training data, the same $\Delta t$ must be used during the auto-regressive inference stage. To investigate this sensitivity and address concerns about the required temporal resolution, we conducted an additional ablation study to evaluate the framework's stability under a coarser temporal resolution. For the cylinder flow case, where the original dataset has a fixed time step of $\Delta t = 0.01s$, we retrained and evaluated the same four model configurations in Section~\ref{sec:harsh_1} on a downsampled dataset with a doubled time step of $\Delta t = 0.02s$. Since the total time period for training was kept constant, this effectively halved the number of snapshots available to the model.

The results, summarized in Table \ref{tab:timestep_ablation}, reveal a stark degradation in performance for all models when trained on the coarser dataset. The MSE for all models increased by at least an order of magnitude. Our best-performing model (Adams-Euler + AW3) saw its MSE rise from 0.002 to 0.212, a hundred-fold increase. This failure is further evidenced by the highly inaccurate Strouhal number predictions. For instance, the direct prediction model predicted a frequency nearly double the ground truth ($St=0.2991$), while the derivative-based methods predicted frequencies less than half the true value ($St=0.065$).

\begin{table}[htb!]
\centering
\caption{Time extrapolation performance comparison of top models when trained with the original time step ($\Delta t = 0.01s$) versus a doubled time step ($\Delta t = 0.02s$). Ground truth Strouhal number is 0.1438.}
\label{tab:timestep_ablation}
\begin{tabular}{l|cc|cc}
\hline
\multirow{2}{*}{\textbf{Model Configuration}} & \multicolumn{2}{c|}{\textbf{Original ($\Delta t = 0.01s$)}} & \multicolumn{2}{c}{\textbf{Doubled ($\Delta t = 0.02s$)}} \\ 
\cline{2-5}
& MSE & $St$ & MSE & $St$ \\ 
\hline
Direct + Vanilla Rollout & 0.011 & 0.1407 & 0.0416 & 0.2991 \\
Forward Euler + AW2 & 0.007 & 0.1489 & 0.1093 & 0.1516 \\
Forward Euler + AW3 & 0.010 & 0.1407 & 0.2278 & 0.0656 \\
Adams-Euler + AW3 & 0.002 & 0.1434 & 0.2120 & 0.0650 \\ 
\hline
\end{tabular}
\end{table}

These findings underscore that while the choice of a time step can be critical, the key factor is whether the training data's temporal resolution is sufficient to capture the core physical dynamics. In this case, halving the number of snapshots made the data too sparse for the models to learn the complex, periodic nature of vortex shedding. This suggests that the original $\Delta t = 0.01s$ was already near the minimal sampling rate required. While it may be feasible to downsample a CFD dataset generated with an unnecessarily fine time step (e.g., one chosen for solver stability rather than to resolve physics), this experiment demonstrates that coarsening a dataset below the temporal resolution required to resolve its core physical dynamics will severely degrade prediction accuracy.

\subsubsection{Generalization across multiple flow scenarios}
\label{sec:harsh_3}

\begin{figure*}[htb!]
    \centering

    \begin{subfigure}[h]{0.6\textwidth}
    \centering
        \includegraphics[width=\textwidth]{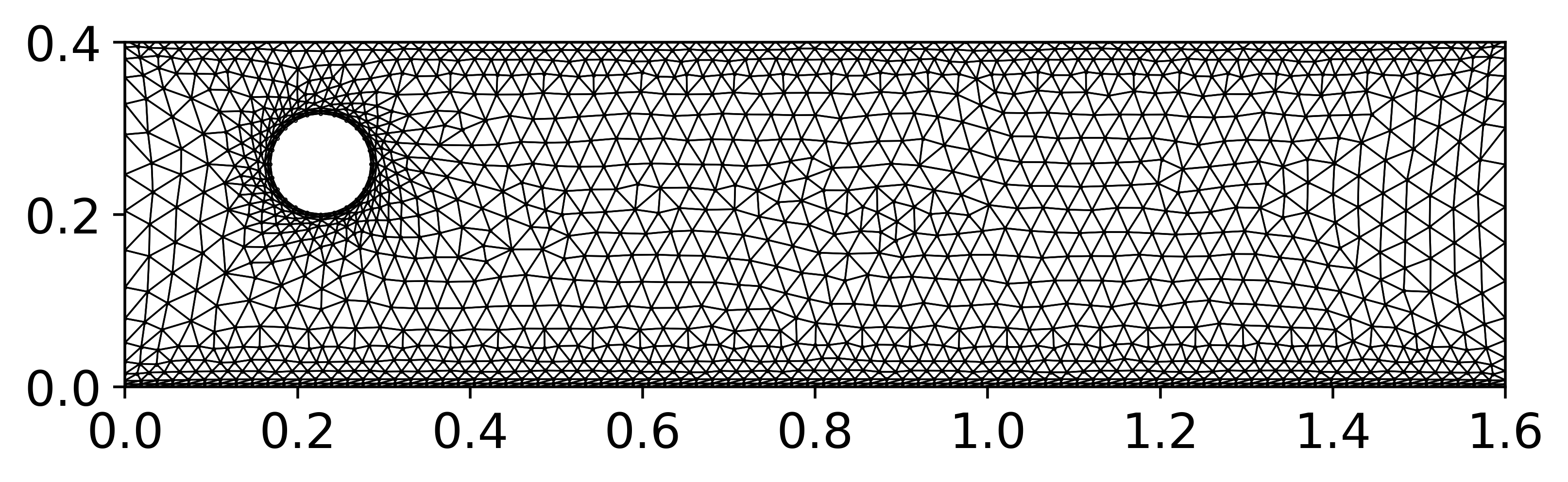}        
    \subcaption{Mesh for Scenario 2}
    \label{fig:5meshes_1}
    \end{subfigure}

    \vfill

    \begin{subfigure}[h]{0.6\textwidth}
    \centering
        \includegraphics[width=\textwidth]{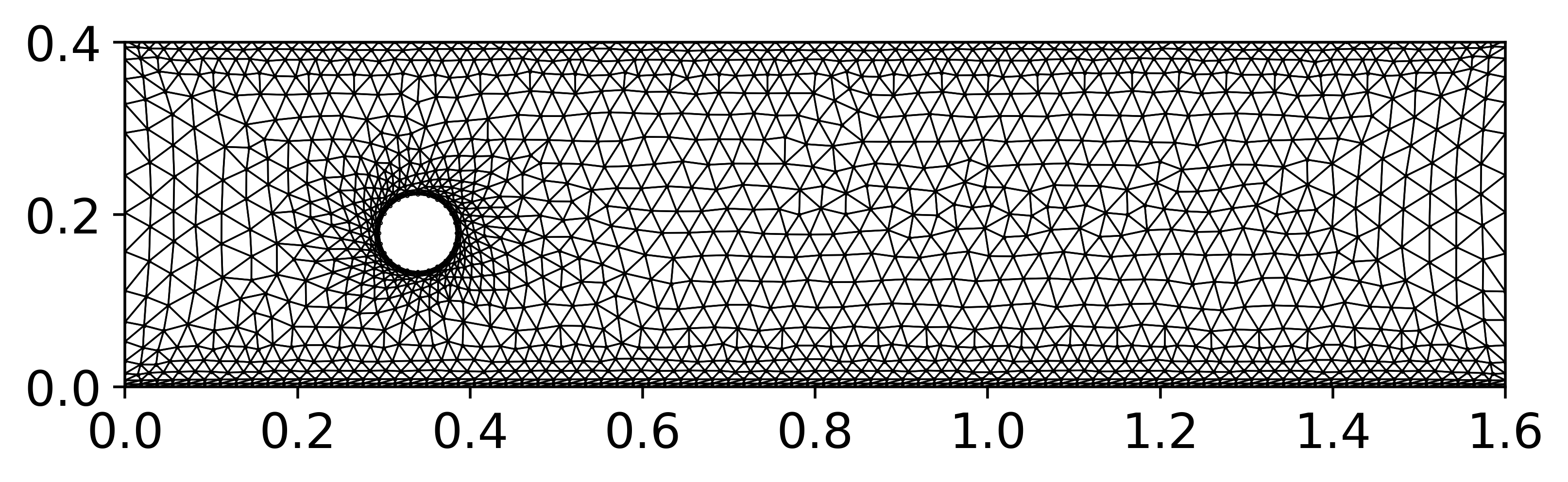}        
    \subcaption{Mesh for Scenario 3}
    \label{fig:5meshes_2}
    \end{subfigure}

    \caption{Two additional meshes used for evaluating the generalization performance across multiple flow scenarios.}\label{fig:multi_scenarios}
\end{figure*}

To evaluate the framework's ability to generalize, we again test the four leading model configurations explored in the previous robustness studies. For this experiment, each model was trained on a combined dataset comprising three distinct flow scenarios. Scenario 1 is the baseline case used throughout this paper, while Scenarios 2 and 3 are new additions with different inlet velocities, cylinder diameters, and mesh configurations (detailed in Figure \ref{fig:multi_scenarios} and Table \ref{tab:multi_scenario_setup}). By exposing the models to more varied physical dynamics during training, this multi-scenario experiment tests their ability (AW3 especially, but including AW2 --- both methods have tunable parameter $k$) to learn a more universal and robust representation of the underlying flow physics.

\begin{table}[H]
\centering
\caption{Physical and mesh properties of the three scenarios used for multi-scenario generalization testing.}
\label{tab:multi_scenario_setup}
\begin{tabular}{l|ccc}
\hline
\multirow{2}{*}{\textbf{Parameter}} & \multicolumn{3}{c}{\textbf{Scenario Type}} \\ 
\cline{2-4}
& Scenario 1 & Scenario 2 & Scenario 3 \\ 
\hline
Inlet $x$-velocity [m/s] & 1.78 & 2.21 & 2.02 \\
Cylinder diameter [m] & 0.074 & 0.116 & 0.089 \\
Number of nodes & 1,946 & 1,852 & 1,925 \\
Number of edges & 11,208 & 10,644 & 11,082 \\ 
Strouhal number & 0.1438 & 0.1472 & 0.1490 \\
\hline
\end{tabular}
\end{table}

The results in Table \ref{tab:multi_scenario_results} highlight the robust performance of the Adams-Euler scheme combined with AW3 when trained on the aggregated multi-scenario dataset. While other models struggle to perform consistently across the varied dynamics, this configuration excels, achieving the lowest MSE for Scenarios 2 and 3 while maintaining strong performance on Scenario 1. Critically, its Strouhal number predictions (0.1462, 0.1441, 0.1428) are consistently accurate across all three scenarios, closely matching their respective ground truths (0.1438, 0.1472, 0.1490). This demonstrates its superior ability to learn from a diverse distribution of physical behaviors and accurately represent the distinct dynamics across different meshes, a key capability for developing more generalizable surrogate models.

\begin{table}[htb!]
\centering
\caption{Time extrapolation performance evaluation (MSE and Strouhal number) of the four models trained on the combined multi-scenario dataset. The ground truth Strouhal numbers are 0.1438, 0.1472, and 0.1490 for Scenarios 1, 2, and 3, respectively.}
\label{tab:multi_scenario_results}
\begin{tabular}{l|cc|cc|cc}
\hline
\multirow{3}{*}{\textbf{Model Configuration}} & \multicolumn{6}{c}{\textbf{Evaluated Scenario}} \\ 
\cline{2-7}
& \multicolumn{2}{c|}{\textbf{Scenario 1}} & \multicolumn{2}{c|}{\textbf{Scenario 2}} & \multicolumn{2}{c}{\textbf{Scenario 3}} \\
& MSE & $St$ & MSE & $St$ & MSE & $St$ \\ 
\hline
Direct + Vanilla Rollout & 0.035 & 0.0088 & 0.098 & 0.1462 & 0.076 & 0.1421 \\
Forward Euler + AW2 & 0.028 & 0.1401 & 0.126 & 0.1387 & 0.078 & 0.1367 \\
Forward Euler + AW3 & 0.024 & 0.1441 & 0.068 & 0.1428 & 0.053 & 0.1421 \\
\textbf{Adams-Euler + AW3} & \textbf{0.026} & \textbf{0.1462} & \textbf{0.051} & \textbf{0.1441} & \textbf{0.042} & \textbf{0.1428} \\ 
\hline
\end{tabular}
\end{table}

% \clearpage
\section{Conclusions and Future Work} \label{sec:conclu}

This study presents a comprehensive framework for enhancing long-term auto-regressive predictions in SciML models through the novel application of numerical time-integration schemes and adaptive multi-step rollout techniques. Our systematic evaluation across canonical 2D PDEs (advection, heat, and Burgers’ equations) first established a key hypothesis: as physical complexity increases, more sophisticated rollout techniques become essential for optimal performance. This trend was decisively validated in our most challenging test case of complex Navier-Stokes dynamics. For this system, our most advanced adaptive weighting strategies (AW2/AW3) proved crucial for achieving robust, long-term accuracy, confirming the insight gained from the simpler systems. By combining the two-step Adams-Bashforth scheme with these adaptive strategies, our lightweight GNN model—containing only 1,177 trainable parameters—demonstrated meaningful effectiveness under harsh constraints, achieving accurate predictions of complex Navier-Stokes dynamics across 350 future time steps and reducing the mean squared error from 0.125 to 0.002. Overall, our integrated methodology delivers an 89\% improvement over fixed-weight multi-step rollout approach (reducing MSE from 0.018 to 0.002) and outperforms standard noise injection by 83\% (reducing MSE from 0.012 to 0.002), while maintaining robustness even on truncated meshes. This powerful yet resource-efficient framework is designed to be model-agnostic, ensuring these advancements can benefit diverse scientific domains without specialized adaptations.

Several promising directions can emerge for future research. First, investigating higher-order time integration schemes beyond the two-step Adams-Bashforth could further enhance prediction stability, particularly for systems with complex temporal dynamics. Second, developing more computationally efficient multi-step rollout strategies would reduce the gradient computation overhead inherent in training across multiple future steps. Third, incorporating domain-specific physical principles—such as conservation laws from the Navier-Stokes equations—directly into the framework could enhance both accuracy and physical consistency while potentially reducing data requirements \cite{yang2024data, yang2025physics}. While this study successfully demonstrates the framework's effectiveness on 2D laminar flows, a crucial next step is to assess its scalability to more complex systems. Future work can therefore focus on extending the framework to three-dimensional simulations and turbulent flows, which may require architectural modifications and the integration of more advanced physics-informed constraints. Furthermore, exploring its applicability to multi-physics coupling problems represents another significant avenue for future research. Finally, a particularly promising direction is to integrate our framework with state-of-the-art architectures, including Fourier neural operators or Transolver \cite{luo2025transolver++}, to validate its model-agnostic benefits and potentially push the boundaries of long-term prediction accuracy.

\section*{Conflicts of Interest} 
The authors declare no conflict of interest.

\section*{Author Contributions}
\noindent\textbf{Sunwoong Yang}: Conceptualization, Data curation, Formal analysis, Investigation, Methodology, Resources, Software, Validation, Visualization, Funding acquisition, Project administration, Writing—original draft, Writing—review \& editing. \textbf{Ricardo Vinuesa}: Project administration, Writing—review \& editing. \textbf{Namwoo Kang}: Funding acquisition, Project administration, Writing—review \& editing.

\section*{Acknowledgments}
This work was supported by the Ministry of Science and ICT of Korea grant (No. RS-2024-00355857, No. 2022-0-00969, and No. 2022-0-00986), the National Research Council of Science \& Technology (NST) grant by the Korea government (MSIT) (No. GTL24031-000), and the Ministry of Trade, Industry \& Energy (RS-2024-00410810, RS-2025-02317327). R.V. was supported by ERC Grant No. 2021-CoG-101043998, DEEPCONTROL. The views and opinions expressed are however those of the author(s) only and do not necessarily reflect those of European Union or European Research Council.

\appendix

\section{On the applicability of Runge-Kutta methods for AR prediction}
\label{sec:app_Runge}

Runge-Kutta (RK) methods \cite{abushaeer2025nonlinear, das2016heat}, especially the fourth-order scheme (RK4), are widely utilized in classical numerical integration due to their high accuracy. However, their direct application to auto-regressive (AR) prediction frameworks presents significant practical challenges, primarily related to computational efficiency. This section outlines why RK methods were not adopted in our AR prediction framework, despite their established advantages in traditional numerical methods.

\subsection{Theoretical background of RK}

The classical RK4 scheme advances the solution of the system $\frac{d\mathbf{u}}{dt} = \mathbf{f}(t, \mathbf{u})$ as follows:
\begin{align}
    k_1 &= \mathbf{f}(t, \mathbf{u}(t)) \\
    k_2 &= \mathbf{f}(t + \Delta t/2, \mathbf{u}(t) + (\Delta t/2)k_1) \\
    k_3 &= \mathbf{f}(t + \Delta t/2, \mathbf{u}(t) + (\Delta t/2)k_2) \\
    k_4 &= \mathbf{f}(t + \Delta t, \mathbf{u}(t) + \Delta tk_3)
\end{align}
with the final update given by:
\begin{equation}
    \mathbf{u}(t + \Delta t) = \mathbf{u}(t) + \frac{\Delta t}{6}(k_1 + 2k_2 + 2k_3 + k_4)
\end{equation}

\subsection{Computational considerations}

The primary challenge of RK4 within AR frameworks is computational complexity. While RK4 theoretically enhances accuracy, each time-step update requires four distinct evaluations of the underlying AI model (corresponding to $k_1$ through $k_4$), each at slightly different inputs. Consequently, both training and inference computational costs increase approximately fourfold compared to simpler schemes like Adams-Euler \cite{zhou2025predicting}, where only one evaluation per timestep is required.

Although RK4 and Adams-Euler methods both suffer from error propagation—a common issue in AR predictions—the repeated computations within a single RK4 step can amplify and propagate errors more severely. Each intermediate RK4 stage depends on prior predictions, compounding inaccuracies within each timestep. This ``multi-stage'' error accumulation can significantly degrade prediction performance over long AR rollouts. In contrast, the Adams-Euler method uses historical derivative information without intermediate stage evaluations (Eq. \ref{eq:time4_0}) using caching, mitigating this within-timestep error amplification and thus offering a balance between computational efficiency and prediction stability.

In summary, while RK4 schemes provide high accuracy in classical numerical contexts, their heavy computational requirements and increased error propagation within AR frameworks make them impractical for our approach. Consequently, the development of AR-specialized RK4 variants that mitigate computational complexity and error propagation remains an open challenge and a promising direction for future research.

% While RK methods offer superior accuracy in traditional numerical integration contexts, their fundamental requirements make them unsuitable for AR prediction tasks. The lack of ground truth for intermediate stages and the compound nature of prediction errors present critical challenges for practical implementation. These limitations explain the reason for our choice of the Adams-Bashforth scheme, which provides a more practical and robust approach to long-term AR prediction.

\section{Evolution of the weights in AW3-based multi-step rollout}
\label{sec:app_AW3}

\begin{figure}[htb!]
    \centering
    \includegraphics[width=0.95\textwidth]{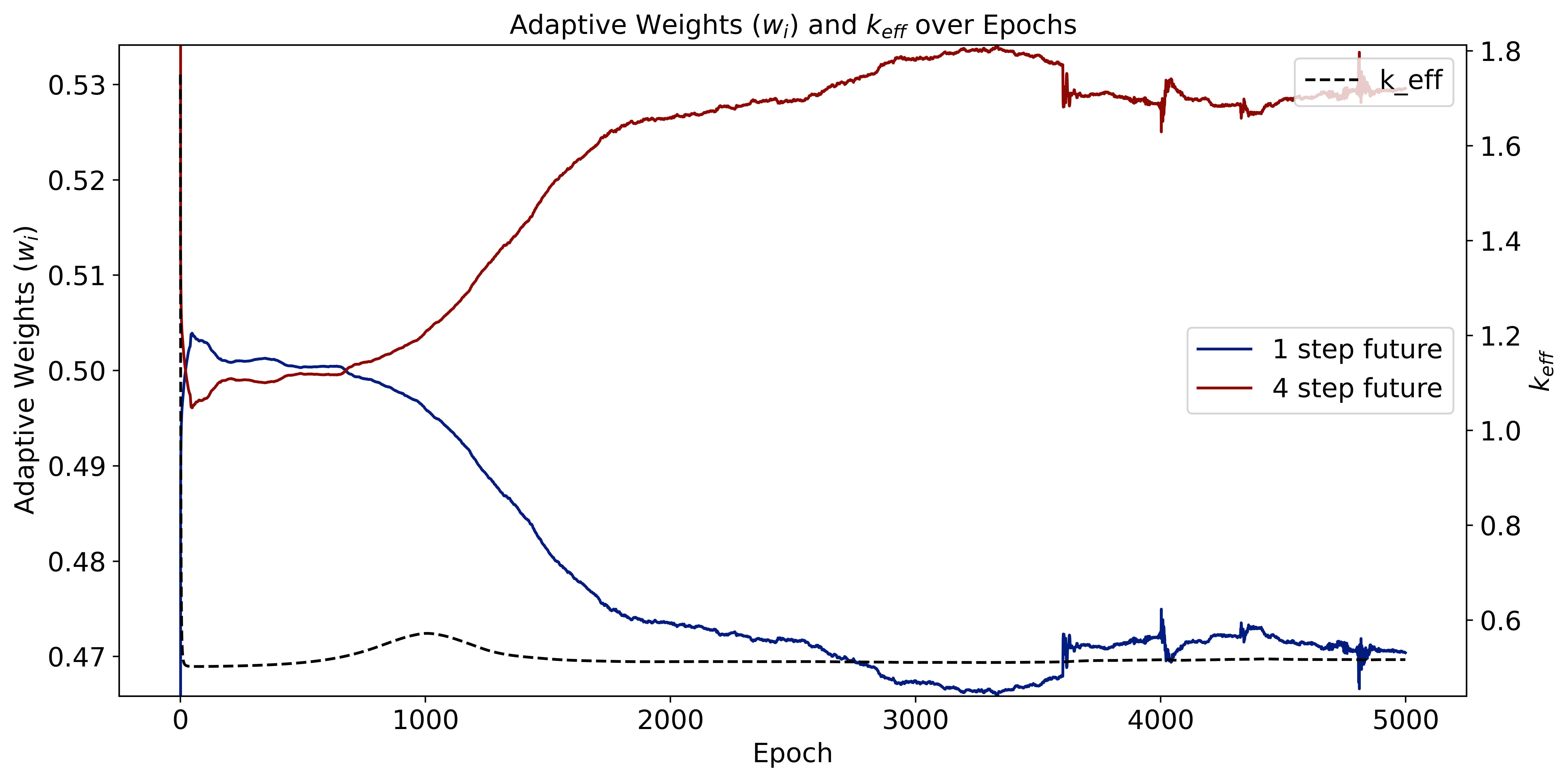}
    \caption{Evolution of adaptive weights for each loss term over epochs in AW3 with the Adams-Euler scheme.}
    \label{fig:adaptive_weights_evolution_AW3}
\end{figure}

In Section \ref{sec:AWmulti}, the weights for each loss term in the AW2-based multi-step rollout are visualized in Figure \ref{fig:adaptive_weights_evolution}. This section extends that analysis by presenting results from the AW3-based approach, using the same model configuration as in Figure \ref{fig:adaptive_weights_evolution} but applying AW3 instead of AW2. Figure \ref{fig:adaptive_weights_evolution_AW3} illustrates these results, revealing that after 1,000 epochs, the model begins to stabilize the weights of the first and last future steps, with the last time-step weight slightly exceeding that of the first. This suggests that AW3, which initially considers both the first and last time-step losses, remains highly effective by gradually shifting its focus to the last time step, ensuring greater accuracy in long-term rollouts.

\bibliographystyle{elsarticle-num-names} 
\bibliography{cas-refs}

\begin{thebibliography}{39}
\expandafter\ifx\csname natexlab\endcsname\relax\def\natexlab#1{#1}\fi
\providecommand{\url}[1]{\texttt{#1}}
\providecommand{\href}[2]{#2}
\providecommand{\path}[1]{#1}
\providecommand{\DOIprefix}{doi:}
\providecommand{\ArXivprefix}{arXiv:}
\providecommand{\URLprefix}{URL: }
\providecommand{\Pubmedprefix}{pmid:}
\providecommand{\doi}[1]{\href{http://dx.doi.org/#1}{\path{#1}}}
\providecommand{\Pubmed}[1]{\href{pmid:#1}{\path{#1}}}
\providecommand{\bibinfo}[2]{#2}
\ifx\xfnm\relax \def\xfnm[#1]{\unskip,\space#1}\fi
%Type = Article
\bibitem[{Solera-Rico et~al.(2024)Solera-Rico, Sanmiguel~Vila, G{\'o}mez-L{\'o}pez, Wang, Almashjary, Dawson, and Vinuesa}]{solera2024beta}
\bibinfo{author}{A.~Solera-Rico}, \bibinfo{author}{C.~Sanmiguel~Vila}, \bibinfo{author}{M.~G{\'o}mez-L{\'o}pez}, \bibinfo{author}{Y.~Wang}, \bibinfo{author}{A.~Almashjary}, \bibinfo{author}{S.~T. Dawson}, \bibinfo{author}{R.~Vinuesa},
\newblock \bibinfo{title}{$\beta$-variational autoencoders and transformers for reduced-order modelling of fluid flows},
\newblock \bibinfo{journal}{Nature Communications} \bibinfo{volume}{15} (\bibinfo{year}{2024}) \bibinfo{pages}{1361}.
%Type = Article
\bibitem[{Liu et~al.(2024)Liu, Zhu, Lu, Sun, and Wang}]{liu2024multi}
\bibinfo{author}{X.-Y. Liu}, \bibinfo{author}{M.~Zhu}, \bibinfo{author}{L.~Lu}, \bibinfo{author}{H.~Sun}, \bibinfo{author}{J.-X. Wang},
\newblock \bibinfo{title}{Multi-resolution partial differential equations preserved learning framework for spatiotemporal dynamics},
\newblock \bibinfo{journal}{Communications Physics} \bibinfo{volume}{7} (\bibinfo{year}{2024}) \bibinfo{pages}{31}.
%Type = Article
\bibitem[{Hasegawa et~al.(2020)Hasegawa, Fukami, Murata, and Fukagata}]{hasegawa2020cnn}
\bibinfo{author}{K.~Hasegawa}, \bibinfo{author}{K.~Fukami}, \bibinfo{author}{T.~Murata}, \bibinfo{author}{K.~Fukagata},
\newblock \bibinfo{title}{Cnn-lstm based reduced order modeling of two-dimensional unsteady flows around a circular cylinder at different reynolds numbers},
\newblock \bibinfo{journal}{Fluid Dynamics Research} \bibinfo{volume}{52} (\bibinfo{year}{2020}) \bibinfo{pages}{065501}.
%Type = Article
\bibitem[{Lee and You(2019)}]{lee2019data}
\bibinfo{author}{S.~Lee}, \bibinfo{author}{D.~You},
\newblock \bibinfo{title}{Data-driven prediction of unsteady flow over a circular cylinder using deep learning},
\newblock \bibinfo{journal}{Journal of Fluid Mechanics} \bibinfo{volume}{879} (\bibinfo{year}{2019}) \bibinfo{pages}{217--254}.
%Type = Article
\bibitem[{List et~al.(2025)List, Chen, Bali, and Thuerey}]{list2025differentiability}
\bibinfo{author}{B.~List}, \bibinfo{author}{L.-W. Chen}, \bibinfo{author}{K.~Bali}, \bibinfo{author}{N.~Thuerey},
\newblock \bibinfo{title}{Differentiability in unrolled training of neural physics simulators on transient dynamics},
\newblock \bibinfo{journal}{Computer Methods in Applied Mechanics and Engineering} \bibinfo{volume}{433} (\bibinfo{year}{2025}) \bibinfo{pages}{117441}.
%Type = Article
\bibitem[{Akhare et~al.(2023)Akhare, Luo, and Wang}]{akhare2023physics}
\bibinfo{author}{D.~Akhare}, \bibinfo{author}{T.~Luo}, \bibinfo{author}{J.-X. Wang},
\newblock \bibinfo{title}{Physics-integrated neural differentiable (pindiff) model for composites manufacturing},
\newblock \bibinfo{journal}{Computer Methods in Applied Mechanics and Engineering} \bibinfo{volume}{406} (\bibinfo{year}{2023}) \bibinfo{pages}{115902}.
%Type = Article
\bibitem[{Taieb et~al.(2012)Taieb, Bontempi, Atiya, and Sorjamaa}]{taieb2012review}
\bibinfo{author}{S.~B. Taieb}, \bibinfo{author}{G.~Bontempi}, \bibinfo{author}{A.~F. Atiya}, \bibinfo{author}{A.~Sorjamaa},
\newblock \bibinfo{title}{A review and comparison of strategies for multi-step ahead time series forecasting based on the nn5 forecasting competition},
\newblock \bibinfo{journal}{Expert systems with applications} \bibinfo{volume}{39} (\bibinfo{year}{2012}) \bibinfo{pages}{7067--7083}.
%Type = Article
\bibitem[{Ahani et~al.(2019)Ahani, Salari, and Shadman}]{ahani2019statistical}
\bibinfo{author}{I.~K. Ahani}, \bibinfo{author}{M.~Salari}, \bibinfo{author}{A.~Shadman},
\newblock \bibinfo{title}{Statistical models for multi-step-ahead forecasting of fine particulate matter in urban areas},
\newblock \bibinfo{journal}{Atmospheric Pollution Research} \bibinfo{volume}{10} (\bibinfo{year}{2019}) \bibinfo{pages}{689--700}.
%Type = Article
\bibitem[{Samal et~al.(2022)Samal, Babu, and Das}]{samal2022multi}
\bibinfo{author}{K.~K.~R. Samal}, \bibinfo{author}{K.~S. Babu}, \bibinfo{author}{S.~K. Das},
\newblock \bibinfo{title}{Multi-output spatio-temporal air pollution forecasting using neural network approach},
\newblock \bibinfo{journal}{Applied Soft Computing} \bibinfo{volume}{126} (\bibinfo{year}{2022}) \bibinfo{pages}{109316}.
%Type = Article
\bibitem[{Wang et~al.(2021)Wang, Zheng, Ji, and Guo}]{wang2021automated}
\bibinfo{author}{H.~Wang}, \bibinfo{author}{Z.~Zheng}, \bibinfo{author}{C.~Ji}, \bibinfo{author}{L.~J. Guo},
\newblock \bibinfo{title}{Automated multi-layer optical design via deep reinforcement learning},
\newblock \bibinfo{journal}{Machine Learning: Science and Technology} \bibinfo{volume}{2} (\bibinfo{year}{2021}) \bibinfo{pages}{025013}.
%Type = Article
\bibitem[{Chang and Tsai(2008)}]{chang2008forecast}
\bibinfo{author}{B.~R. Chang}, \bibinfo{author}{H.~F. Tsai},
\newblock \bibinfo{title}{Forecast approach using neural network adaptation to support vector regression grey model and generalized auto-regressive conditional heteroscedasticity},
\newblock \bibinfo{journal}{Expert systems with applications} \bibinfo{volume}{34} (\bibinfo{year}{2008}) \bibinfo{pages}{925--934}.
%Type = Article
\bibitem[{Asadi et~al.(2012)Asadi, Tavakoli, and Hejazi}]{asadi2012new}
\bibinfo{author}{S.~Asadi}, \bibinfo{author}{A.~Tavakoli}, \bibinfo{author}{S.~R. Hejazi},
\newblock \bibinfo{title}{A new hybrid for improvement of auto-regressive integrated moving average models applying particle swarm optimization},
\newblock \bibinfo{journal}{Expert Systems with Applications} \bibinfo{volume}{39} (\bibinfo{year}{2012}) \bibinfo{pages}{5332--5337}.
%Type = Article
\bibitem[{Wang et~al.(2024)Wang, Sankaran, and Perdikaris}]{wang2024respecting}
\bibinfo{author}{S.~Wang}, \bibinfo{author}{S.~Sankaran}, \bibinfo{author}{P.~Perdikaris},
\newblock \bibinfo{title}{Respecting causality for training physics-informed neural networks},
\newblock \bibinfo{journal}{Computer Methods in Applied Mechanics and Engineering} \bibinfo{volume}{421} (\bibinfo{year}{2024}) \bibinfo{pages}{116813}.
%Type = Article
\bibitem[{Liu et~al.(2022)Liu, Nath, and Cai}]{liu2022causality}
\bibinfo{author}{L.~Liu}, \bibinfo{author}{K.~Nath}, \bibinfo{author}{W.~Cai},
\newblock \bibinfo{title}{A causality-deeponet for causal responses of linear dynamical systems},
\newblock \bibinfo{journal}{arXiv preprint arXiv:2209.08397}  (\bibinfo{year}{2022}).
%Type = Inproceedings
\bibitem[{Nghiem et~al.(2023)Nghiem, Nguyen, Nguyen, and Nguyen}]{nghiem2023causal}
\bibinfo{author}{T.~X. Nghiem}, \bibinfo{author}{T.~Nguyen}, \bibinfo{author}{B.~T. Nguyen}, \bibinfo{author}{L.~Nguyen},
\newblock \bibinfo{title}{Causal deep operator networks for data-driven modeling of dynamical systems},
\newblock in: \bibinfo{booktitle}{2023 IEEE International Conference on Systems, Man, and Cybernetics (SMC)}, \bibinfo{organization}{IEEE}, \bibinfo{year}{2023}, pp. \bibinfo{pages}{1136--1141}.
%Type = Article
\bibitem[{Raissi et~al.(2019)Raissi, Perdikaris, and Karniadakis}]{raissi2019physics}
\bibinfo{author}{M.~Raissi}, \bibinfo{author}{P.~Perdikaris}, \bibinfo{author}{G.~E. Karniadakis},
\newblock \bibinfo{title}{Physics-informed neural networks: A deep learning framework for solving forward and inverse problems involving nonlinear partial differential equations},
\newblock \bibinfo{journal}{Journal of Computational physics} \bibinfo{volume}{378} (\bibinfo{year}{2019}) \bibinfo{pages}{686--707}.
%Type = Article
\bibitem[{Lu et~al.(2021)Lu, Jin, Pang, Zhang, and Karniadakis}]{lu2021learning}
\bibinfo{author}{L.~Lu}, \bibinfo{author}{P.~Jin}, \bibinfo{author}{G.~Pang}, \bibinfo{author}{Z.~Zhang}, \bibinfo{author}{G.~E. Karniadakis},
\newblock \bibinfo{title}{Learning nonlinear operators via deeponet based on the universal approximation theorem of operators},
\newblock \bibinfo{journal}{Nature machine intelligence} \bibinfo{volume}{3} (\bibinfo{year}{2021}) \bibinfo{pages}{218--229}.
%Type = Article
\bibitem[{Li et~al.(2020)Li, Kovachki, Azizzadenesheli, Liu, Bhattacharya, Stuart, and Anandkumar}]{li2020fourier}
\bibinfo{author}{Z.~Li}, \bibinfo{author}{N.~Kovachki}, \bibinfo{author}{K.~Azizzadenesheli}, \bibinfo{author}{B.~Liu}, \bibinfo{author}{K.~Bhattacharya}, \bibinfo{author}{A.~Stuart}, \bibinfo{author}{A.~Anandkumar},
\newblock \bibinfo{title}{Fourier neural operator for parametric partial differential equations},
\newblock \bibinfo{journal}{arXiv preprint arXiv:2010.08895}  (\bibinfo{year}{2020}).
%Type = Article
\bibitem[{Pfaff et~al.(2020)Pfaff, Fortunato, Sanchez-Gonzalez, and Battaglia}]{pfaff2020learning}
\bibinfo{author}{T.~Pfaff}, \bibinfo{author}{M.~Fortunato}, \bibinfo{author}{A.~Sanchez-Gonzalez}, \bibinfo{author}{P.~W. Battaglia},
\newblock \bibinfo{title}{Learning mesh-based simulation with graph networks},
\newblock \bibinfo{journal}{arXiv preprint arXiv:2010.03409}  (\bibinfo{year}{2020}).
%Type = Inproceedings
\bibitem[{Venkatraman et~al.(2015)Venkatraman, Hebert, and Bagnell}]{venkatraman2015improving}
\bibinfo{author}{A.~Venkatraman}, \bibinfo{author}{M.~Hebert}, \bibinfo{author}{J.~Bagnell},
\newblock \bibinfo{title}{Improving multi-step prediction of learned time series models},
\newblock in: \bibinfo{booktitle}{Proceedings of the AAAI Conference on Artificial Intelligence}, volume~\bibinfo{volume}{29}, \bibinfo{year}{2015}.
%Type = Article
\bibitem[{Kim et~al.(2025)Kim, Park, Kim, Yu, Chang, Woo, Yang, and Kang}]{kim2024physics}
\bibinfo{author}{J.~Kim}, \bibinfo{author}{J.~Park}, \bibinfo{author}{N.~Kim}, \bibinfo{author}{Y.~Yu}, \bibinfo{author}{K.~Chang}, \bibinfo{author}{C.-S. Woo}, \bibinfo{author}{S.~Yang}, \bibinfo{author}{N.~Kang},
\newblock \bibinfo{title}{Physics-constrained graph neural networks for spatio-temporal prediction of drop impact on oled display panels},
\newblock \bibinfo{journal}{Expert Systems with Applications}  (\bibinfo{year}{2025}) \bibinfo{pages}{126907}.
%Type = Article
\bibitem[{Zhang et~al.(2025)Zhang, Liu, Chen, Wei, Wu, and Dai}]{zhang2025goal}
\bibinfo{author}{X.~Zhang}, \bibinfo{author}{J.~Liu}, \bibinfo{author}{C.~Chen}, \bibinfo{author}{L.~Wei}, \bibinfo{author}{Z.~Wu}, \bibinfo{author}{W.~Dai},
\newblock \bibinfo{title}{Goal-driven long-term marine vessel trajectory prediction with a memory-enhanced network},
\newblock \bibinfo{journal}{Expert Systems with Applications} \bibinfo{volume}{263} (\bibinfo{year}{2025}) \bibinfo{pages}{125715}.
%Type = Article
\bibitem[{Gao et~al.(2024)Gao, Han, Fan, Sun, Liu, Duan, and Wang}]{gao2024bayesian}
\bibinfo{author}{H.~Gao}, \bibinfo{author}{X.~Han}, \bibinfo{author}{X.~Fan}, \bibinfo{author}{L.~Sun}, \bibinfo{author}{L.-P. Liu}, \bibinfo{author}{L.~Duan}, \bibinfo{author}{J.-X. Wang},
\newblock \bibinfo{title}{Bayesian conditional diffusion models for versatile spatiotemporal turbulence generation},
\newblock \bibinfo{journal}{Computer Methods in Applied Mechanics and Engineering} \bibinfo{volume}{427} (\bibinfo{year}{2024}) \bibinfo{pages}{117023}.
%Type = Article
\bibitem[{Jeon et~al.(2022)Jeon, Lee, and Kim}]{jeon2022finite}
\bibinfo{author}{J.~Jeon}, \bibinfo{author}{J.~Lee}, \bibinfo{author}{S.~J. Kim},
\newblock \bibinfo{title}{Finite volume method network for the acceleration of unsteady computational fluid dynamics: Non-reacting and reacting flows},
\newblock \bibinfo{journal}{International Journal of Energy Research} \bibinfo{volume}{46} (\bibinfo{year}{2022}) \bibinfo{pages}{10770--10795}.
%Type = Article
\bibitem[{Jeon et~al.(2024)Jeon, Lee, Vinuesa, and Kim}]{jeon2024residual}
\bibinfo{author}{J.~Jeon}, \bibinfo{author}{J.~Lee}, \bibinfo{author}{R.~Vinuesa}, \bibinfo{author}{S.~J. Kim},
\newblock \bibinfo{title}{Residual-based physics-informed transfer learning: A hybrid method for accelerating long-term cfd simulations via deep learning},
\newblock \bibinfo{journal}{International Journal of Heat and Mass Transfer} \bibinfo{volume}{220} (\bibinfo{year}{2024}) \bibinfo{pages}{124900}.
%Type = Inproceedings
\bibitem[{Sanchez-Gonzalez et~al.(2020)Sanchez-Gonzalez, Godwin, Pfaff, Ying, Leskovec, and Battaglia}]{sanchez2020learning}
\bibinfo{author}{A.~Sanchez-Gonzalez}, \bibinfo{author}{J.~Godwin}, \bibinfo{author}{T.~Pfaff}, \bibinfo{author}{R.~Ying}, \bibinfo{author}{J.~Leskovec}, \bibinfo{author}{P.~Battaglia},
\newblock \bibinfo{title}{Learning to simulate complex physics with graph networks},
\newblock in: \bibinfo{booktitle}{International conference on machine learning}, \bibinfo{organization}{PMLR}, \bibinfo{year}{2020}, pp. \bibinfo{pages}{8459--8468}.
%Type = Article
\bibitem[{Yang et~al.(2024)Yang, Vinuesa, and Kang}]{yang2024enhancing}
\bibinfo{author}{S.~Yang}, \bibinfo{author}{R.~Vinuesa}, \bibinfo{author}{N.~Kang},
\newblock \bibinfo{title}{Enhancing graph u-nets for mesh-agnostic spatio-temporal flow prediction},
\newblock \bibinfo{journal}{arXiv preprint arXiv:2406.03789}  (\bibinfo{year}{2024}).
%Type = Article
\bibitem[{Zhou and Farimani(2025)}]{zhou2025predicting}
\bibinfo{author}{A.~Zhou}, \bibinfo{author}{A.~B. Farimani},
\newblock \bibinfo{title}{Predicting change, not states: An alternate framework for neural pde surrogates},
\newblock \bibinfo{journal}{Computer Methods in Applied Mechanics and Engineering} \bibinfo{volume}{441} (\bibinfo{year}{2025}) \bibinfo{pages}{117990}.
%Type = Article
\bibitem[{Hussain et~al.(2025)Hussain, Lin, Waqas, and Al-Mdallal}]{hussain2025integrating}
\bibinfo{author}{M.~Hussain}, \bibinfo{author}{D.~Lin}, \bibinfo{author}{H.~Waqas}, \bibinfo{author}{Q.~M. Al-Mdallal},
\newblock \bibinfo{title}{Integrating artificial intelligence and machine learning with numerical simulation for enhanced thermal performance of ternary nanofluid},
\newblock \bibinfo{journal}{Journal of Computational Design and Engineering} \bibinfo{volume}{12} (\bibinfo{year}{2025}) \bibinfo{pages}{62--77}.
%Type = Inproceedings
\bibitem[{Wu et~al.(2022)Wu, Wang, Zhang, Ying, Cao, Sosic, Jalali, Hamam, Maucec, and Leskovec}]{wu2022learning}
\bibinfo{author}{T.~Wu}, \bibinfo{author}{Q.~Wang}, \bibinfo{author}{Y.~Zhang}, \bibinfo{author}{R.~Ying}, \bibinfo{author}{K.~Cao}, \bibinfo{author}{R.~Sosic}, \bibinfo{author}{R.~Jalali}, \bibinfo{author}{H.~Hamam}, \bibinfo{author}{M.~Maucec}, \bibinfo{author}{J.~Leskovec},
\newblock \bibinfo{title}{Learning large-scale subsurface simulations with a hybrid graph network simulator},
\newblock in: \bibinfo{booktitle}{Proceedings of the 28th ACM SIGKDD Conference on Knowledge Discovery and Data Mining}, \bibinfo{year}{2022}, pp. \bibinfo{pages}{4184--4194}.
%Type = Article
\bibitem[{Elman(1993)}]{elman1993learning}
\bibinfo{author}{J.~L. Elman},
\newblock \bibinfo{title}{Learning and development in neural networks: The importance of starting small},
\newblock \bibinfo{journal}{Cognition} \bibinfo{volume}{48} (\bibinfo{year}{1993}) \bibinfo{pages}{71--99}.
%Type = Article
\bibitem[{Zhou et~al.(2024)Zhou, Lorsung, Hemmasian, and Farimani}]{zhou2024strategies}
\bibinfo{author}{A.~Zhou}, \bibinfo{author}{C.~Lorsung}, \bibinfo{author}{A.~Hemmasian}, \bibinfo{author}{A.~B. Farimani},
\newblock \bibinfo{title}{Strategies for pretraining neural operators},
\newblock \bibinfo{journal}{arXiv preprint arXiv:2406.08473}  (\bibinfo{year}{2024}).
%Type = Inproceedings
\bibitem[{Gao and Ji(2019)}]{gao2019graph}
\bibinfo{author}{H.~Gao}, \bibinfo{author}{S.~Ji},
\newblock \bibinfo{title}{Graph u-nets},
\newblock in: \bibinfo{booktitle}{international conference on machine learning}, \bibinfo{organization}{PMLR}, \bibinfo{year}{2019}, pp. \bibinfo{pages}{2083--2092}.
%Type = Article
\bibitem[{Ogoke et~al.(2021)Ogoke, Meidani, Hashemi, and Farimani}]{ogoke2021graph}
\bibinfo{author}{F.~Ogoke}, \bibinfo{author}{K.~Meidani}, \bibinfo{author}{A.~Hashemi}, \bibinfo{author}{A.~B. Farimani},
\newblock \bibinfo{title}{Graph convolutional networks applied to unstructured flow field data},
\newblock \bibinfo{journal}{Machine Learning: Science and Technology} \bibinfo{volume}{2} (\bibinfo{year}{2021}) \bibinfo{pages}{045020}.
%Type = Article
\bibitem[{Yang et~al.(2024)Yang, Kim, Hong, Yee, Maulik, and Kang}]{yang2024data}
\bibinfo{author}{S.~Yang}, \bibinfo{author}{H.~Kim}, \bibinfo{author}{Y.~Hong}, \bibinfo{author}{K.~Yee}, \bibinfo{author}{R.~Maulik}, \bibinfo{author}{N.~Kang},
\newblock \bibinfo{title}{Data-driven physics-informed neural networks: A digital twin perspective},
\newblock \bibinfo{journal}{Computer Methods in Applied Mechanics and Engineering} \bibinfo{volume}{428} (\bibinfo{year}{2024}) \bibinfo{pages}{117075}.
%Type = Article
\bibitem[{Yang et~al.(2025)Yang, Lee, and Kang}]{yang2025physics}
\bibinfo{author}{S.~Yang}, \bibinfo{author}{Y.~Lee}, \bibinfo{author}{N.~Kang},
\newblock \bibinfo{title}{Physics-guided multi-fidelity deeponet for data-efficient flow field prediction},
\newblock \bibinfo{journal}{arXiv preprint arXiv:2503.17941}  (\bibinfo{year}{2025}).
%Type = Article
\bibitem[{Luo et~al.(2025)Luo, Wu, Zhou, Xing, Di, Wang, and Long}]{luo2025transolver++}
\bibinfo{author}{H.~Luo}, \bibinfo{author}{H.~Wu}, \bibinfo{author}{H.~Zhou}, \bibinfo{author}{L.~Xing}, \bibinfo{author}{Y.~Di}, \bibinfo{author}{J.~Wang}, \bibinfo{author}{M.~Long},
\newblock \bibinfo{title}{Transolver++: An accurate neural solver for pdes on million-scale geometries},
\newblock \bibinfo{journal}{arXiv preprint arXiv:2502.02414}  (\bibinfo{year}{2025}).
%Type = Article
\bibitem[{AbuShaeer et~al.(2025)AbuShaeer, Abu-Bakr, Eid, Elsaid, Alqarni, and Abu-Nab}]{abushaeer2025nonlinear}
\bibinfo{author}{Z.~AbuShaeer}, \bibinfo{author}{A.~F. Abu-Bakr}, \bibinfo{author}{M.~R. Eid}, \bibinfo{author}{E.~M. Elsaid}, \bibinfo{author}{A.~J. Alqarni}, \bibinfo{author}{A.~K. Abu-Nab},
\newblock \bibinfo{title}{Nonlinear acoustic multicavitation with external ultrasound field in complex fluids: Numerical investigation},
\newblock \bibinfo{journal}{Journal of Computational Design and Engineering}  (\bibinfo{year}{2025}) \bibinfo{pages}{qwaf055}.
%Type = Article
\bibitem[{Das et~al.(2016)Das, Sharma, and Sarkar}]{das2016heat}
\bibinfo{author}{K.~Das}, \bibinfo{author}{R.~P. Sharma}, \bibinfo{author}{A.~Sarkar},
\newblock \bibinfo{title}{Heat and mass transfer of a second grade magnetohydrodynamic fluid over a convectively heated stretching sheet},
\newblock \bibinfo{journal}{Journal of Computational Design and Engineering} \bibinfo{volume}{3} (\bibinfo{year}{2016}) \bibinfo{pages}{330--336}.

\end{thebibliography}
% \bibliographystyle{elsarticle-num-names} 
% \bibliography{cas-refs}
% \printbibliography
% \bibliographystyle{model5-names}
% \bibliography{cas-refs}
% \biboptions{authoryear}





% \end{thebibliography}
\end{document}